\documentclass[review]{elsarticle}
\usepackage{hyperref}
\usepackage{float}
\usepackage{verbatim} 
\usepackage{natbib}
\restylefloat{figure}
\floatstyle{plaintop} 
\restylefloat{table}
\usepackage{xcolor}
\usepackage{amssymb}
\usepackage{amsmath}
\usepackage[font=small]{subcaption}
\usepackage{algorithm}
\usepackage{algpseudocode}
\usepackage{comment}
\usepackage[para,online,flushleft]{threeparttable}
\usepackage{array}
\usepackage{hyperref}
\newcolumntype{P}[1]{>{\centering\arraybackslash}p{#1}}
\newcolumntype{M}[1]{>{\centering\arraybackslash}m{#1}}

\journal{Healthcare Analytics}

\begin{document}
\begin{frontmatter}

\title{
A Multi-Agent Reinforcement Learning Framework for Public Health Decision Analysis}
\author[label1]{Dinesh Sharma}
\ead{dsharma1@usf.edu}

\author[label2]{Ankit Shah}
\ead{ankit@iu.edu}

\author[label3]{Chaitra Gopalappa \corref{cor1}}
\ead{chaitrag@umass.edu}

\cortext[cor1]{Corresponding author.}
\address[label1]{University of South Florida, Department of Industrial and Management Systems Engineering, 4202 E Fowler Ave, Tampa, 33620, FL, USA}
\address[label2]{Indiana University, Department of Operations and Decision Technologies, 1309 E. 10th Street, Bloomington, 47405, IN, USA}
\address[label3]{University of Massachusetts Amherst, Department of Mechanical and Industrial Engineering, 300 Massachusetts Ave, Amherst, 01003, MA, USA}
\begin{abstract}
Human immunodeficiency virus (HIV) is a major public health concern in the United States (U.S.), with about 1.2 million people living with it and about 35,000 newly infected each year. There are considerable geographical disparities in HIV burden and care access across the U.S. The 'Ending the HIV Epidemic (EHE)’ initiative by the U.S. Department of Health and Human Services aims to reduce new infections by 90\% by 2030, by improving coverage of diagnoses, treatment, and prevention interventions and prioritizing jurisdictions with high HIV prevalence. One of the approaches towards achieving this objective includes developing intelligent decision-support systems that can help optimize resource allocation and intervention strategies. Existing decision analytic models either focus on individual cities or aggregate national data, failing to capture jurisdictional interactions critical for optimizing intervention strategies. To address this, we propose a multi-agent reinforcement learning (MARL) framework that enables jurisdiction-specific decision-making while accounting for cross-jurisdictional epidemiological interactions. Our framework functions as an intelligent resource optimization system, helping policymakers strategically allocate interventions based on dynamic, data-driven insights. Experimental results across jurisdictions in California and Florida demonstrate that MARL-driven policies outperform traditional single-agent reinforcement learning approaches by reducing new infections under fixed budget constraints. Our study highlights the importance of incorporating jurisdictional dependencies in decision-making frameworks for large-scale public initiatives. By integrating multi-agent intelligent systems, decision analytics, and reinforcement learning, this study advances expert systems for government resource planning and public health management, offering a scalable framework for broader applications in healthcare policy and epidemic management.
\end{abstract}





\end{frontmatter}

\section{Introduction}
\label{intro}

Human immunodeficiency virus (HIV) remains a significant public health challenge in the United States (U.S.), with over 1.2 million individuals living with HIV. While there is no cure, early diagnosis and antiretroviral therapy (ART) effectively contribute to viral load suppression (VLS), preventing transmissions~\citep{usstatistics}. Additionally, pre-exposure prophylaxis (PrEP) has proven to reduce HIV acquisition by 99\%~\citep{hivprep}. Despite these advances, reducing new HIV infections is a persistent challenge, with significant geographic and demographic disparities in infection rates. In response, the U.S. Department of Health and Human Services launched the 'Ending the HIV Epidemic (EHE)' initiative in 2019, aiming to reduce HIV incidence by 75\% in key jurisdictions by 2025 and by 90\% across the country by 2030 \citep{usehe, hivaheadfaq}. Meeting these ambitious EHE targets requires scaling up exiting interventions such as (i) testing for early diagnosis of HIV infections, (ii) rapid and effective treatment with ART to achieve VLS, and (iii) adequate HIV prevention measures, including PrEP for groups at high-risk of infection, as well as improving how limited resources are allocated across jurisdictions and risk groups in a way that maximizes epidemic control.

Existing models in HIV literature that have evaluated the goals of the EHE initiative can be classified into two categories: independent jurisdictional models, where decisions are assessed for each jurisdiction individually, and national models, where decisions are evaluated at the aggregated national level. These models face notable limitations. National level models \cite{khatami2021reinforcement,khurana2018impact} aggregate decisions at the national level, thereby overlooking population heterogeneity of local epidemics and disparities in care and treatment across jurisdictions. On the other hand, independent jurisdictional models \cite{krebs2019developing,zang2020development,nosyk2019ending,wheatley2022cost} evaluate decisions specific to each jurisdiction but fail to account for epidemic interactions across jurisdictions, particularly those occurring through sexual partnership mixing. Another limitation of existing models is their reliance on scenario-based analyses, where a small number of pre-selected scenarios are simulated for comparison. Given the large number of jurisdictions and the multiple possible strategies for sequentially scaling up interventions over the multi-year horizon of the EHE initiative, such analyses are insufficient for identifying an optimal strategy. ~\citet{khatami2021reinforcement} recently demonstrated the use of reinforcement learning (RL), a sequential decision-making approach, to identify optimal intervention scale-up strategies in a national model. However, this approach focused on a single strategy for the U.S. as a whole, ignoring the jurisdictional heterogeneity across the country.

To address these gaps in existing HIV modeling literature, we propose a decision-support framework based on multi-agent reinforcement learning (MARL), leveraging deep reinforcement learning (DRL) techniques. In this approach, each jurisdiction is represented as an autonomous agent that learns intervention strategies specific to its epidemic context, while accounting for cross-jurisdictional interactions. By modeling HIV prevention and treatment decisions as a sequential optimization problem, MARL enables dynamic allocation of resources to testing, ART retention, and PrEP scale-up under budget constraints. This adaptive approach contrasts with traditional optimization methods and static models, offering a pathway to more efficient and effective epidemic control.

The objective of this study is threefold: (i) to demonstrate the feasibility of MARL for informing HIV intervention strategies, (ii) to compare its performance with classical optimization baselines and existing modeling approaches, and (iii) to evaluate its effectiveness in two representative states, California and Florida, as case studies.

The main contributions of this study are as follows:

\begin{enumerate}
    \item \textbf{A novel decision-support framework for HIV policy design}. We introduce a MARL approach that identifies intervention strategies tailored to the unique epidemic context of each jurisdiction, enabling adaptive allocation of testing, treatment, and prevention resources while accounting for cross-jurisdictional interactions.

    \item \textbf{Benchmarking against existing models}. We compare MARL with two single-agent reinforcement learning (SARL) approaches: national aggregated SARL (A-SARL)and independent jurisdiction SARL (I-SARL), which represent two prevalent models in current HIV literature, as well as classical optimization baselines, providing the systematic benchmark of MARL against methods commonly used in HIV modeling.

    \item \textbf{Evaluation of policy-relevant scenarios}. We assess intervention strategies under various budget constraints, analyzing how dynamic allocations can accelerate progress toward EHE targets and exploring the trade-offs between testing, treatment, and prevention.

   \item \textbf{Decision-analytic tool for policy analyses}. The model serves as a suitable decision analytic tool for national-level analyses of investments in testing, ART retention, and PrEP  for achieving the goal of ending the HIV epidemic.

\end{enumerate}

The rest of the paper is organized as follows: Section \ref{lit_review} discusses related literature. Section \ref{methodology} offers an overview of our simulation model and proposed decision analytics approaches. Section \ref{exp} presents the experimental setup of our study, while Section \ref{results} provides the computational results. Section \ref{discuss} discusses findings and limitations, and Lastly, Section \ref{conclude} provides paper's conclusions.

\section{Literature Review} \label{lit_review}

Several comprehensive reviews \cite{avancena2020optimization, wang2024optimal, silal2021operational} have summarized decision-analytic and optimization-based frameworks used in epidemic modeling and public health resource allocation. Building on these foundations, this section reviews broader decision-analytic approaches across public health domains, including HIV and COVID-19 intervention modeling. 
\citet{kok2015optimizing} developed an optimal resource allocation model using a system dynamics approach to the continuum of HIV care, aiming to minimize new HIV infections. \citet{gopalappa2012cost} and \citet{lin2016cost} focused on assessing the cost-effectiveness of various interventions in HIV prevention. Additionally, several studies have developed optimization models for allocating HIV funds to enhance the effectiveness of HIV prevention efforts \cite{lasry2011model,zaric2001optimal,lasry2012allocating,yaylali2016theory,juusola2016hiv,gromov2018numerical,yaylali2018optimal}. In the context of another recent public health challenge, COVID-19, several studies have utilized optimization models to determine vaccine prioritization strategies when vaccine supply is limited \cite{matrajt2021vaccine,yu2021dynamic,han2021time,pino2023optimization}. \citet{pino2023optimization} developed a linear programming model with the objective of minimizing both primary and breakthrough infections while considering vaccine effectiveness, waning immunity, various age groups, population data, and weekly vaccination rates. The model provided essential guidance to policymakers, as it allowed the identification of efficient vaccination schedule and allocation strategy across different regions. \citet{olayiwola2023caputo} studied the impact of high-risk quarantine and vaccination strategies on COVID-19 transmission and showed that, while both strategies reduce virus prevalence, their combined application leads to a faster decline in disease prevalence and new cases, highlighting the need for multiple intervention approaches to be applied simultaneously. In terms of other resource allocation, \citet{silveira2024multi} developed multi-stage optimization approach to plan the location and distribution of intensive care unit beds in case of a shortage of beds during epidemic events using the case study from COVID-19 data. Several studies \cite{libotte2020determination,tsay2020modeling,rawson2020and} have investigated optimal control strategies for managing the COVID-19 pandemic.  However, the above models either use static optimization, which is not suitable for sequential decision analyses, i.e., evaluating different ways to scale up interventions over time toward the goal of ending an epidemic, or they fail to account for regional or population differences in intervention measures and overall decision-making.

RL is a sequential decision-analyses technique, which makes it ideal for the analysis of public health decisions, such as resource allocation during a pandemic, monitoring and testing strategies, and adaptive sampling for hidden populations \cite{weltz2022reinforcement}. The application of RL in the public health domain has experienced significant growth in recent times. Several studies \cite{kwak2021deep,awasthi2022vacsim,kompella2020reinforcement,kumar2021recurrent} used DRL models to optimize various COVID-19 mitigation policies. \citet{libin2021deep} employed DRL to model the spread of pandemic influenza in Great Britain. \citet{bednarski2021collaborative} investigated the use of RL models to facilitate the redistribution of medical equipment, enhancing the public health response in preparation for future crises. The DRL algorithms used in the above works include Deep Q-Network (DQN) \cite{mnih2013playing}, Soft Actor-Critic (SAC) \cite{haarnoja2018soft}, and Proximal Policy Optimization (PPO) \cite{schulman2017proximal}. Notably, PPO has demonstrated strong performance in addressing problems with high-dimensional (epidemic) state spaces and continuous action (decision) spaces.

Recently, MARL algorithms have found applications across various fields, including resource allocation, robot path planning, production systems, and maintenance management \cite{oroojlooy2022review}. \citet{lowe2017multi} proposed multi-agent actor critic method approach, that utilized centralized training decentralized execution (CTDE) structure. \citet{yu2022surprising} formulated their problem as decentralized partially observable markov decision process (Dec-POMDP) with similar CTDE structure to \cite{lowe2017multi} and extended the PPO to multi-agent setting as MAPPO (Multi-Agent PPO). \citet{chu2019multi} and \citet{yang2020urban} utilized MARL in traffic signal control. \citet{nasir2019multi}, \citet{lin2018efficient} and \citet{yu2020multi} used various multi-agent algorithms in wireless networks, fleet management and heating, ventilation, and air conditioning, respectively. To the best of our knowledge, MARL has not been applied to public health.

\section{Methodology}\label{methodology}
In this paper, we formulate MARL to find jurisdiction-specific optimal intervention policies to achieve the EHE goal, i.e., move towards 90\% reduction in HIV infections in the U.S. The general framework of MARL includes a) a decentralized Markov decision process (Dec-MDP) to formulate the decision-making problem, b) a  multi-jurisdictional simulation model to evaluate a policy (sequence of decisions) and c) a solution algorithm for guiding policy selection. In this context, 
we evaluated independent proximal policy optimization (IPPO) and CTDE.

 \begin{figure}[h]
\centering  
\includegraphics[width=0.95\linewidth]{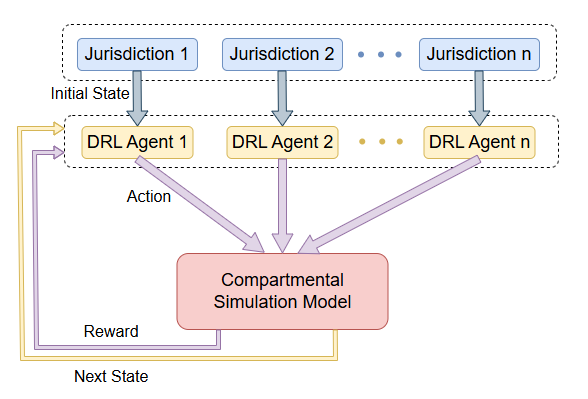} 
\caption{Schematic of the multi-agent reinforcement learning (MARL) framework.}\label{fig:ma_overview_diagram}
\end{figure}

Figure \ref{fig:ma_overview_diagram} provides a schematic overview of our MARL framework where the actions from an agent are passed to the compartmental simulation model to simulate the population and get the next state and reward. The rest of the methods is organized as follows: in \ref{sim_model} we discuss the simulation, in \ref{decmdp_formulation} we discuss formulation of Dec-MDP, and in \ref{drl_methodology} we discuss the solution algorithms. 

\subsection{Simulation Model} \label{sim_model}
\begin{figure}[h]
\centering  
\includegraphics[width=0.99\linewidth]{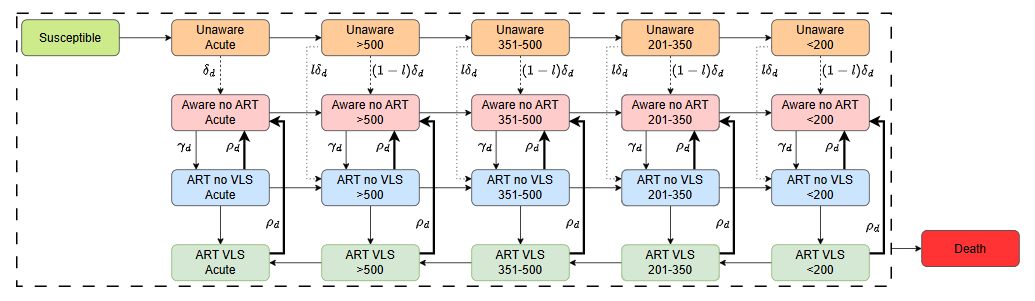} 
\caption{Overview of compartmental simulation model.}\label{fig:compartmental_model}
\end{figure}

The disease progression and transmission of HIV is simulated through the compartmental simulation model developed in \cite{tatapudi2022evaluating}. An overview of this compartmental simulation is provided in Figure \ref{fig:compartmental_model}. The model encompasses four stages within the care continuum: Unaware, Aware no ART, ART no VLS, and ART VLS. Each of these care continuum stages is further divided into five disease progression stages based on CD4 count: Acute, CD4 $> 500$, CD4 351-500, CD4 201-350, and CD4 $< 200$. Additionally, the model includes compartments for individuals who are susceptible to infection and those who have succumbed to the disease, resulting in a total of 22 compartments.
In the model, $\delta_{d}$ represents the diagnostic rate, $\rho_{d}$ represents the dropout rate, and $\gamma_{d}$ represents the rate of entry to care in disease stage $d$. Furthermore, the parameter $l$ signifies the proportion of individuals who are linked to care within three months of diagnosis. We model three transmission risk groups,
heterosexual males ($HM$ ), heterosexual females ($HF$), and men who have sex with men ($MSM$). We modeled mixing across the transmission risk groups, and mixing across jurisdictions, using data specific to transmission risk groups. Details on input data, calibration, and model validation are presented in \cite{tatapudi2022evaluating}. 
In summary, each jurisdiction is a compartmental model with 66 compartments, with a mixing matrix modeling the sexual partnership mixing across jurisdictions.

\subsubsection{Model validation}

The simulation model was validated by comparing the model estimated new incidence with estimates from the U.S. National HIV Surveillance System (NHSS). Further details about the model development, including comprehensive calibration, sensitivity analysis, model verification, and model validation are presented in \cite{tatapudi2022evaluating}.

\subsection{Decentralized Markov Decision Process} \label{decmdp_formulation}
The general formulation of Dec-MDP is a tuple $(J, S, O, a, P, R, \gamma)$. 
$J$ is the number of agents representing the total jurisdictions. $S$ is the state space, and corresponds to the overall epidemic state. $o^{j}$ is the observation space for agent $j$ and $O$ represents the set of observations where $O = [o^{1}, o^{2}, ..., o^{J}]$. $a^{j}$ is the action space for agent $j$. Consequently, the total collection of actions, denoted as $a$, is defined as the Cartesian product: $a = a^{1} \times a^{2} \times \ldots \times a^{J}$. $P$ is a state transition function that determines the state transition from $s \in S$ at $t$ to $s' \in S$ at $t+1$. $R$ is the set of reward functions ($R^{j}$) of each agent $j$. Lastly, the discount factor $\gamma \in [0,1]$ is utilized to adjust for future rewards.
The objective of the problem is to then solve for the optimal  policy $\pi(a^{j}|o^{j})$, which is a function that defines what action $a^{j}$ to take when the observation is $o^{j}$. We discuss below, the elements of the Dec-MDP specific to this application.

\subsubsection{Observation and State}
We formulate the observation for a jurisdiction $j \in J$ at time $t$ as, 
\begin{equation}\label{eq:1}
    o_{t}^{j} = [p_{k,j,t}, \mu_{u,k,j,t}, \mu_{a,k,j,t}, \mu_{ART,k,j,t}, \mu_{prep,k,j,t}; \forall k \in \{HM, HF, MSM\}]
\end{equation}


\begin{table}[t]
\centering
\begin{tabular}{|M{1.5cm}|M{9cm}|}
\hline
\textbf{Notation}      & \textbf{Definition}    \\ \hline
 $p_{k,j,t}$ & proportion of people living with HIV (PWH), diagnosed or undiagnosed, for each risk group $k \in \{HM, HF, MSM\}$ in jurisdiction $j \in {J}$ at time $t$ \\ 
 \hline
 $\mu_{u,k,j,t}$ & proportion of PWH unaware of their infection for each risk group $k \in \{HM, HF, MSM\}$ in jurisdiction $j \in {J}$ at time $t$ \\
 \hline
 $\mu_{a,k,j,t}$ & proportion of PWH aware of their infection but not on antiretroviral therapy (ART) for each risk group $k \in \{HM, HF, MSM\}$ in jurisdiction $j \in {J}$ at time $t$ \\
 \hline
 $\mu_{ART,k,j,t}$ & proportion of PWH aware of their infection and on ART or on viral load suppression (VLS) for each risk group $k \in \{HM, HF, MSM\}$ in jurisdiction $j \in {J}$ at time $t$ \\
 \hline
 $\mu_{prep,k,j,t}$ & proportion of people on pre-exposure prophylaxis (PrEP) for each risk group $k \in \{HM, HF, MSM\}$ in jurisdiction $j \in {J}$ at time $t$ \\ \hline
\end{tabular}
\caption{Components of state space in markov decision process (MDP)}
\label{rl_state_table}
\end{table}


Table \ref{rl_state_table} details the components of state space given in Equation \ref{eq:1}. Each of the above elements can take values from 0 to 1 on the real scale. Thus, all possible combinations of $o^{j}$ form the observation space for jurisdiction $j \in {J}$.

We represent a state $S_{t}$ at time $t$ as the set of  observations $o_{t}^{j}$ from all the jurisdictions $j \in J$, i.e., 
\begin{equation}\label{eq:2}
    S_{t} = [o_{t}^{1},o_{t}^{2},...o_{t}^{J}]
\end{equation}

\subsubsection{Action}
We formulate an action $a_{t}^{j}$ for agent $j \in J$ at time $t$ as a vector representing the three intervention strategies: testing, retention-in-care and treatment, and PrEP. 
$a_{t}^{j}$ is represented as, 
\begin{equation}\label{eq:3}
    a_{t}^{j} = [\textbf{a}_{unaware,k,t}^{j}, \textbf{a}_{ART,k,t}^{j}, \textbf{a}_{prep,k,t}^{j}; \forall k \in \{HM,HF,MSM\}]
\end{equation}

where, $\textbf{a}_{unaware,k,t}^{j}$ denotes the proportion decrease in $\mu_{u,k,j,t}$, $\textbf{a}_{ART,k,t}^{j}$ denotes the proportion increase in $\mu_{ART,k,j,t}$, and $\textbf{a}_{prep,k,t}^{j}$ denotes the proportion increase in $\mu_{prep,k,j,t}$, for each risk group $k \in \{HM, HF, MSM\}$ in jurisdiction $j \in {J}$ at time $t$. Each  of the above elements can take values from 0 to 1 on the real scale. Thus, all possible combinations of $a^{j}$ form the action space for jurisdiction $j \in {J}$.
 Note that, instead of directly formulating testing and retention-in-care using corresponding rates, we formulated as above because of its attractive mathematical properties that help efficiently constrain the number of action choices and thus improve the chance of convergence of an RL algorithm, as proved in \cite{khatami2021reinforcement}.

\subsubsection{State Transition Function}
The state transition function calculates the probabilities of a system transitioning from state $S_{t}$ to the next state $S_{t+1}$ when an action $a_{t}$ is taken. For this problem, the state transition probabilities are not available, hence we use the simulator to provide the Markov jumps from one state to another under the action taken by the agent(s).

\subsubsection{Reward}
Rewards reflect the impact of an action on the current state. Immediate rewards are computed as the negative of the total new infections, as our objective is to minimize this value over time. In addition to this, the reward function also contains a penalty term that drives interventions to be within allocated budget. The rewards $R_{a_{t}^{j}}^{j}(S_{t}^{j}, S_{t+1}^{j})$
for an agent $j \in J$ transitioning from state $S_{t}^{j}$ to the next state $S_{t+1}^{j}$ under action $a_{t}^{j}$ can be estimated as follows:
\begin{equation}\label{eq:4}
    R_{t}^{j} = -\displaystyle \sum_{j,k} i_{k,j,t} - P_{t}^{j}
\end{equation}
where, $i_{k,j,t}$ represents the new infections for risk groups $k \in \{HM, HF, MSM\}$ in jurisdiction $j$ at time $t$.
$P_{t}^{j}$ is the penalty term, which is calculated as, $P_{t}^{j} = \sum_{j} C_{t}^{j} - B_{t}$, the difference between the sum of HIV related costs $C_{t}^{j}$ for all jurisdictions and the budget $B_{t}$ for the state at time $t$. Notably, $C_{t}^{j}$ is estimated from the simulation model, using unit cost input functions from  \cite{khatami2021reinforcement} (discussed in Experimental Setup). We evaluate multiple budget constraints (discussed in Experimental Setup).

\subsection{Deep Reinforcement Learning (DRL) Solution Methodology} \label{drl_methodology}
The goal of the DRL algorithm is to solve for the optimal \textit{policy}, i.e., the optimal sequence of \textit{actions} over the period 2019 to 2030, for each jurisdiction. In the MARL framework, each agent samples an action and interacts with the environment, represented by the simulator. Although each agent acts independently, their simultaneous interactions within the same environment can influence the actions taken by all of them. Consequently, each agent receives a reward from the simulator and transitions to the next state. Repeating this process over multiple iterations, each agent learns to take better actions to eventually converge to the optimal policy. In this paper, we use the IPPO, multi-agent extension of PPO, to solve the MDP. In addition, we use two variants of multi-agent learning, CTDE and CTDE (action), to compare the performance of various multi-agent reinforcement learning frameworks. The details of these reinforcement learning algorithms and their implementations are provided in \ref{marl_algorithm}.



\section{Experimental Setup} \label{exp}

The simulation model in each jurisdiction was initialized to match HIV in the U.S. in 2018, using data from NHSS, and are discussed in \cite{tatapudi2022evaluating}. All inputs to the epidemic model, including natural disease progression, and sexual behavioral, and transmissions related parameters, and model validation are discussed in \cite{tatapudi2022evaluating}. We focused our analyses on two states California (CA) and Florida (FL), as each have multiple EHE priority counties. 
 The specific jurisdictions used in each state are detailed below.

\textbf{California}: Alameda County, Los Angeles County, Orange County CA, Riverside County, Sacramento County, San Bernardino County, San Diego County, (rest of) California

\textbf{Florida}: Broward County, Duval County, Hillsborough County, Miami-Dade County, Orange County FL, Palm Beach County, Pinellas County, (rest of) Florida

We conducted three experiments, the first to compare MARL with SARL and traditional optimization approaches, the second to compare different MARL algorithms, and the third to estimate optimal policy under varying constraints and its impact on reaching EHE goals. We discuss each experiment below.   

The first experiment compared optimal outcomes (HIV incidence over time) from MARL, A-SARL, and I-SARL, by solving for optimal policy under a fixed budget and similar objective function (Table \ref{marl_sarl_table}). MARL is the proposed method. A-SARL and I-SARL are representative of national and independent models in the literature respectively. This experiment also compared MARL with classical optimization approaches, a myopic approach and a Cross-Entropy Method (CEM). Unlike MARL, which is a sequential decision-making optimization approach, a myopic method and CEM are both static optimization approaches. While dynamic programming is more suitable for sequential decision-making they are not applicable for continuous state and action spaces. Therefore, we compare our algorithm with myopic method and CEM as they are the most suitable for problems with continuous state and action spaces.

MARL: The formulations and solution methods for MARL setup were discussed in \ref{decmdp_formulation} and \ref{drl_methodology}. In MARL, each jurisdiction within a state is an agent, evaluating an optimal policy for its jurisdiction. At each decision-making step (every
year), as each agent is evaluating its optimal policy, and all agents are simultaneously doing so, the state transitions and rewards are influenced by the actions take by all agents.

A-SARL: In the aggregated-SARL, we have one agent (CA or FL). Though the simulation itself contains the multiple interacting jurisdictions within each State (CA or FL), the markov decision process (MDP) problem (state, action, and reward in equations (\ref{eq:1}), (\ref{eq:2}), and (\ref{eq:3}), respectively), is formulated as a single agent (representing the State CA or FL) by aggregating across all jurisdictions. The single agent solves for one optimal policy for the full state.

I-SARL: In independent-SARL, the MDP is formulated by treating each jurisdiction as an agent, but the epidemic environment assumes no jurisdictional interactions, i.e., there is no jurisdictional mixing. Each agent solves for an optimal policy for its own jurisdiction, however, as there is no epidemic interactions, the actions of other agents do not impact its state transitions. To evaluate the impact of overlooking jurisdictional interactions, individually trained models of each jurisdiction are applied on the full simulation that includes jurisdictional mixing.

Static myopic optimization \cite{powell2007approximate}: This approach solves for the optimal policy incrementally for each year starting from the first year, using optimal decisions from the previous year as input to the next year. 

Cross-entropy method (CEM) \cite{rubinstein2004cross}: CEM is a stochastic optimization approach that formulates the full policy (all 12 years) as a static decision variable. An overview of both methods are presented in \ref{optimization_formulation}. 



\begin{table}[t]
\centering
\begin{tabular}{|M{1.5cm}|M{3cm}|M{6cm}|}
 \hline
 \textbf{Name} & \textbf{Objective} & \textbf{Cost Constraint} \\ 
 \hline
 MARL1 (IPPO) & Minimize total infection of the state & Sum of total cost of the state $\leq$ Initial budget of the state\\
 \hline
 A-SARL  & Minimize total infection of the state & Sum of total cost of the state $\leq$ Initial budget of the state\\
 \hline
 I-SARL  & Minimize total infection of each jurisdiction (assuming no mixing) & Sum of total cost for a jurisdiction $\leq$ Initial budget of the jurisdiction (the distribution of the budget for each jurisdiction was taken from the MARL output)  \\ \hline
\end{tabular}
\caption{Overview of MARL, A-SARL, and I-SARL}
\label{marl_sarl_table}
\end{table}


The second experiment compared MARL algorithms from Section \ref{marl_algorithm} (IPPO, CTDE and CTDE (action)), using incidence as the outcome metric. Table \ref{ippo_ctde_table} details the setup of the experiment. The three methods differ in the input given to the critic network of the DRL agent while keeping everything else constant. The best performing algorithm was then used for experiment three.


\begin{table}[t]
\centering
\begin{threeparttable}
\begin{tabular}{|M{1.5cm}|M{2.5cm}|M{3.5cm}|M{3cm}|}
\hline
 \textbf{Name} & \textbf{Objective} & \textbf{Cost Constraint}  & \textbf{Critic Input}\\
 \hline
 MARL1 (IPPO) & Minimize total infection of the state & Sum of total cost of the state $\leq$ Total budget of the state* &  Observation of each agent\\
 \hline
 MARL2 (CTDE) & Minimize total infection of the state & Sum of total cost of the state $\leq$ Total budget of the state &  Observation of all the agent\\
 \hline
 MARL3 (CTDE (action)) & Minimize total infection of the state & Sum of total cost of the state $\leq$ Total budget of the state &  Observation of all the agents and the action of the remaining agents \\ \hline
\end{tabular}
\begin{tablenotes}
    \item[*] Action space and budget kept at baseline for all three algorithms. Baseline action uses intervention scale-up following past trends whereas baseline budget is minimum budget required to maintain the status-quo intervention levels.
\end{tablenotes}
\end{threeparttable}
\caption{Comparison of different MARL algorithms}
\label{ippo_ctde_table}
\end{table}


The third experiment evaluated the impact of different intervention constraints on HIV incidence (outcome of interest). The previous two experiments used a 'baseline' action space, where the intervention scale-up was set to follow past trends, had no penalty if the EHE goals were not achieved, and used a 'baseline budget', which was set as the minimum cost needed to maintain status-quo intervention levels. We also implemented he baseline scenario to all 96 jurisdictions to demonstrate the scalability of the model.

In this third experiment, to analyze the influence of intervention scale-up, incidence penalty, and budget constraints, we incrementally modified each of these constraints, as follows.  Scenario 2 and above in Table \ref{scenario_table} doubled the action space (see Action space constraints below). Scenario 3 and above assign a penalty if the EHE goal of 90\% incidence reduction by 2030 in not reached (see Reward function penalty constraints below). 
Scenarios 4 and above evaluate different increments in budget constraints (see Budget constraints below). We restricted this third experiment to only two states as preliminary analyses to understand level of scale-up needed to reach the EHE goals.
 
Action space constraints (for intervention scale-up): We evaluated two action spaces, i.e., options for intervention scale-up. One follows past trends (baseline), and another, an optimistic scale-up . Specifically, for the period 2015 to 2019, there was a 1.6 \% decrease in unaware proportion (14.9 \% in 2015 to 13.3 \% in 2019) \cite{hivunawarenum} and a 4.8 \% increase in VLS proportion (59.8 \% in 2015 to 64.6 \% in 2019) \cite{hivvlsnum}. Thus, for baseline unaware proportion (see \ref{eq:2}), we set the  feasible range of action choices as 0 to 0.5 \%  decrease per year, and for ART proportion (see \ref{eq:2}), we set the feasible range of action choices as 0 to 4 \% increase per year. Similarly, the percentage of people with PrEP indicators that have been prescribed PrEP increased from 13.2 \% in 2017 to 30.1 \% in 2021, approximately 4\% increase per year \cite{hivprepnum}.Therefore, for baseline proportion on PrEP (see \ref{eq:2}), we set the feasible range of action choices as 0-4 \% per year. In the aggressive scenario, we assumed scale-up can occur at twice the past trend. Specifically we increased the maximum action range to 1\% for decrease in unaware proportion, 8\% for increase in ART proportions, and 8\% for increase in PrEP proportions.

Reward function penalty constraints: To add a penalty if the EHE incidence reduction goals were not met, we modified the reward function as in Equation \ref{eq:14}.

\begin{equation}\label{eq:14}
    R_{t}^{j} = -\displaystyle \sum_{j,k} i_{k,j,t} - P_{t}^{j}
\end{equation}

where, $P_{j}^{t}$ is the penalty associated with budget and EHE goals of infection. The penalty is given by Equation \ref{eq:15}, which is comprised of three components. The first component (Equation \ref{eq:16}), applies a penalty if budget is exceeded. The second and third components add penalties if an EHE jurisdiction dose not meet the 75\% and 90\% reduction in incidence by 2025 and 2030, respectively (equivalent to $t=7$ and $t=12$) ( Equations \ref{eq:17} and \ref{eq:18}).

\begin{equation}\label{eq:15}
    P_{t}^{j} = P_{t}^{'j} + P_{t}^{''j} + P_{t}^{'''j}
\end{equation}

where,

\begin{equation}\label{eq:16}
P_{t}^{'j} =
\begin{cases}
    \sum_{j} C_{t}^{j} - B_{t}, & \text{if } \sum_{j} C_{t}^{j} > B_{t} \\
    0,                                          & \text{else}
\end{cases}
\end{equation}

At $t=7$,
\begin{equation}\label{eq:17}
P_{t}^{''j} =
\begin{cases}
    (\sum_{k} i_{k,j,t} - 0.25 I_{0}^{j})\times1000, & \text{if } \sum_{k} i_{k,j,t} > 0.25 I_{0}^{j} \\
    0,                                          & \text{else}
\end{cases}
\end{equation}

At $t=12$, 
\begin{equation}\label{eq:18}
P_{t}^{'''j} =
\begin{cases}
    (\sum_{k} i_{k,j,t} - 0.1 I_{0}^{j})\times100, & \text{if } \sum_{k} i_{k,j,t} > 0.1 I_{0}^{j} \\
    0,                                          & \text{else}
\end{cases}
\end{equation}

\begin{table}
\centering
\begin{tabular}{|M{1.4cm}|M{1.1cm}|M{1.6cm}|M{1.9cm}|M{1.5cm}|M{1.2cm}|M{1.8cm}|}

 \hline
 \textbf{Scenario} & \textbf{Name} & \textbf{Objective} & \textbf{Cost Constraint} & \textbf{Action Space
 } & \textbf{EHE Penalty} & \textbf{Budget}\\
 \hline
 Scenario 1 & MARL1 & Minimize total infection & Sum of total cost $\leq$ Total budget & Baseline & No & Baseline \\
 \hline
 Scenario 2 & MARL4 & Minimize total infection & Sum of total cost $\leq$ Total budget & Optimistic & No & Baseline \\
 \hline
 Scenario 3 & MARL5 & Minimize total infection & Sum of total cost $\leq$ Total budget & Optimistic & Yes & Baseline \\
 \hline
Scenario 4 & MARL6 & Minimize total infection & Sum of total cost $\leq$ Total budget & Optimistic & Yes & 2$\times$Baseline (CA), 5$\times$Baseline (FL) \\
 \hline
Scenario 5 & MARL7 & Minimize total infection & Sum of total cost $\leq$ Total budget & Optimistic & Yes & 3$\times$Baseline (CA), 10$\times$Baseline (FL) \\
 \hline
Scenario 6 & MARL8 & Minimize total infection & Sum of total cost $\leq$ Total budget & Optimistic & Yes & 5$\times$Baseline (CA), 15$\times$Baseline (FL) \\
 \hline
Scenario 7 & MARL9 & Minimize total infection & Sum of total cost $\leq$ Total budget & Optimistic & Yes & 7$\times$Baseline (CA), 18$\times$Baseline (FL) \\ \hline
\end{tabular}
\caption{Comparison of various scenarios with action and budget scale-up}
\label{scenario_table}
\end{table}

Budget constraints: We evaluated five budget constraints. The 'Baseline budget' was set as the cost needed to maintain the status-quo interventions, which was estimated as an outcome of the simulation in 2019 using unit costs discussed in the next sub-section. The estimated baseline budget for Florida and California was \$ $4\times10^{8}$ and \$ $15\times10^{8}$, respectively. We then evaluated, through trial and error, the minimum increment in budget needed to reach the EHE goal, which was 7 times and 18 times the baseline budget for California and Florida, respectively. To evaluate the changes at different-levels of budget scale-up, we evaluated   budget constraints at increments of 2, 3, 5 and 7 times the baseline budget for California, and 5, 10, 15 and 18 times the baseline budget for Florida.

Unit costs: The total HIV related cost is comprised of three components: testing, retention-in-care, and prevention, corresponding to the three interventions (discussed in the action space). HIV testing costs are estimated in the simulation model based on the number tested as per the testing intervention (part of \textit{action} see (\ref{eq:3})) and the unit cost of testing. Unit cost estimation are discussed in detail in \cite{khatami2021reinforcement}. Briefly, unit costs are determined by considering that testing can occur in clinical or non-clinical settings, and each setting has fixed and variable costs. Fixed costs include infrastructure and staff. Variable costs include the cost per person tested, including cost of the test and cost of outreach intervention programs. The retention-in-care cost is estimated in the simulation model based on the number of persons retained in care and treatment as per the retention-in-care intervention  (part of \textit{action} see (\ref{eq:3}))  and unit cost per person. Unit costs include cost of treatment and outreach programs. The unit cost estimation and sources for an outreach program are discussed in detail in \cite{khatami2021reinforcement}. The prevention (PrEP) costs are estimated in the simulation model based on the number of persons on PrEP as per the intervention taken (part of \textit{action} see (\ref{eq:3})) and unit cost per person. We used unit cost of PrEP from  \cite{sansom2021optimal}, which includes an annual cost (for care monitoring) and drug cost.

\subsection{Hyperparameter Selection} \label{hyperparam_select}

Hyperparameters were selected through literature-based values \cite{schulman2017proximal, raffin2021stablebaselines3} and grid-based search (10 combinations were experimented).  The combination that had good stability and convergence on average reward were chosen (see Table \ref{hyperparameter}; see Appendix \ref{hyperparamters_tune} for further details).

\begin{table}[H]
\centering
\begin{tabular}{|M{3cm}|M{4cm}|M{3cm}|}
\hline
\textbf{Hyperparameter}  & \textbf{Grid Search Values \cite{schulman2017proximal, raffin2021stablebaselines3} } & \textbf{Selected Values}       \\ 
\hline
Discount factor          & [0.95, 0.99]  &0.99 \\ \hline 
Learning rate      & [1e-4, 3e-4, 1e-3]  &0.0003 \\ \hline
Clip parameter           & [0.1, 0.2]  &0.2 \\ \hline
K epochs                 &  20 &20 \\ \hline
Optimizer                & Adam  & Adam \\ \hline
Entropy Coefficient      & 0.01  & 0.01 \\ \hline
\end{tabular}
\caption{List of hyperparameters used in MARL training}
\label{hyperparameter}
\end{table}

\section{Results}\label{results}

\subsection{Baseline Comparison}

\begin{figure}[t]
  \centering
  \begin{subfigure}{0.45\textwidth}
    \includegraphics[width=1\textwidth]{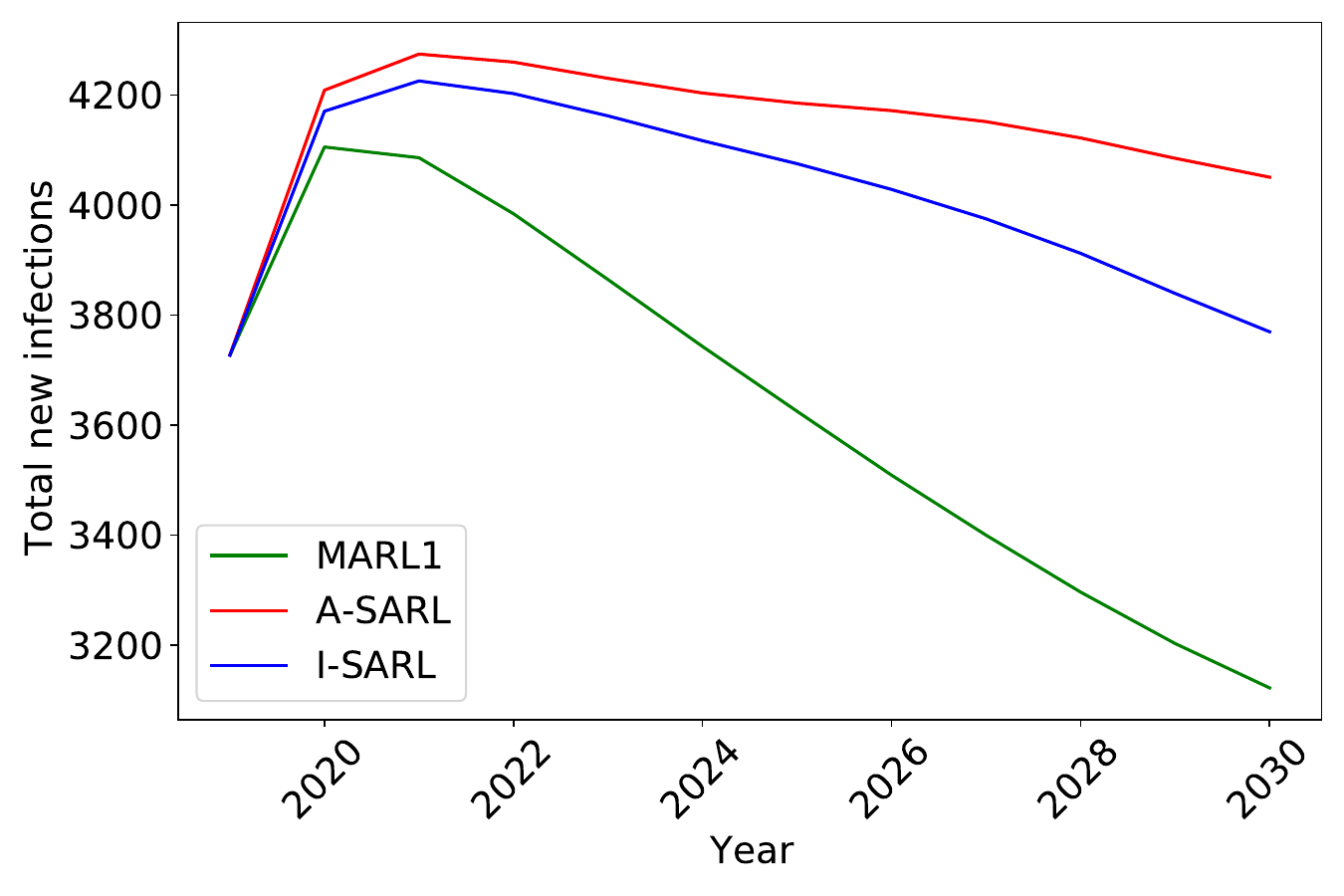}
      \caption[size=12]{}
      \label{fig:inf_marl_sarl_CA}
  \end{subfigure}
  \begin{subfigure}{0.45\textwidth}
    \includegraphics[width=1\textwidth]{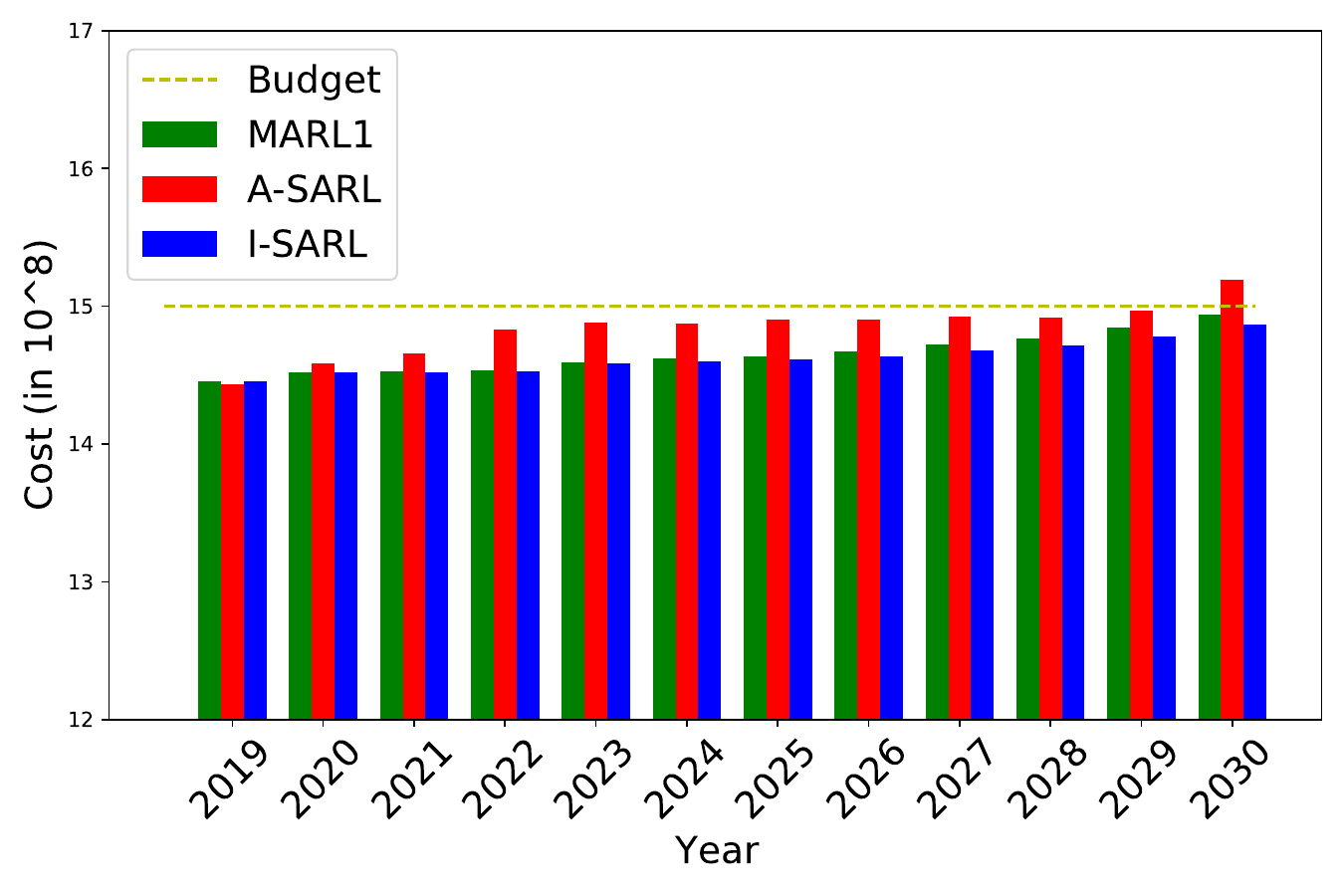}
      \caption[Image 2]{}
      \label{fig:total_cost_marl_sarl_CA}
  \end{subfigure}
  \caption[Images]{(a) HIV incidence and (b) intervention cost under MARL, A-SARL and I-SARL in California.}
  \label{fig:total_inf_ca}
\end{figure}

\begin{figure}[t]
  \centering
  \begin{subfigure}{0.45\textwidth}
    \includegraphics[width=1\textwidth]{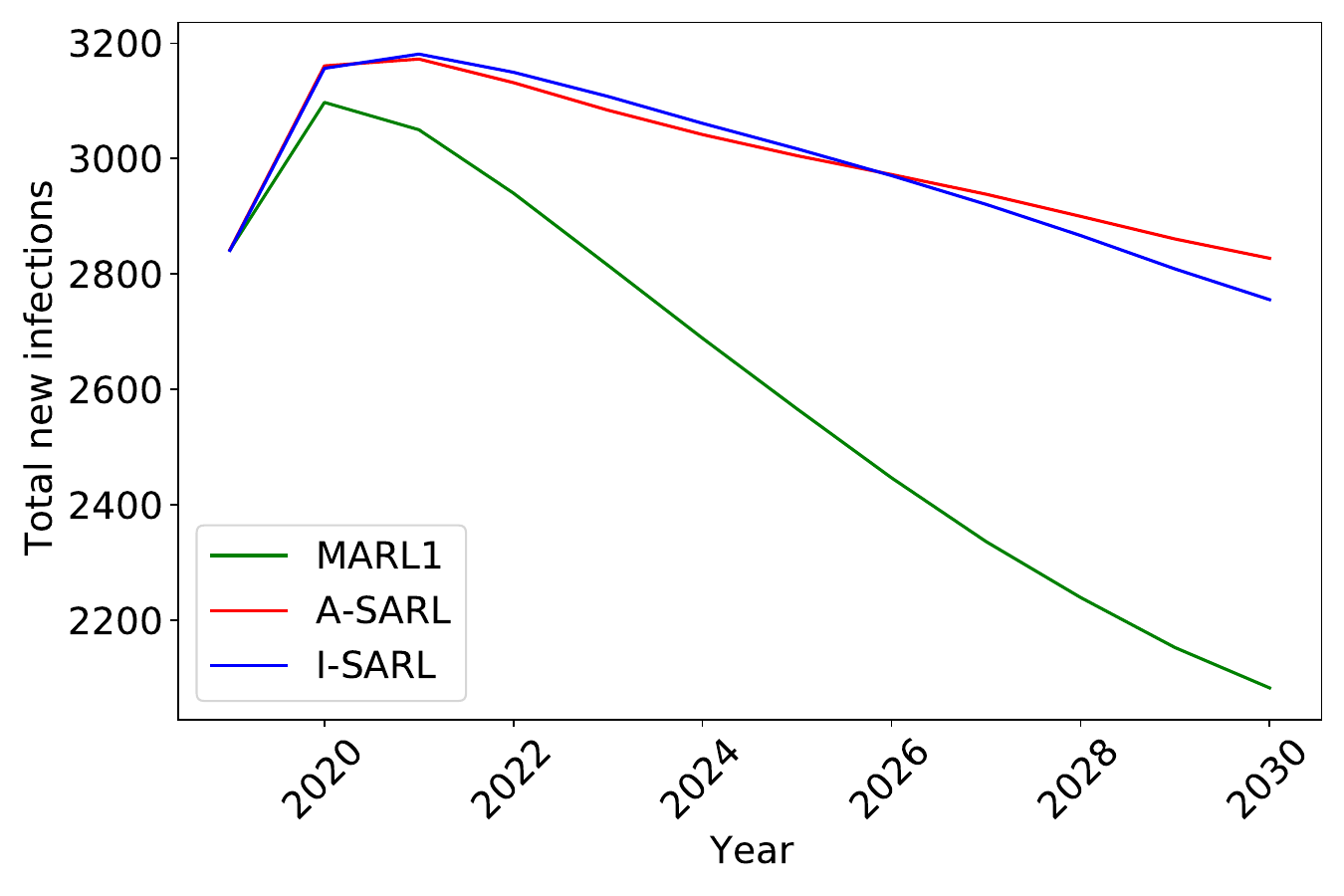}
      \caption[Image 1]{}
      \label{fig:inf_marl_sarl_FL}
  \end{subfigure}
  \begin{subfigure}{0.45\textwidth}
    \includegraphics[width=1\textwidth]{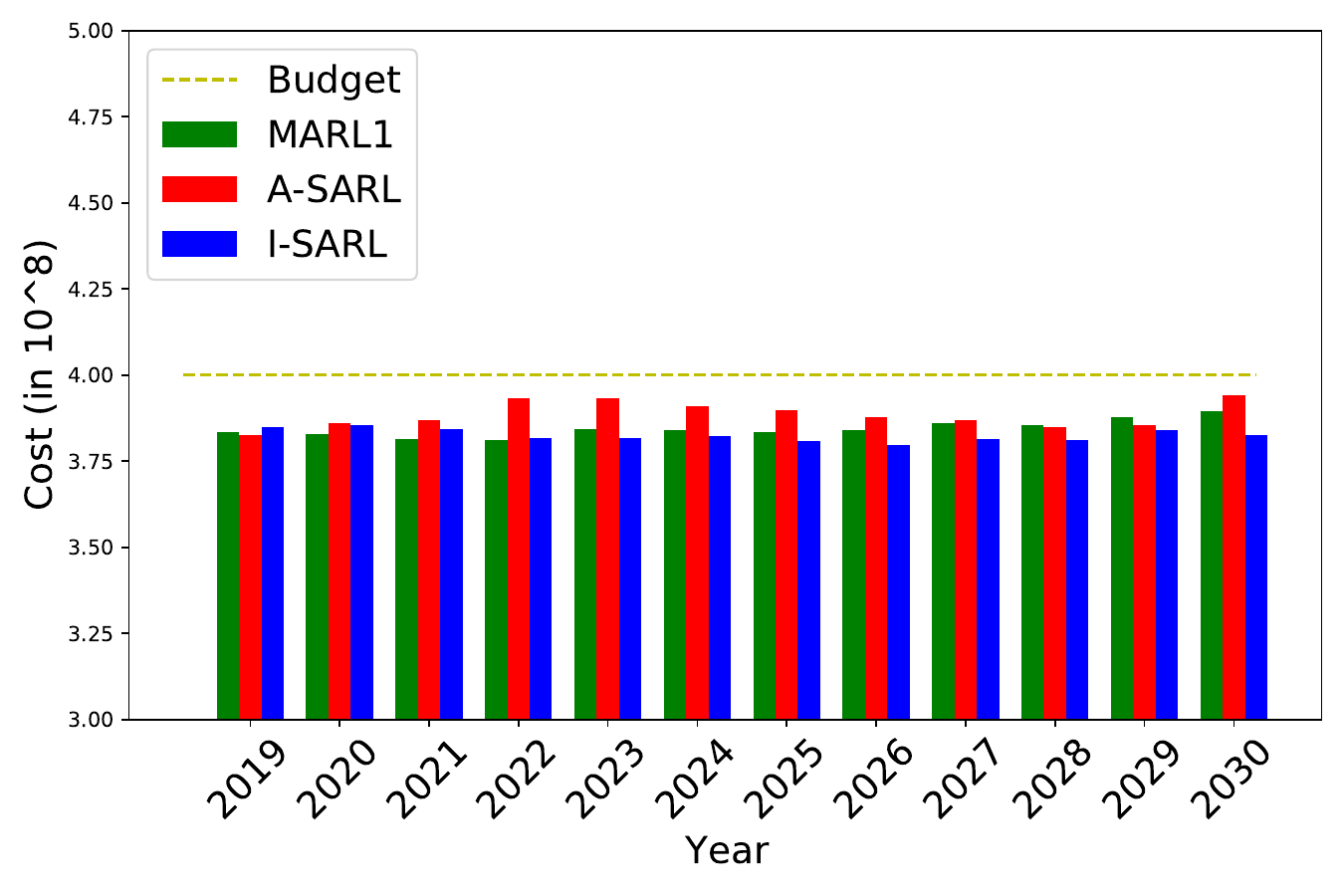}
      \caption[Image 2]{}
      \label{fig:total_cost_marl_sarl_FL}
  \end{subfigure}
  \caption[Images]{(a) HIV incidence and (b) intervention cost under MARL, A-SARL and I-SARL in Florida.}
  \label{fig:total_inf_fl}
\end{figure}

Comparing incidence and costs outcomes in optimal policies from MARL, A-SARL, and I-SARL, as expected, MARL significantly outperforms the other two in both California and Florida (Figures \ref{fig:total_inf_ca} and \ref{fig:total_inf_fl}). The pattern is consistent even when comparing incidence in each jurisdiction within these states (see Appendix 
\ref{fig:incidence_florida}). As expected, results from A-SARL suggest that intervention choices will be biased towards larger jurisdictions. If larger jurisdictions have higher prevalence (as was the case here), it will result in scale-up in interventions for all jurisdictions, thus incurring costs without additional impact in lower prevalence jurisdictions. If large jurisdictions have low prevalence, we may expect jurisdictions with high prevalence to not receive the necessary intervention scale-up. Results from I-SARL suggest that ignoring jurisdictional mixing can lead to inefficient intervention choices.


\begin{figure}[t]
  \centering
  \begin{subfigure}{0.45\textwidth}
    \includegraphics[width=1\textwidth]{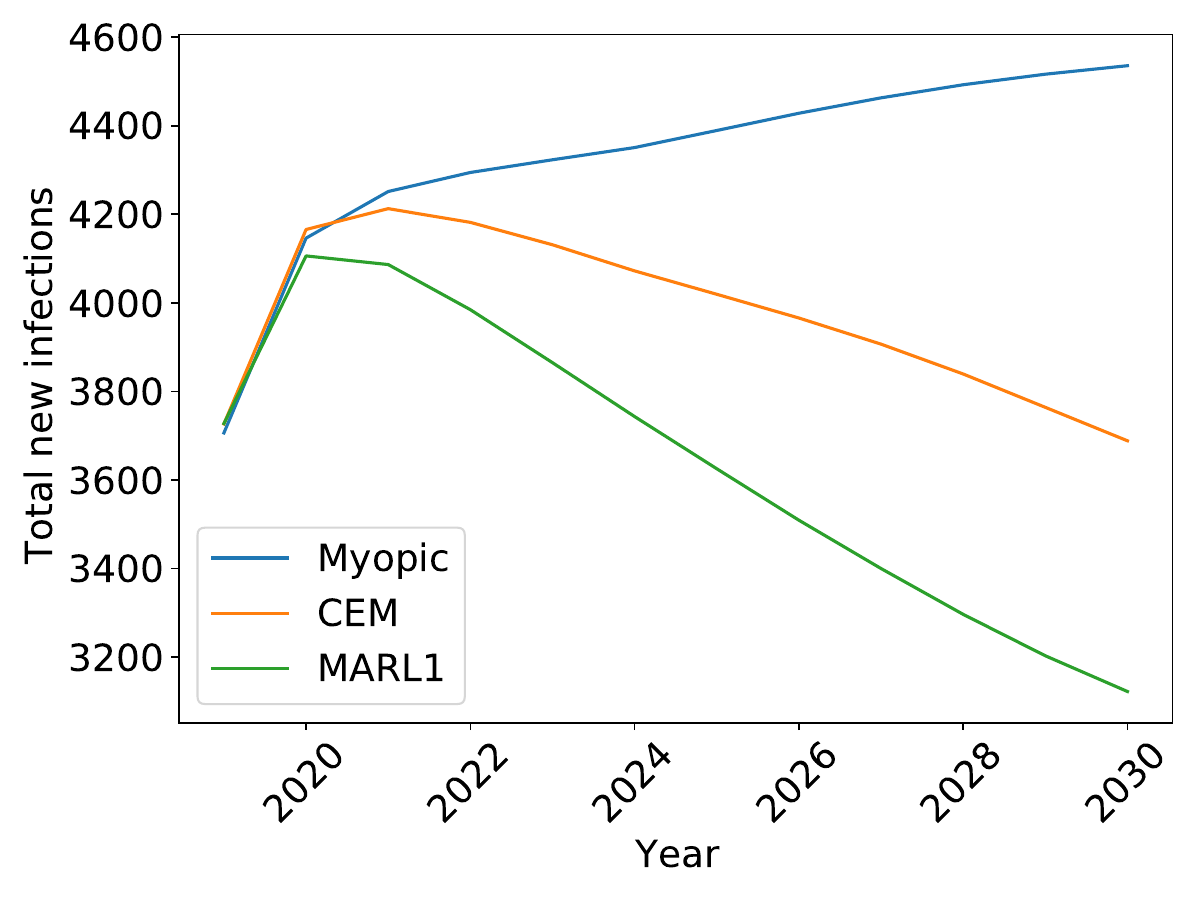}
      \caption[size=12]{}
      \label{fig:inf_myopic_cem_marl_CA}
  \end{subfigure}
  \begin{subfigure}{0.45\textwidth}
    \includegraphics[width=1\textwidth]{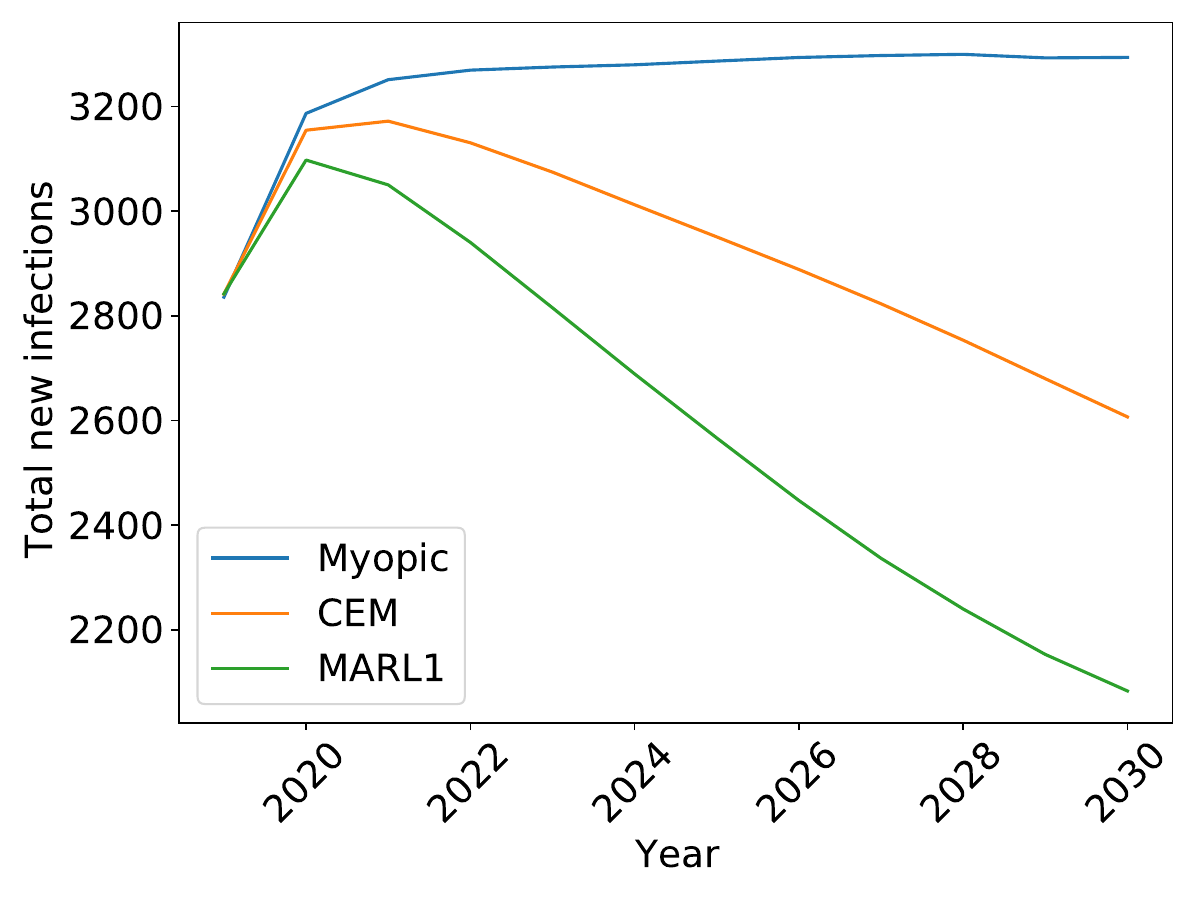}
      \caption[Image 2]{}
      \label{fig:inf_myopic_cem_marl_fl}
  \end{subfigure}
  \caption[Images]{HIV incidence under MARL, Myopic, and CEM policies in (a) California and (b) Florida.}
  \label{fig:total_inf_myopic_cem_marl_ca_fl}
\end{figure}

\begin{figure}[htbp]
  \centering
  \begin{subfigure}{0.45\textwidth}
    \includegraphics[width=1\textwidth]{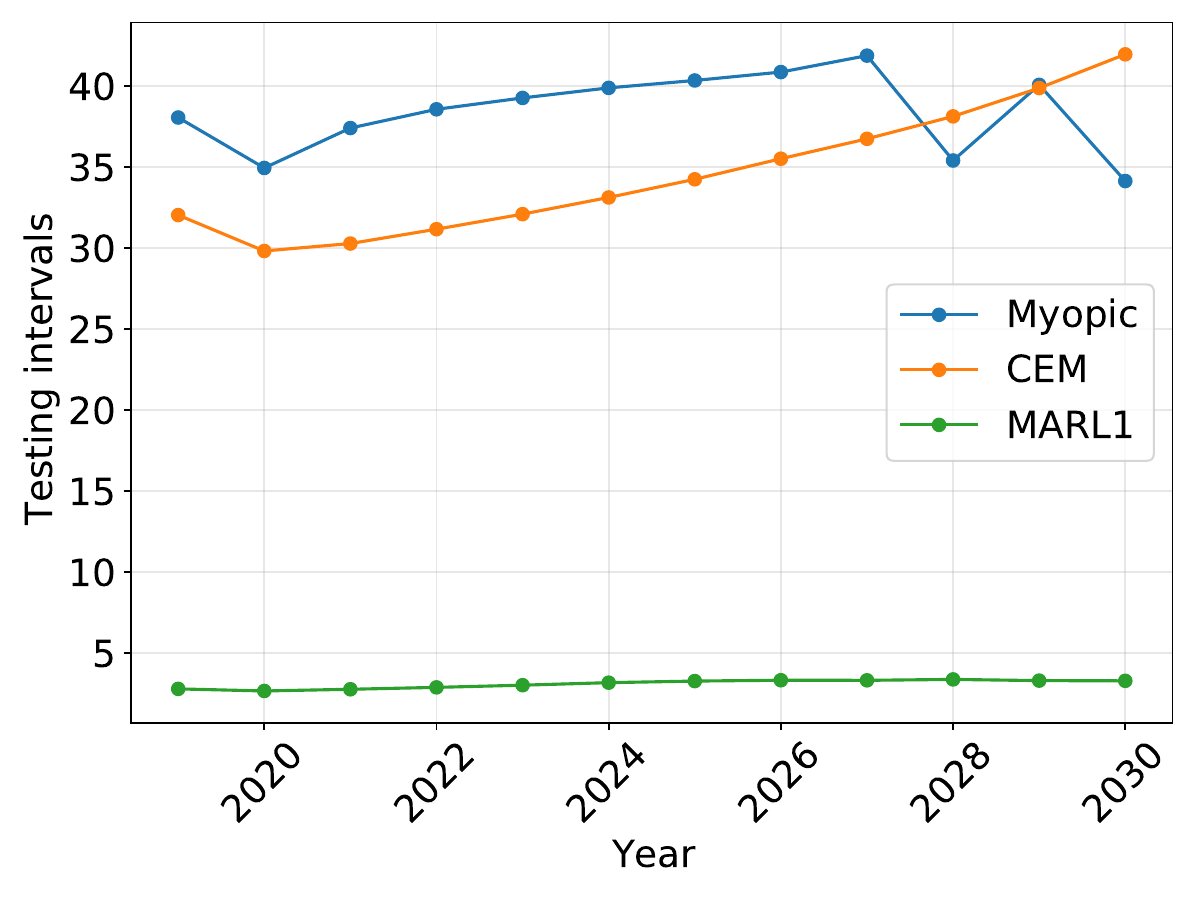}
      \caption[size=12]{}
      \label{fig:testing_myopic_cem_marl_broward}
  \end{subfigure}
  \begin{subfigure}{0.45\textwidth}
    \includegraphics[width=1\textwidth]{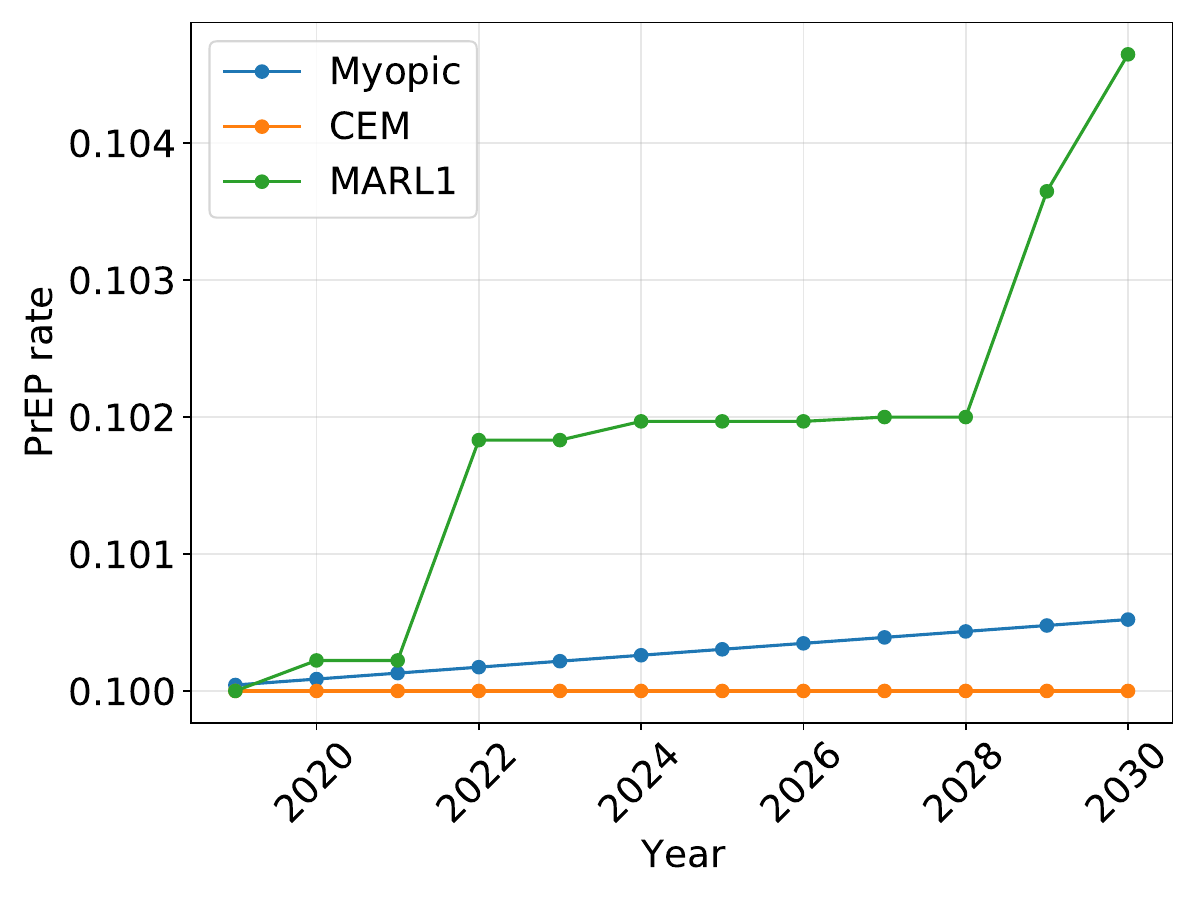}
      \caption[Image 2]{}
      \label{fig:prep_myopic_cem_marl_broward}
  \end{subfigure}
  \caption[Images]{(a) Testing interval (in years) and (b) PrEP rate for MSM under MARL, Myopic, and CEM in Broward County, Florida.}
  \label{fig:testing_prep_broward}
\end{figure}

\begin{figure}[htbp]
  \centering
  \begin{subfigure}{0.45\textwidth}
    \includegraphics[width=1\textwidth]{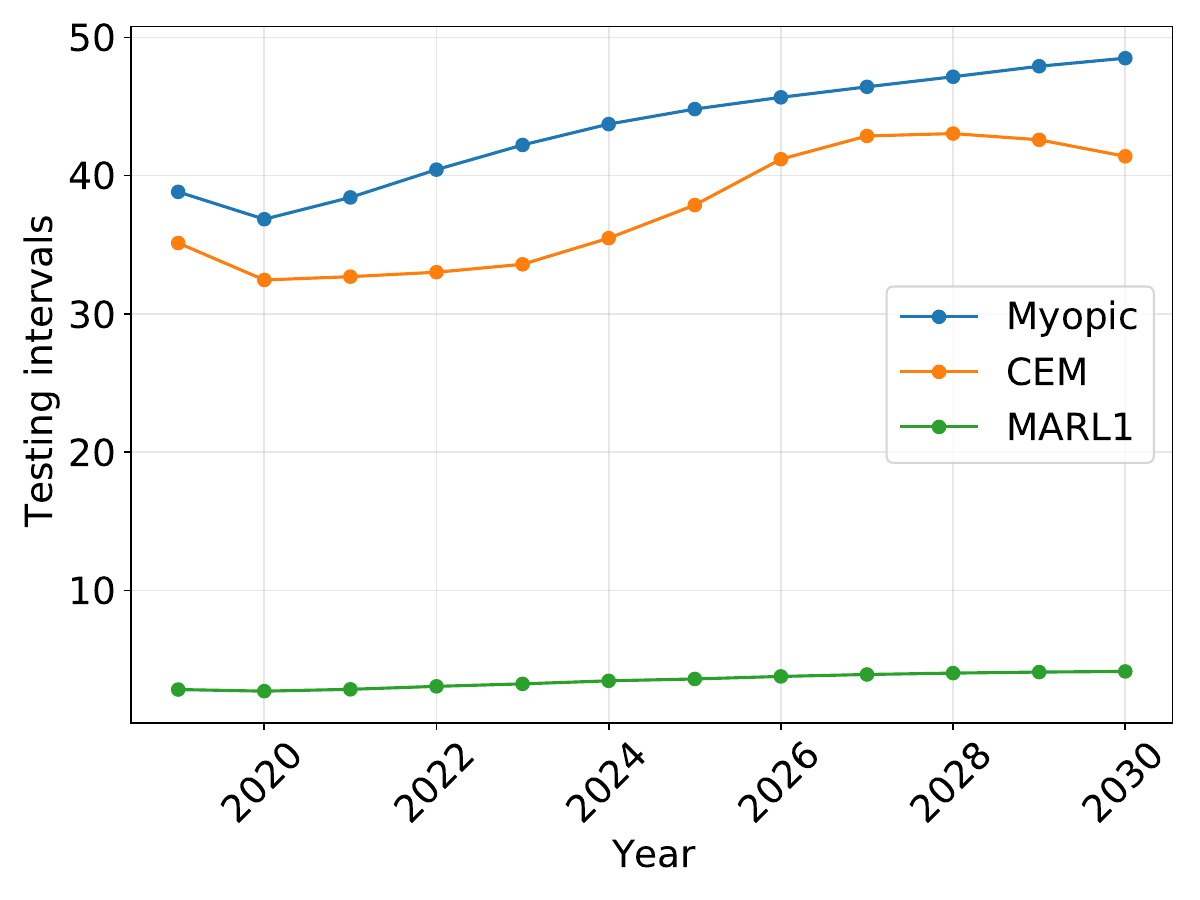}
      \caption[size=12]{}
      \label{fig:testing_myopic_cem_marl_restfl}
  \end{subfigure}
  \begin{subfigure}{0.45\textwidth}
    \includegraphics[width=1\textwidth]{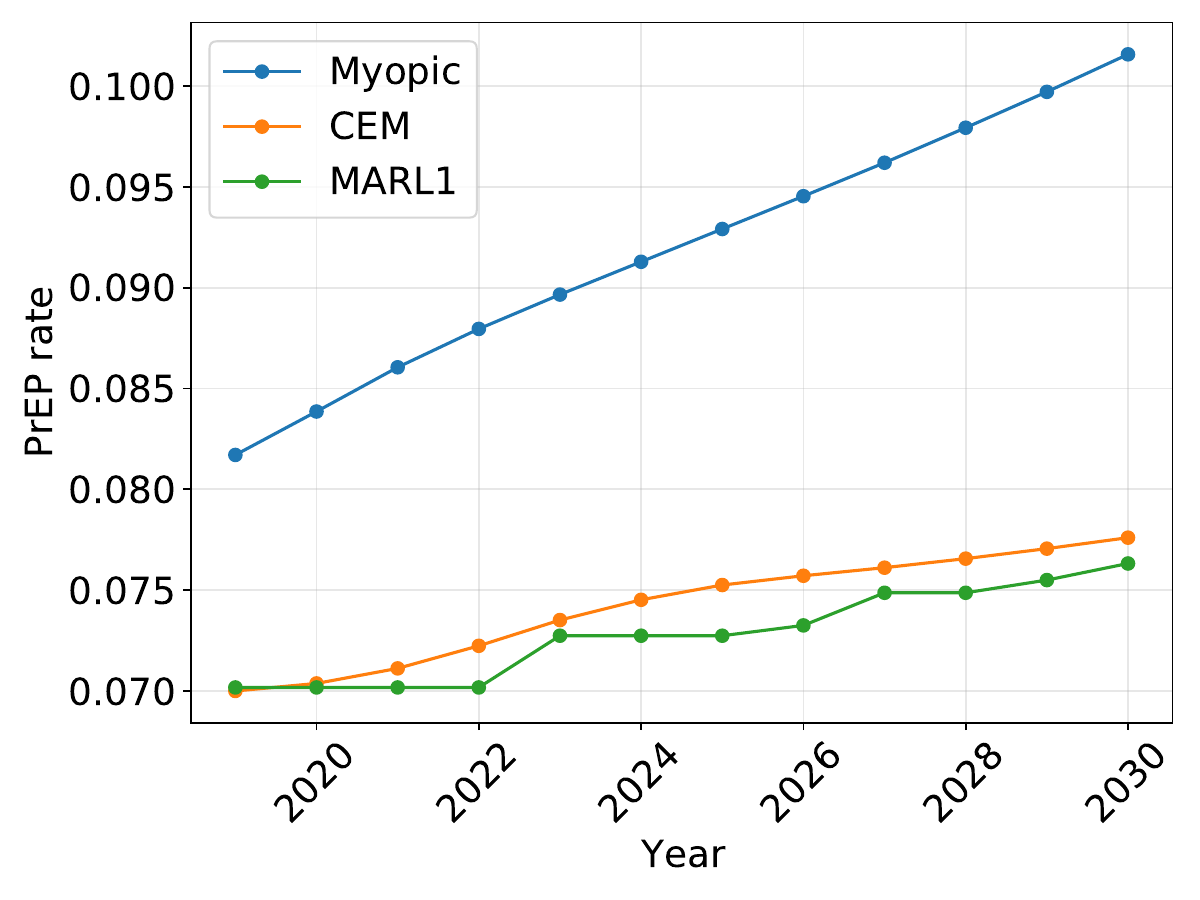}
      \caption[Image 2]{}
      \label{fig:prep_myopic_cem_marl_restfl}
  \end{subfigure}
  \caption[Images]{(a) Testing interval (in years) and (b) PrEP rate for MSM under MARL, Myopic, and CEM in Rest-of-Florida.}
  \label{fig:testing_prep_restfl}
\end{figure}

Compared to outcomes from myopic approach and CEM optimization methods , MARL had significantly lower HIV incidence (Figure \ref{fig:total_inf_myopic_cem_marl_ca_fl}). 
Resulting policies (Figure \ref{fig:testing_prep_broward} and \ref{fig:testing_prep_restfl}) indicate that the myopic approach likely focused on immediate gains, as expected. It favored lower testing even in Broward county, a high HIV burden county, for higher PrEP in rest-of-Florida that has low HIV burden (in addition to PrEP in Broward at similar levels of MARL). This is expected from myopic policy because, while the impact of PrEP occurs in the same time step, the full impact of testing is realized at a future time-step, after receiving treatment and reaching viral suppression. While PrEP was similar in both MARL and CEM,  unlike MARL, CEM favored less frequent testing for Broward and Rest-of-Florida (Figure \ref{fig:testing_prep_broward} and \ref{fig:testing_prep_restfl}) . Though CEM considers long-term gains, it likely converged to a local optima. We can expect convergence issues with CEM given the significantly large state and action space to accommodate all years, and transmission dynamics.  We can expect that the dynamics resulting from delayed effect of testing can cause more issues with convergence than PrEP.

\subsection{Comparison of MARL Frameworks}

\begin{figure}[H]
  \centering
  \begin{subfigure}{0.45\textwidth}
    \includegraphics[width=1\textwidth]{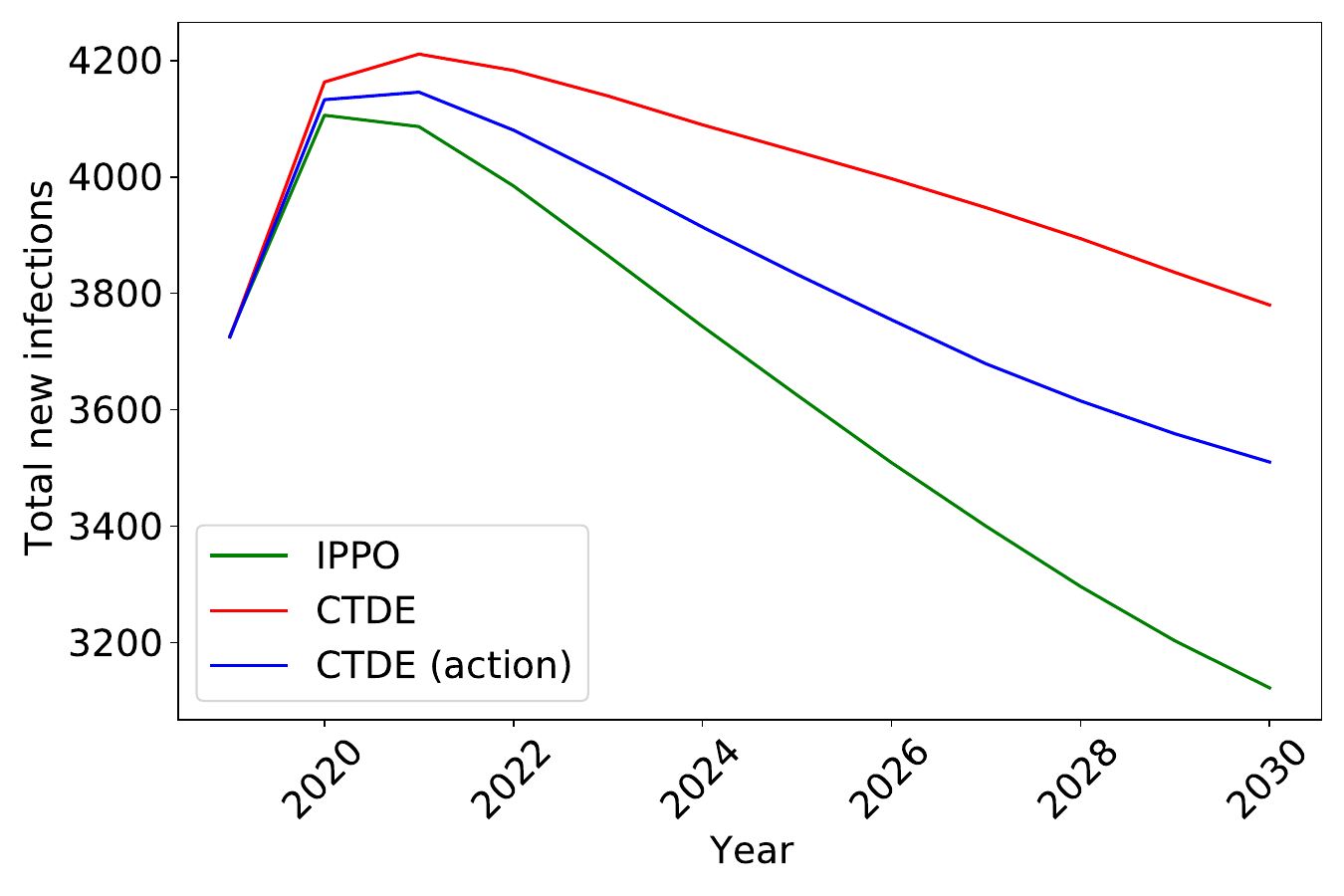}
      \caption[Image 1]{}
      \label{fig:inf_ippo_ctde_CA}
  \end{subfigure}
  \begin{subfigure}{0.45\textwidth}
    \includegraphics[width=1\textwidth]{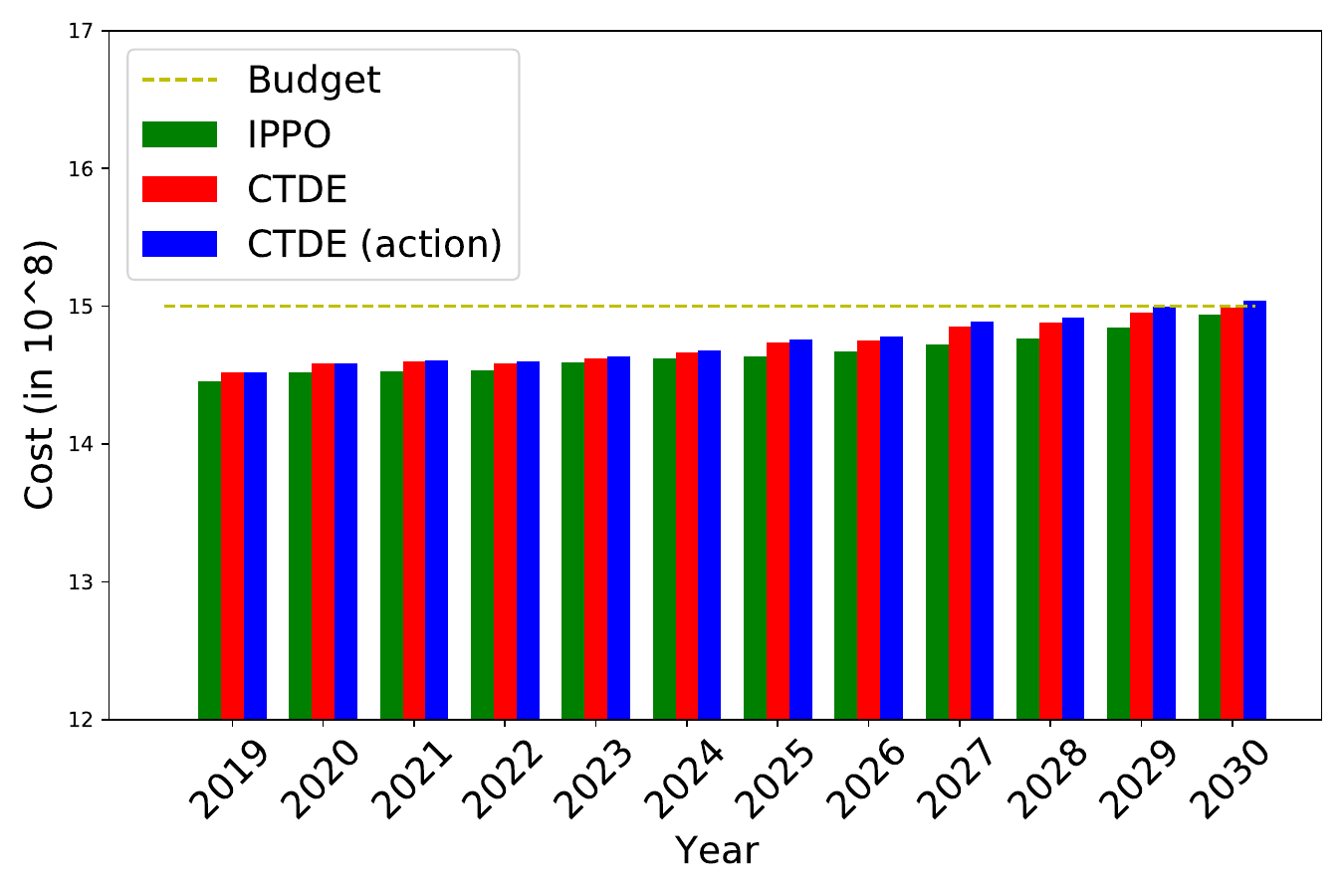}
      \caption[Image 2]{}
      \label{fig:total_cost_ippo_ctde_CA}
  \end{subfigure}
  \caption[Images]{(a) HIV incidence and (b) intervention cost under MARL frameworks: IPPO, CTDE, and CTDE (action) in California.}
  \label{fig:marl_comparison_ca}
\end{figure}

\begin{figure}[H]
  \centering
  \begin{subfigure}{0.45\textwidth}
    \includegraphics[width=1\textwidth]{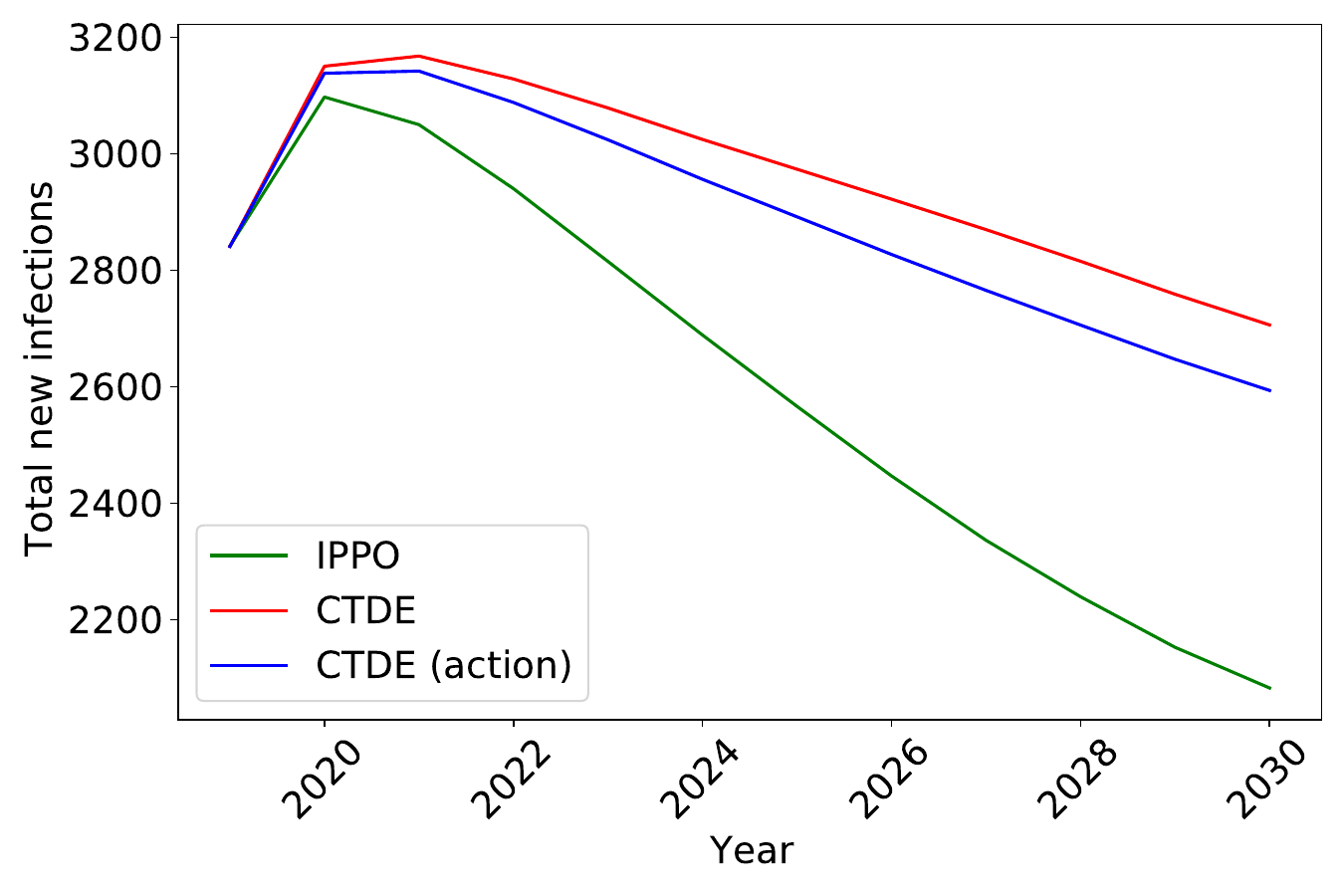}
      \caption[Image 1]{}
      \label{fig:inf_ippo_ctde_FL}
  \end{subfigure}
  \begin{subfigure}{0.45\textwidth}
    \includegraphics[width=1\textwidth]{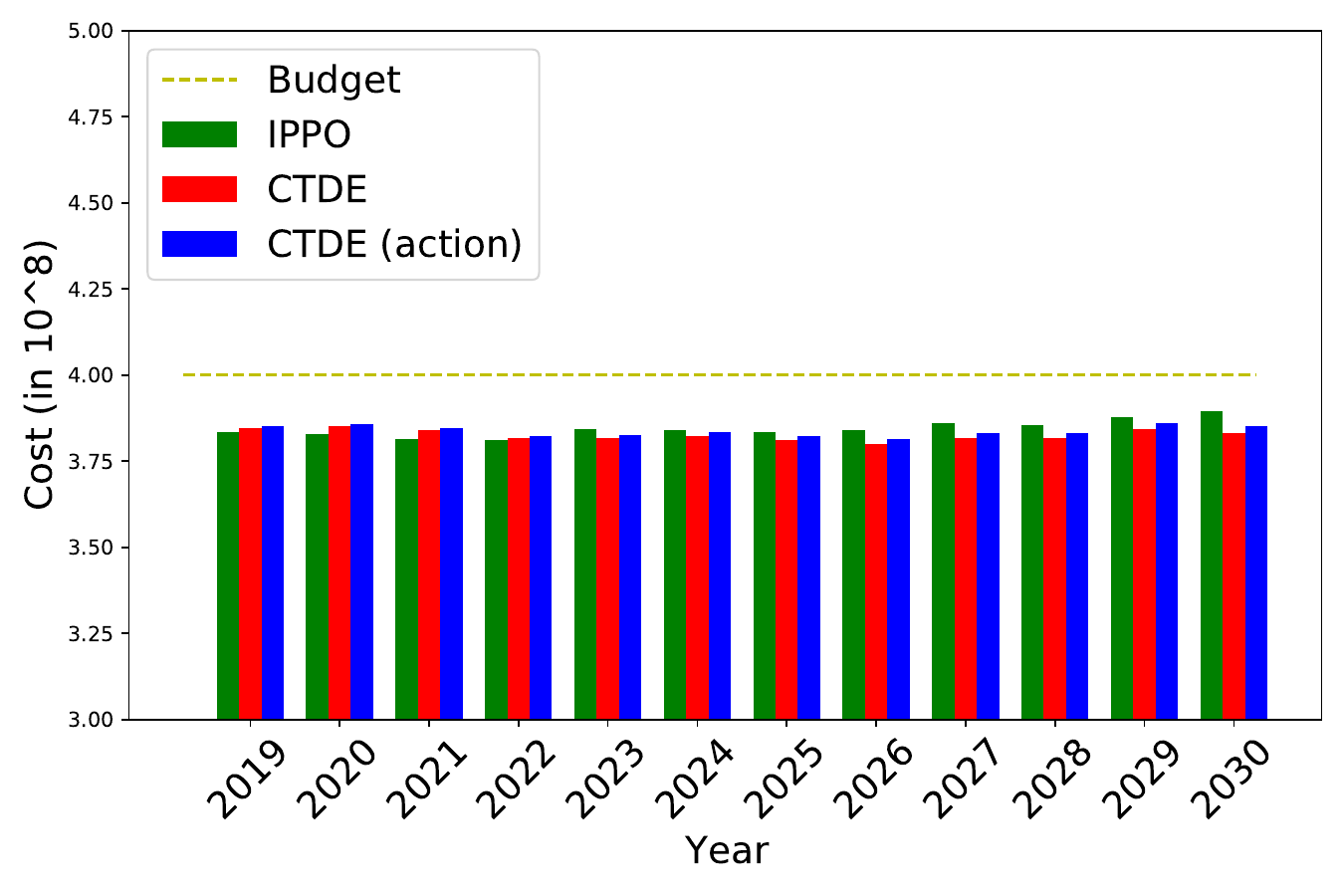}
      \caption[Image 2]{}
      \label{fig:total_cost_ippo_ctde_FL}
  \end{subfigure}
  \caption[Images]{(a) HIV incidence and (b) intervention cost under MARL frameworks: IPPO, CTDE, and CTDE (action) in Florida.}
  \label{fig:marl_comparison_fl}
\end{figure}

Comparing outcomes across MARL algorithms (IPPO, CTDE, and CTDE (action)) (Figure \ref{fig:marl_comparison_ca}a and \ref{fig:marl_comparison_fl}a), IPPO performs the best. To recollect, the critic in IPPO uses only its state whereas the critic in CTDE uses the state of every agent, and thus, the input layer of the critic network in CTDE is $J\times$ that of the critic network in IPPO. It is likely that because of constant jurisdictional mixing matrix, the effects of jurisdictional interactions are integrally learned into the epidemic dynamics through the critic network, and the significantly larger dimension of the critic in CTDE affects the stability in training. 


 \subsection{Comparison of Budget Scenarios}
 
\begin{figure}[h]
\centering  
\includegraphics[width=0.80\linewidth]{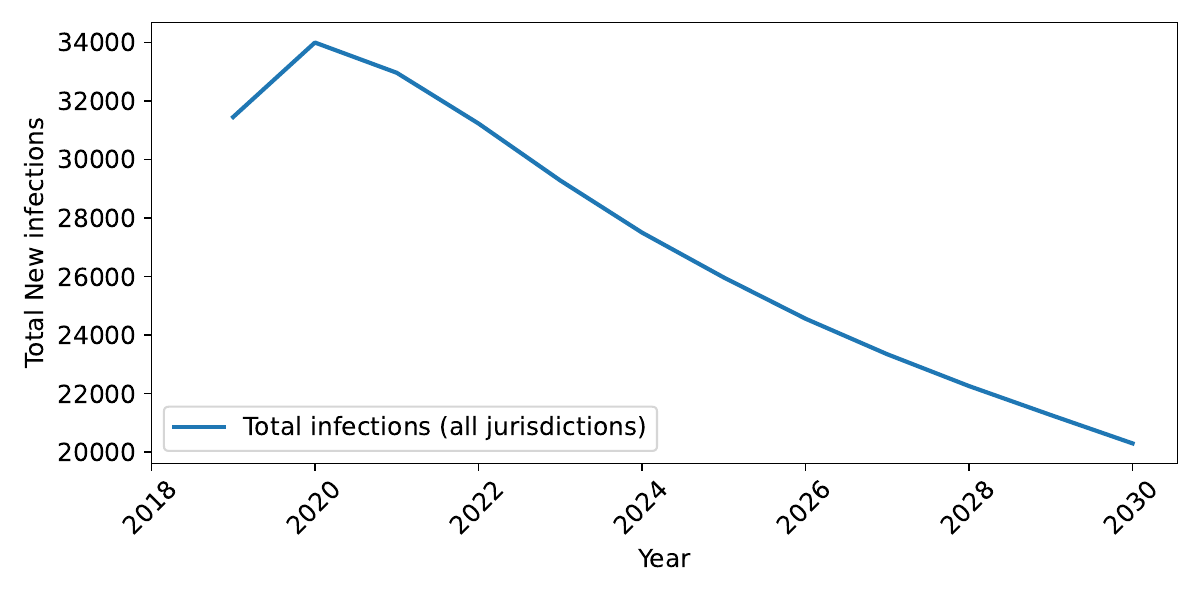} 
\caption{HIV incidence across 96 jurisdictions under IPPO baseline budget scenario.}
\label{fig:inf_96_jur}
\end{figure}

\begin{figure}
  \centering
  \begin{subfigure}{0.45\textwidth}
    \includegraphics[width=1\textwidth]{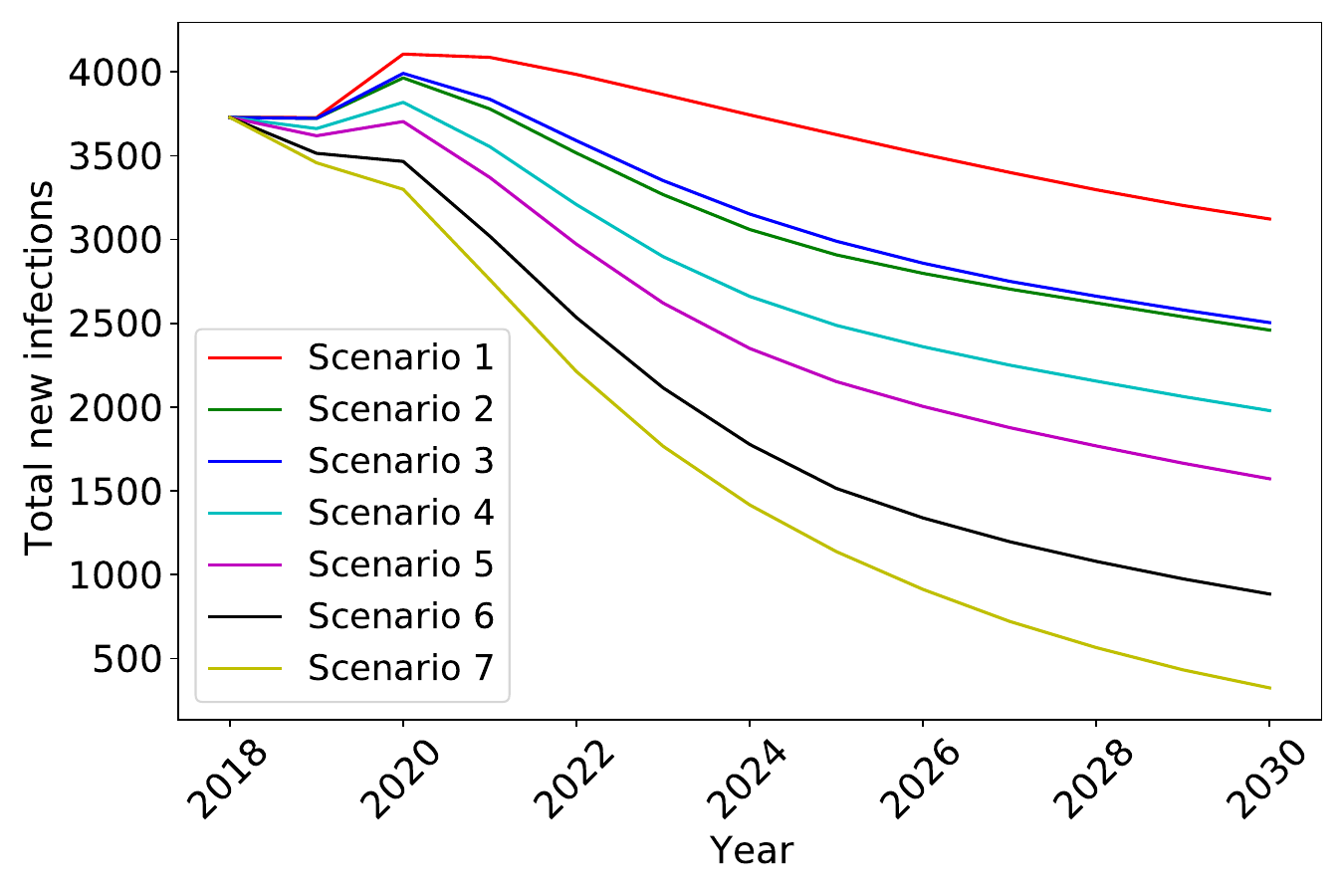}
      \caption[Image 1]{}
      \label{fig:scenario_CA}
  \end{subfigure}
  \begin{subfigure}{0.45\textwidth}
    \includegraphics[width=1\textwidth]{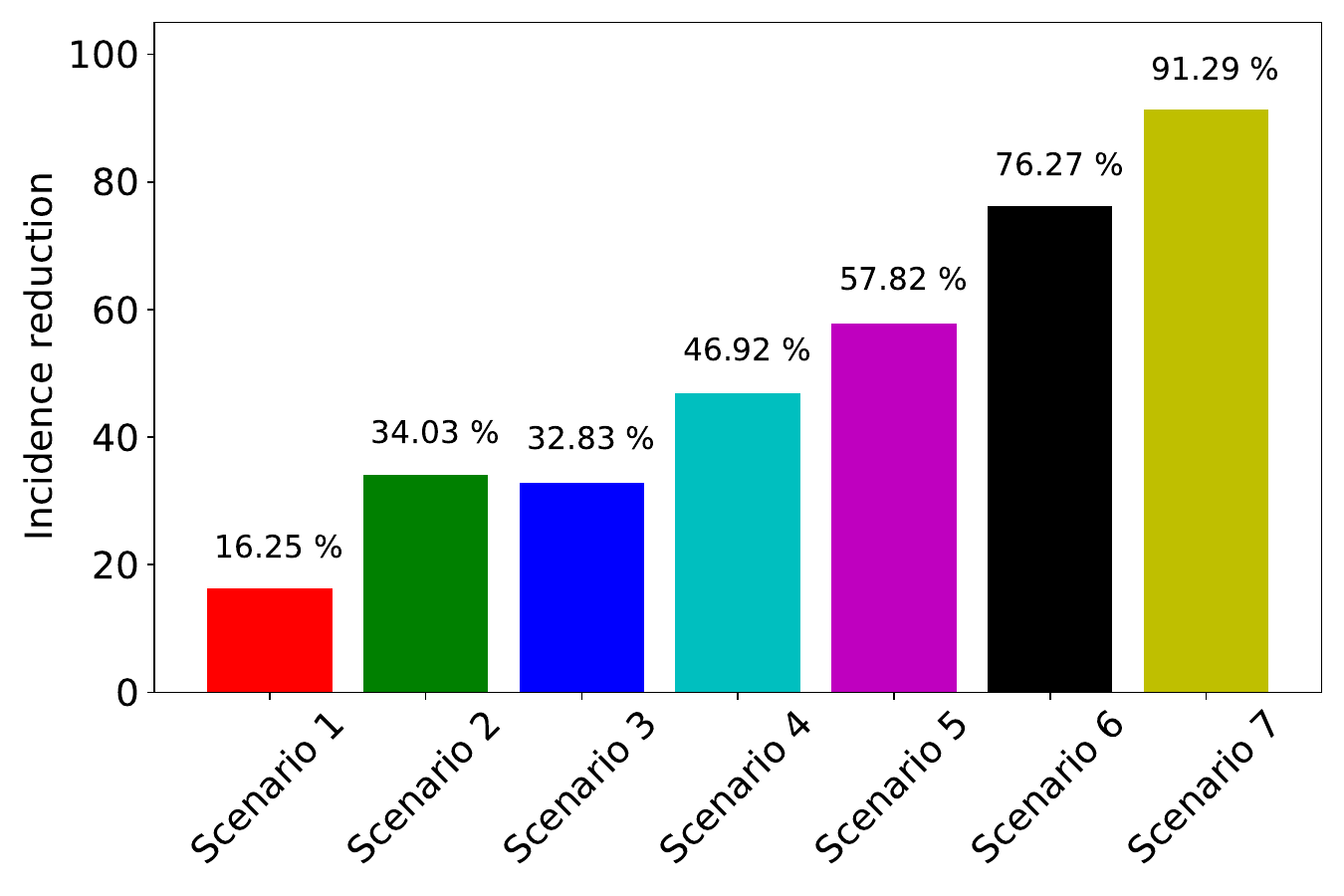}
      \caption[Image 2]{}
      \label{fig:scenario_change_CA}
  \end{subfigure}
  \caption[Images]{(a) Total incidence and b) incidence reduction under different budget allocations in California.}
  \label{fig:incidence_CA}
\end{figure}

\begin{figure}
  \centering
  \begin{subfigure}{0.45\textwidth}
    \includegraphics[width=1\textwidth]{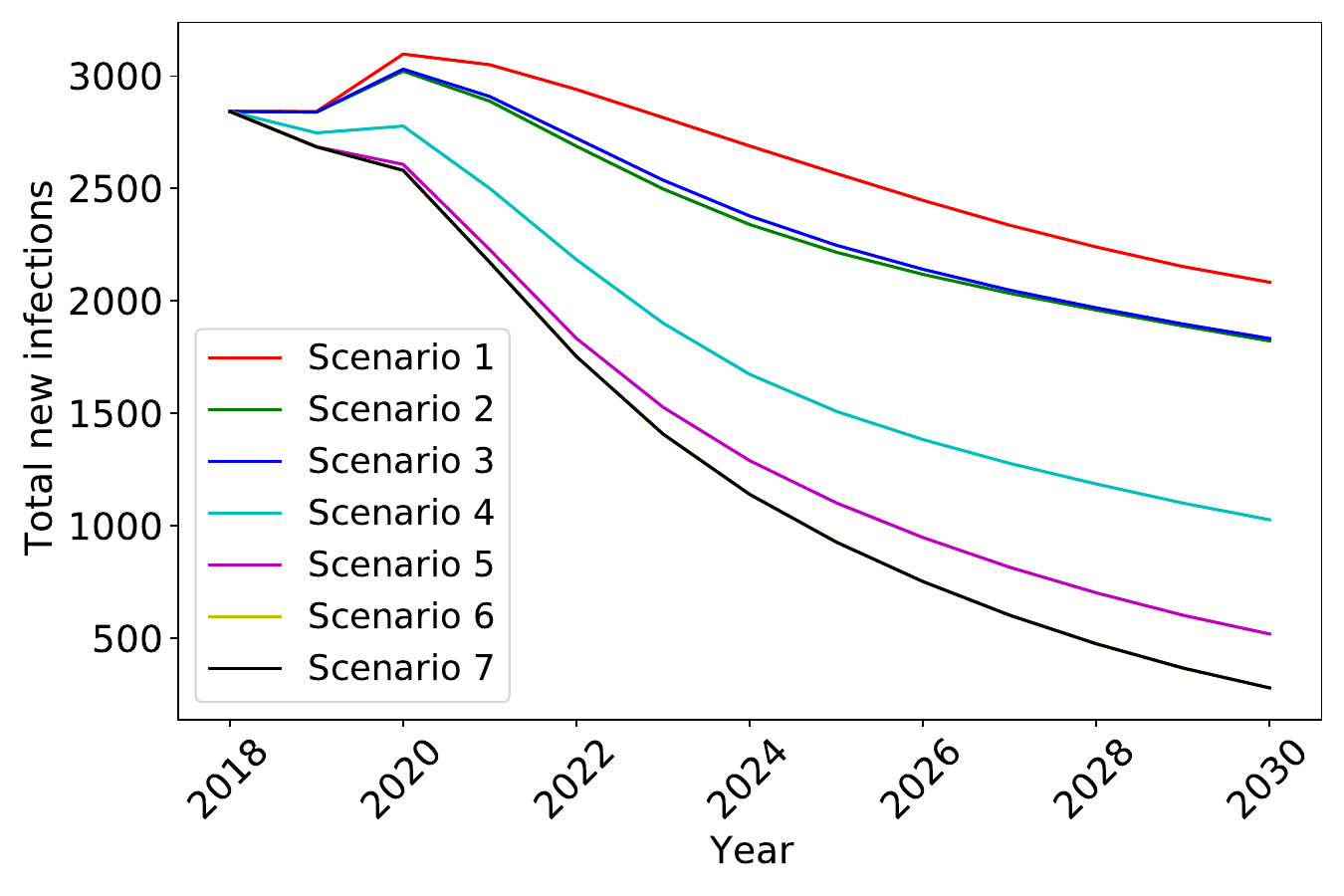}
      \caption[Image 1]{}
      \label{fig:scenario_FL}
  \end{subfigure}
  \begin{subfigure}{0.45\textwidth}
    \includegraphics[width=1\textwidth]{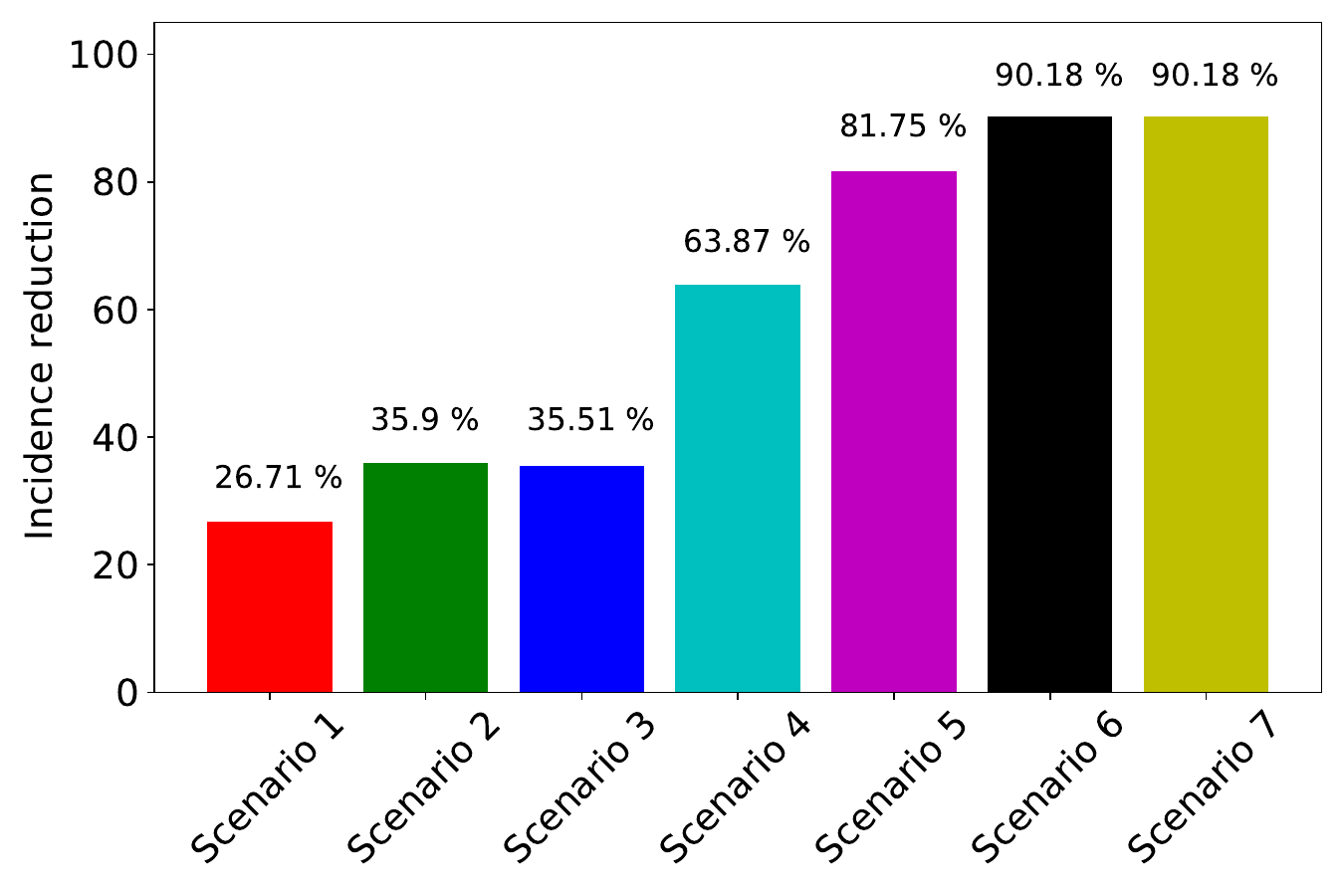}
      \caption[Image 2]{}
      \label{fig:scenario_change_FL}
  \end{subfigure}
  \caption[Images]{(a) Total incidence and b) incidence reduction under different budget allocations in Florida.}
  \label{fig:incidence_FL}
\end{figure}

\begin{figure}
  \centering
  \begin{subfigure}{0.45\textwidth}
    \includegraphics[width=1\textwidth]{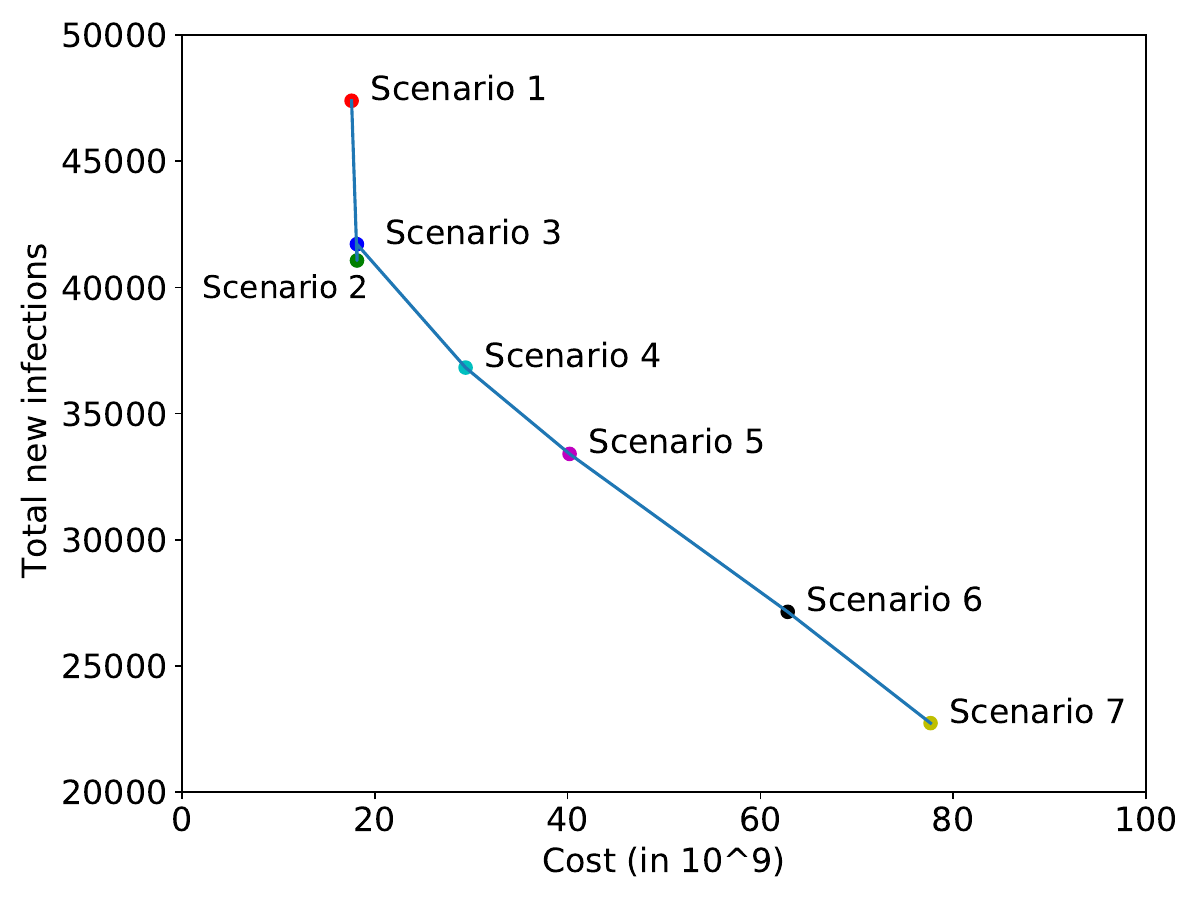}
      \caption[Image 1]{}
      \label{fig:cost_inf_CA}
  \end{subfigure}
  \begin{subfigure}{0.45\textwidth}
    \includegraphics[width=1\textwidth]{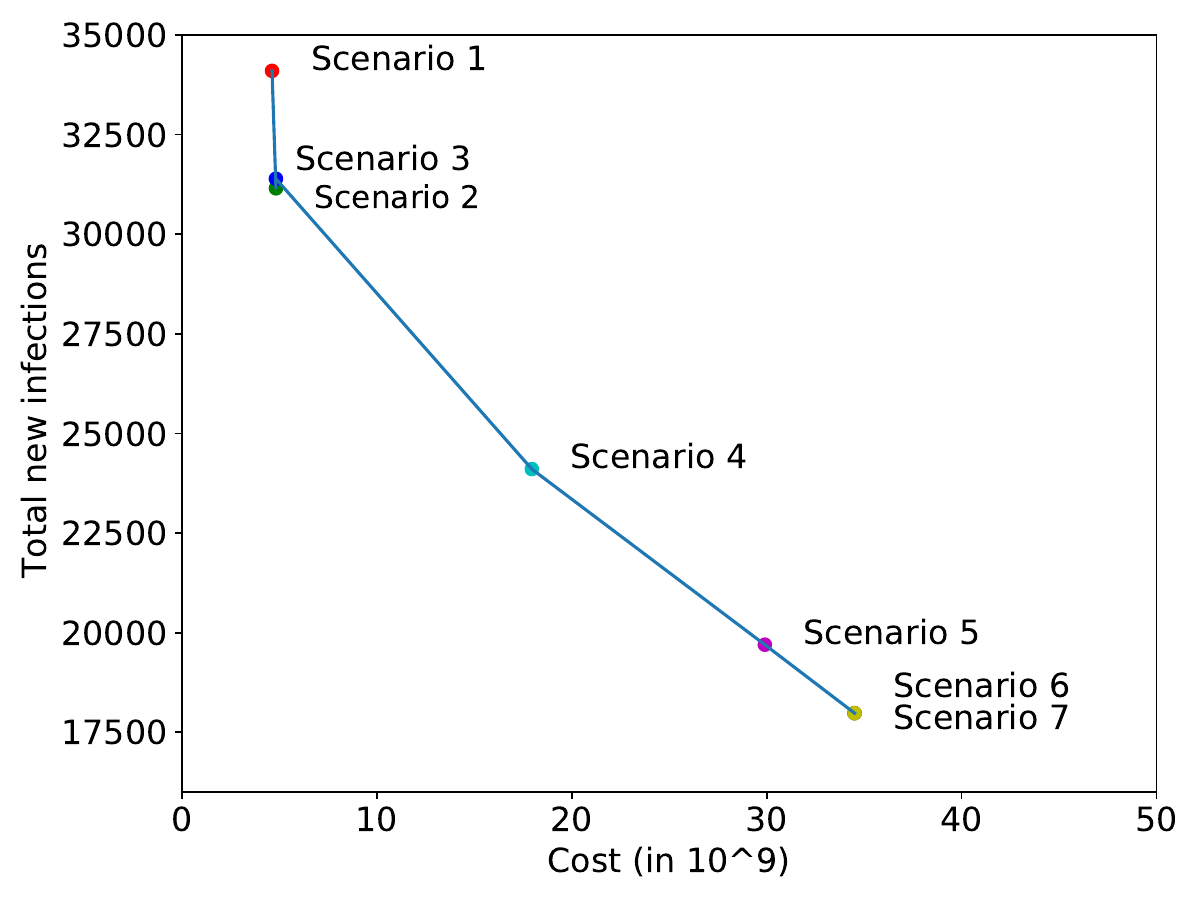}
      \caption[Image 2]{}
      \label{fig:cost_inf_FL}
  \end{subfigure}
  \caption[Images]{Total infection versus intervention cost across scenarios in (a) California and (b) Florida.}
  \label{fig:images}
\end{figure}

We used IPPO for the last set of analyses (Table \ref{scenario_table}). Results suggest that continuing with baseline intervention scale-up and baseline budget when applied to all 96 jurisdictions will result in an incidence reduction of 35\%, by the end of 2030 compared to 2019 (Figure \ref{fig:inf_96_jur}). We next compare across scenarios to evaluate the influence of intervention and budget scale-up for the two test cases (CA and FL). The baseline scenario (Scenario 1) achieved 16\% and 26\% reduction in incidence in CA and FL, respectively (Figures \ref{fig:incidence_CA} and \ref{fig:incidence_FL}), by the end of 2030 compared to 2019. Aggressive intervention scale-up (Scenario 2) achieves a significantly higher incidence reduction (Figures \ref{fig:incidence_CA} and \ref{fig:incidence_FL}) with same costs (Figure \ref{fig:cost_inf_CA} and \ref{fig:cost_inf_FL}), mainly because of faster scale-up in ART (Figure 
\ref{fig:art_proportions_florida}), suggesting that higher spending in retention-in-care programs to increase ART and VLS will be offset from the costs saved from preventing new infections. Addition of penalties alone, for not reaching EHE goal, (Scenario 3) did not have much change as the costs are already at the budget limit. The EHE goal of 90\% reduction in incidence occurred when the annual budget constraint was relaxed to 7 times (Scenario 7) and 15 times (Scenario 6) that of the costs spent in 2019 for CA and FL, respectively. However, the total costs incurred over the period 2019 to 2030 in these scenarios were about 4 times and 7 times that of the costs in the baseline Scenario 1. This is because the optimal policy in some years could be at a lower cost than the budget constraint.  

\subsection{Optimal Intervention Policies}

\begin{figure}
\begin{center}
\includegraphics[width=0.8\linewidth]{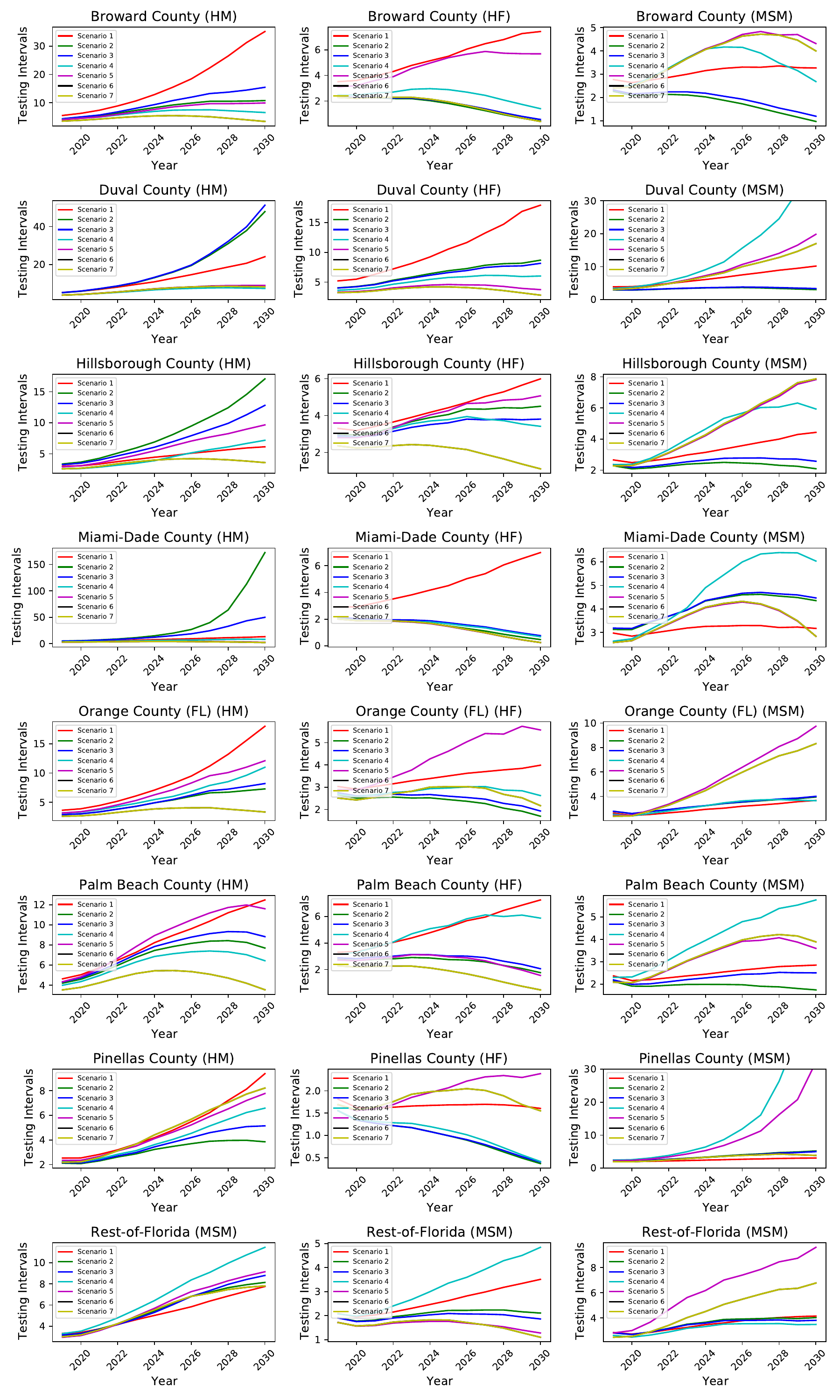}
\end{center}
\caption{Testing interval (in years) scale-up under the optimal policy for HM, HF, and MSM across jurisdictions in Florida.}
\label{fig:testing_frequency_florida}
\end{figure}

\begin{figure}
\begin{center}
\includegraphics[width=0.85\linewidth]{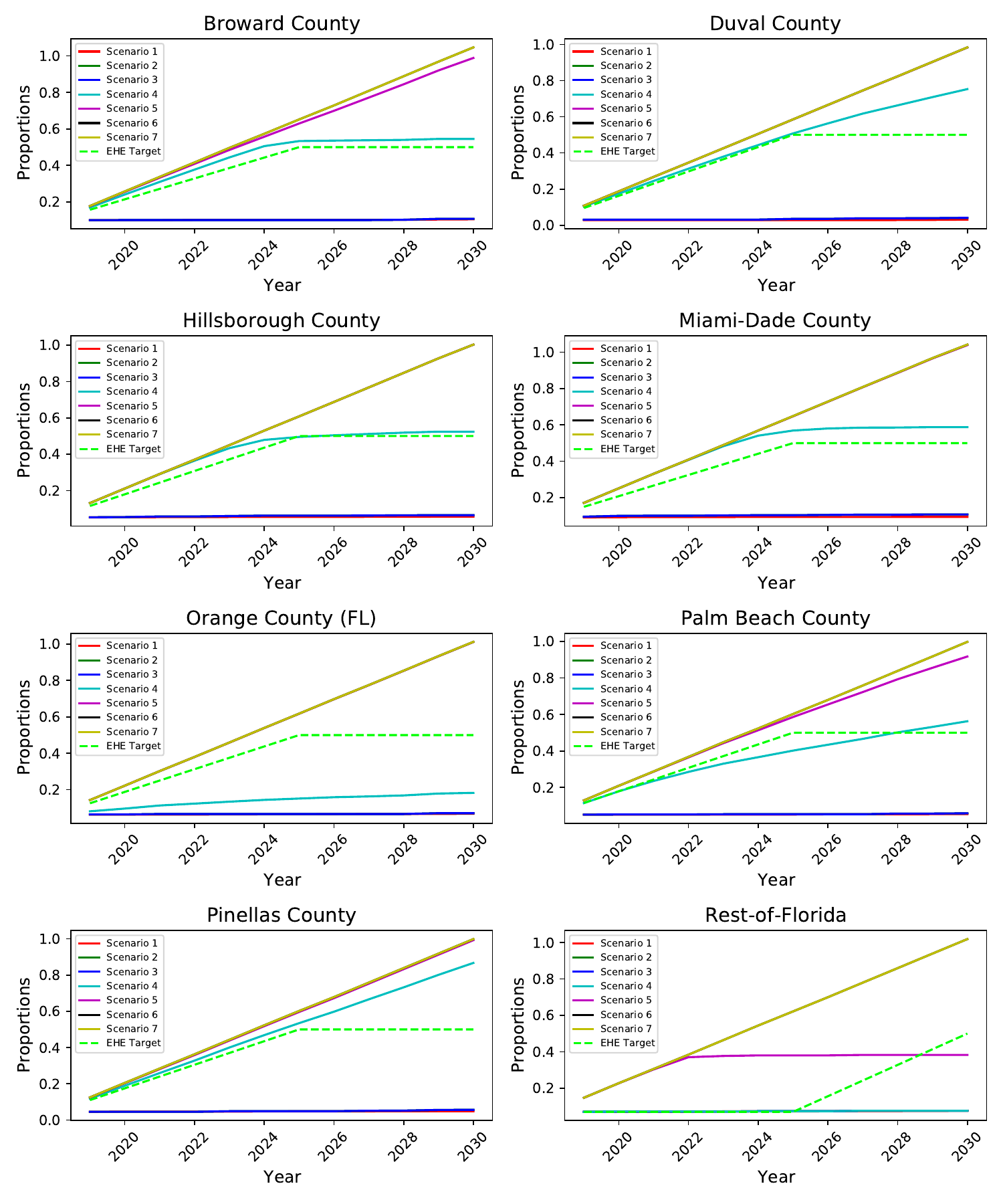}
\end{center}
\caption{ PrEP scale-up under the optimal policy for MSM across jurisdictions in Florida. EHE target denotes the national Ending the HIV Epidemic goal.}
\label{fig:prep_proportions_fl}
\end{figure}

Figure \ref{fig:testing_frequency_florida} and Figure \ref{fig:prep_proportions_fl} show the optimal policies for testing intervals and PrEP allocation for each jurisdiction in FL, corresponding to Scenarios 1 to 7 in Table \ref{scenario_table}. The optimal policies for CA followed a very similar pattern to that of FL, thus only the detailed results for FL have been presented here. In general, the findings suggest that, if budget is limited to current levels (Scenarios 1 to 3), the optimal policy is to first prioritize scale-up of ART, i.e., allocate resources to retention-in-care to achieve viral suppression. These results were consistent when the model was scaled to all 96 jurisdictions. Second, when budget is increased from current levels, to scale-up ART and PrEP, and, as they collectively decrease new infections, testing frequency could be reduced. Third, if the budget was further increased, the optimal policy is to scale-up all three interventions, which resulted in reaching the 90\% EHE incidence goal. As shown in \ref{optimal_policiies_appendix}, comparing across jurisdictions, while ART scale-up was equivalent and high in all jurisdictions indicating prioritizing treatment of all infections, the scale-up in PrEP and testing varied across jurisdictions, generally allocating more resources to jurisdictions and populations with higher risk of infection.

\subsection{Sensitivity Analysis}

We tested the robustness of our learned policy by conducting a sensitivity analysis on HIV related costs and jurisdictional mixing, using the Baseline scenario (Scenario 1). 
HIV related costs: We varied the HIV unit costs of testing, treatment, and PrEP by $\pm 20 \%$ in both California and Florida. Table \ref{tab:cost_sensitivity} gives the overview of  the three scenarios, base, low unit cost ($-20\%$) and high unit cost ($+20\%$). The total cost increased or decreased proportionally to the costs, violating budget constraints in the high cost scenario. However,  new infections did not change significancy suggesting the variability in cost assumptions would impact the financial feasibility, but given the high costs of treatment, the optimal policy would favor preventing those infections.

\begin{table}[t]
\centering
\begin{threeparttable}
\begin{tabular}{|M{2.2cm}|M{1.8cm}|M{2cm}|M{1.8cm}|M{1.8cm}|}
\hline
\multicolumn{5}{|c|}{\textbf{California (CA)}} \\
\hline
\textbf{Scenario} & \textbf{Budget (billion \$)} & \textbf{Total Cost (billion \$)} & \textbf{Budget Violation (billion \$)} & \textbf{\% Above Budget} \\
\hline
Base        & 18 & 17.6 & 0 & 0 \\
Low(-20\%) & 18 & 14.08 & 0 & 0 \\
High(+20\%)& 18 & 21.12 & 3.12 & 17.33 \\
\hline
\multicolumn{5}{|c|}{\textbf{Florida (FL)}} \\
\hline
\textbf{Scenario} & \textbf{Budget (billion \$)} & \textbf{Total Cost (billion \$)} & \textbf{Budget Violation (billion \$)} & \textbf{\% Above Budget} \\
\hline
Base        & 4.8 & 4.61 & 0 & 0 \\
Low (-20\%) & 4.8 & 3.69 & 0 & 0 \\
High (+20\%)& 4.8 & 5.53 & 0.73 & 15.21 \\
\hline
\end{tabular}
\begin{tablenotes}
    \item[*] Budget violation is calculated as $\max(0, \text{Total Cost} - \text{Budget})$. 
    \item[*] \% Above Budget = $\frac{\text{Budget Violation}}{\text{Budget}} \times 100$.
\end{tablenotes}
\end{threeparttable}
\caption{Sensitivity analysis of annual costs relative to budget under base, low (-20\%), and high (+20\%) cost assumptions for California (CA) and Florida (FL).}
\label{tab:cost_sensitivity}
\end{table}

\begin{figure}
  \centering
  \begin{subfigure}{0.45\textwidth}
    \includegraphics[width=1\textwidth]{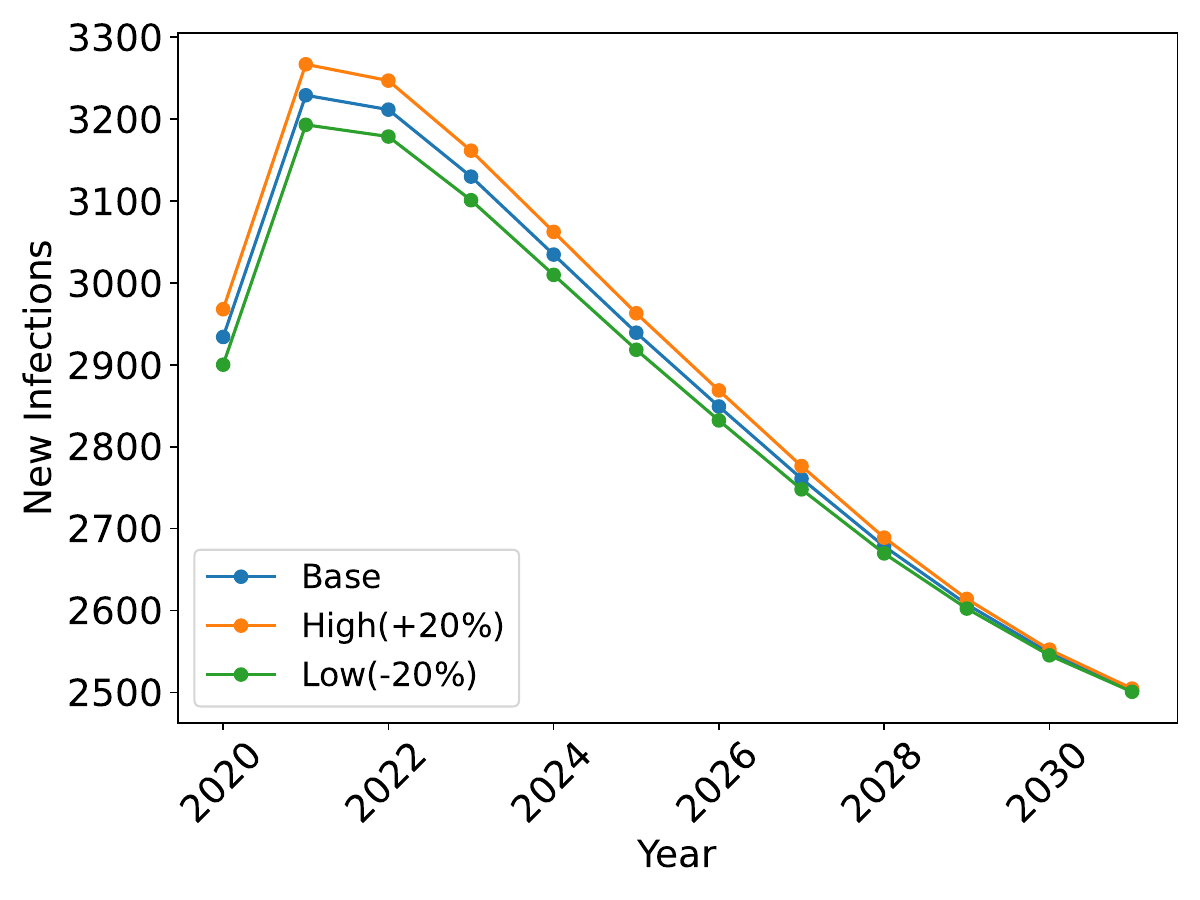}
      \caption[Image 1]{}
      \label{fig:sensitivity_CA}
  \end{subfigure}
  \begin{subfigure}{0.45\textwidth}
    \includegraphics[width=1\textwidth]{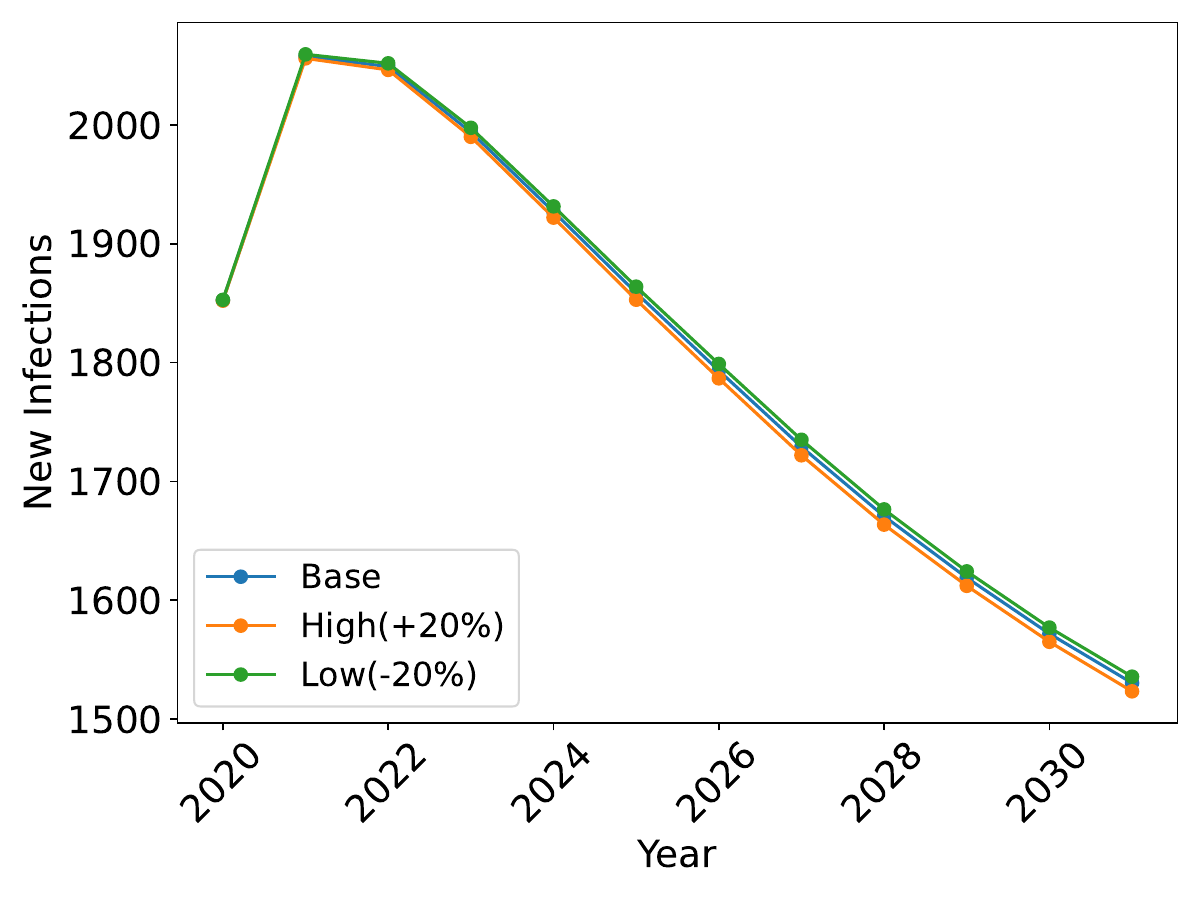}
      \caption[Image 2]{}
      \label{fig:sensitivity_FL}
  \end{subfigure}
  \caption[Images]{Total MSM infections under base, low (-20\%) and high (+20\%) within-jurisdiction mixing in (a) California and (b) Florida.}
  \label{fig:sensitivity_mixing}
\end{figure}

Sensitivity analyses on jurisdictional mixing: As men-who-have-sex-with-men (MSM) risk group are at the highest risk of infection, we focused the sensitivity analysis on this group. Specifically, we increased/decreased mixing within the jurisdiction by $\pm20\%$. Epidemic trajectories in all three scenarios were similar (Figure \ref{fig:sensitivity_mixing}). This indicates that the learned policies are not highly sensitive to moderate changes in behavioral patterns. Although extreme changes in mixing assumptions might alter the epidemic dynamics, the policies from MARL are robust to reasonable changes in mixing assumptions.

\section{Discussion}\label{discuss}

Building on the results presented above, the following discussion interprets the key findings of the proposed framework, emphasizing their policy relevance, methodological contributions, and study limitations.

\subsection{Key Insights and Policy Implications} \label{insights}

This paper presented a MARL framework for optimizing intervention strategies (testing, treatment, and prevention) aimed at ending the HIV epidemic. The proposed MARL approach enables each jurisdiction to make autonomous, data-driven decisions tailored to its specific epidemic context, while considering cross-jurisdictional interactions. By incorporating dynamic decision-making and inter-jurisdictional epidemic modeling, our framework offers a significant advancement over current decision-analytic approaches, such as A-SARL and I-SARL, which represent national aggregated models and independent jurisdiction models, respectively. Comparative analysis shows that both national aggregated and independent jurisdiction models tend to result in sub-optimal decisions due to the lack of consideration for interactions across jurisdictions. Our results further suggest that the impact of intervention strategies can be enhanced by expanding intervention measures and allocating additional resources.

The intervention policy learned by MARL framework showed a context-sensitive prioritization of resources across time. For example, the agent initially invested more heavily in expanding testing and retention, leveraging the immediate impact of these interventions on reducing undiagnosed infections and onward transmission. As the epidemic evolved and budget constraints persisted, the policy adaptively reallocated resources to retention efforts while the testing frequency started to increase, recognizing the compounding benefits of sustained viral suppression through ART adherence. As the budget allocation increased, more resources were allocated towards PrEP and the proportion of MSM under PrEP started to increase leading to significant reduction in new infections. When enough resources are available (scenario 7 of our experiment), the PrEP proportion increased linearly to get all of the MSM population under PrEP while the testing frequency increased initially and decreased again, leading to more testing efforts to reduce the overall infections in the case of adequate resource availability. This highlights the ability of the MARL framework to learn the long-term payoff of early versus deferred investments, given a constrained budget envelope. Importantly, the agent made strategic trade-offs, such as reducing one intervention temporarily to intensify another, indicating that flexible resource allocation over time can outperform fixed or proportional spending across interventions.

Conventional HIV models often evaluate intervention scenarios under fixed resource allocations, using deterministic optimization or scenario testing. By contrast, our DRL approach learns policies dynamically through interaction with a simulated epidemic environment and directly incorporates financial constraints into the decision-making process. The MARL agent integrates the budget constraint into the policy learning loop, allowing it to find intervention mixes that are both cost-feasible and epidemiologically effective. From a policy perspective, our results suggest that dynamic budget-aware intervention policies can lead to more efficient use of limited resources over time. Health departments could use such tools to identify when and how to shift funding between testing programs, PrEP scale-up, and retention services in order to maximize epidemic control. For example, if the learned policy shows that early investments in PrEP yield long-term cost savings via averted infections, this could inform reallocation of funds away from less impactful services in early program years. However, real-world implementation of the policies by translating model recommendations into action should acknowledge the challenges like operational flexibility, inter-program coordination, and political will to shift resources dynamically.

\subsection{Limitations}\label{limit}

While the proposed MARL framework offers substantial advancements, several limitations should be acknowledged. First, the scope of this work was limited to presenting a suitable decision-making tool for resource allocation specific to each jurisdiction in the context of epidemic mixing. We applied the model to all 96 jurisdictions only for the baseline scenario as proof-of-concept on the scalability of the model, making simplistic assumptions on budget estimates being proportion to epidemic burden due to data unavailability. Nonetheless our experimental analyses on California and Florida, on multiple intervention scale-up and budgetary constraints, provided some insights on the significant influence of these variables. Future work would include working with public health agencies to set realistic budget scenarios specific to each jurisdiction, and program scale-up constraints, to identify optimal choices within these system constraints. 

Second, following the EHE plan, we added penalty constraints on overall incidence reduction. Future work could consider adding constraints specific to each jurisdiction for more equitable allocation. 

Third, the model assumes homogeneous behavior within each risk group and static mixing patterns across time. These assumptions may underestimate heterogeneity in sexual networks and evolving dynamics of HIV transmission. Our sensitivity analysis varying MSM mixing by $\pm20\%$ indicated that infection trajectories were robust to moderate perturbations, but more detailed behavioral and network data are needed to improve policy realism.

Additionally, our model focused exclusively on HIV transmission through sexual contact, without considering other transmission routes such as injecting drug use. We assumed static demographics throughout the simulation period and made simplifying assumptions in the absence of certain data, such as partnership mixing between every jurisdiction pair. The simulation environment, while rich, may not fully capture heterogeneity in adherence behavior, network effects, or regional variation in HIV risk. Furthermore, we imposed constraints on the action space, assumed PrEP eligibility for only MSM, and considered all MSM populations are eligible for PrEP. While the model accounted for costs, it used simplified assumptions about intervention unit costs and did not include non-monetary constraints (e.g., staff capacity, supply chain issues). These assumptions, while necessary for model tractability, may limit the generalizability of our findings.

\section{Conclusions}\label{conclude}

This study demonstrates the potential of multi-agent reinforcement learning to inform HIV prevention and treatment strategies under dynamic epidemic and budgetary conditions. By modeling jurisdictions as interacting agents, the MARL framework consistently outperformed classical optimization baselines, including myopic optimization and the cross-entropy method, and single-agent reinforcement learning baselines by adaptively reducing new infections while respecting budget constraints. The results highlight the value of sequential, adaptive decision-making in capturing long-term epidemic dynamics and guiding resource allocation. By providing jurisdiction-specific, data-driven recommendations for intervention strategies, this framework has the potential to significantly impact government resource planning and public health management. It allows policymakers to make more informed, optimized decisions about where to allocate resources, which interventions to prioritize, and how to scale them effectively across different regions. 

Future research can expand this work in several directions. Methodologically, the MARL framework can be extended to incorporate realistic budget and non-monetary constraints, as well as stochastic budget fluctuations in funding during policy implementation. From a modeling perspective, incorporating behavioral heterogeneity, dynamic sexual network structures, and additional transmission routes such as injection drug use would increase the realism of epidemic dynamics. Integrating equity objectives could further support fairer allocation of prevention and treatment resources across jurisdictions. Finally, collaborative implementation with public health agencies could enable real-world validation and policy translation, generating actionable insights to inform the next phase of national HIV prevention and treatment strategies.

\section*{Glossary}

\begin{itemize}
    
\item \textbf{Ending the HIV Epidemic (EHE)} - Initiative by U.S. Department of Health and Health Service (HHS) to reduce new HIV infections in U.S. by 75\% by 2025 and by 90\% by 2030

\item \textbf{Jurisdiction} - Counties and states that contribute significantly to total HIV infections and are used in the HIV modeling

\item \textbf{HIV prevalence} - Number of persons living with HIV at a given time regardless of their time of infection, whether the person has received the diagnosis, or the stage of HIV disease

\item \textbf{Antiretroviral Therapy (ART)} - Treatment of HIV that involves taking a combination of HIV medicines everyday

\item \textbf{Viral Load Suppression (VLS)} - Having less than 200 copies of HIV per milliliter of blood

\item \textbf{Pre-exposure Prophylaxis (PrEP)} - Medicine taken by people at risk for HIV to prevent getting HIV from sex or injection drug use 

\item \textbf{HIV care continuum} - Steps that people with HIV take from diagnosis to achieving and maintaining viral load suppression 

\item \textbf{Retention-in-care} - Adherence to drug regimens and other clinical or lifestyle-change recommendations made by the providers

\item \textbf{Compartmental Simulation Model} - Model with various compartments where each individual in a population can be on one compartment and total population is the sum of all the b compartments

\item \textbf{CD4 Count} - Blood test that measures the number of CD4 cells in the blood

\item \textbf{Heterosexual Male/Female} - People who have heterosexual contact

\item \textbf{Men who have sex with men (MSM)} - People who have male to male sexual contact and men who have sexual contact with both men and women

\item \textbf{Diagnostic rate} - Estimated number of new HIV diagnosis among the people unaware of their HIV infection

\item \textbf{Dropout rate} - Estimated number of people leaving the HIV treatment regime among the people in HIV care

\item \textbf{Independent Proximal Policy Optimization (IPPO)} - Multi-agent extension of Proximal Policy Optimization where each agent has the visibility of its observation only

\item \textbf{Centralized Training with Decentralized Execution (CTDE)} - Multi-agent framework where each agent makes decision based on its observation and evaluates based on the overall observation of all the agents

\end{itemize}




\section*{Data availability}
Data will be made available on request.

\appendix


\section{Baseline Optimization Formulations}\label{optimization_formulation}

\subsection{Myopic Optimization}

At each time step $t$, the myopic baseline chooses an intervention allocation 
$a_t \in \mathbb{R}^{72}$ (9 actions $\times$ 8 jurisdictions) that minimizes 
immediate new infections, subject to a budget constraint. 
The immediate reward is defined as:

\begin{equation}\label{eq:19}
    R(s_t,a_t) = - \text{NewInf}(s_t,a_t) \;-\; \kappa \cdot \max\!\bigl(0, C(s_t,a_t) - B_t \bigr)
\end{equation}

where
\begin{itemize}
    \item $\text{NewInf}(s_t,a_t)$ = projected new infections in period $t$,
    \item $C(s_t,a_t)$ = intervention cost under $a_t$,
    \item $B_t$ = available budget in period $t$,
    \item $\kappa \gg 1$ = penalty weight for overspending.
\end{itemize}

The optimization problem is:

\begin{align}\label{eq:20}
    \max_{a_t} \quad & R(s_t, a_t) \\
    \text{s.t.} \quad & a_t \in \mathcal{A}
\end{align}

where $\mathcal{A}$ is the feasible action space (e.g., bounds on scale-up rates for testing, ART, and PrEP).  
This is a greedy, one-step optimization repeated at each decision epoch.

\vspace{1em}

\subsection{Cross-Entropy Method (CEM)}

The CEM baseline performs a stochastic search over multi-period intervention sequences 
$a_{1:T} = (a_1,\ldots,a_T)$. 
The cumulative reward is defined as:

\begin{equation}\label{eq:21}
    R(s_{1:T},a_{1:T}) \;=\; - \sum_{t=1}^T \gamma^{t-1}\, \text{NewInf}(s_t,a_t) 
\;-\; \kappa \sum_{t=1}^T \max\!\bigl(0, C(s_t,a_t) - B_t \bigr)
\end{equation}

where
\begin{itemize}
    \item $\gamma \in (0,1]$ is the discount factor,
    \item the first term penalizes cumulative new infections,
    \item the second term penalizes budget violations across the horizon.
\end{itemize}

The optimization problem is:

\begin{align}\label{eq:22}
    \max_{a_{1:T}} \quad & R(s_{1:T},a_{1:T}) \\
    \text{s.t.} \quad & a_t \in \mathcal{A}, \quad \forall t.
\end{align}

\section{Deep Reinforcement Learning Algorithm}\label{marl_algorithm}

Among the various DRL algorithms, PPO~\cite{schulman2017proximal} is  known for its stability and efficiency. It is an on-policy algorithm that can be used for environments with continuous action spaces and very large state spaces. Our research problem has the characteristics that are well suited for such an approach. PPO adopts the concepts from trust region policy optimization (TRPO) \cite{schulman2015trust}. The objective in TRPO, called surrogate advantage, is calculated as:
\begin{equation}\label{eq:5}
    L^{CPI}(\theta) = \hat{E}_{t}\left[ \dfrac{\pi_{\theta}(a_{t}|s_{t})}{\pi_{\theta_{l}}(a_{t}|s_{t})}\hat{A}_{t}\right] = \hat{E}_{t}[r_{t}(\theta)\hat{A}_{t}]
\end{equation}
The probability ratio $r_{t}(\theta)$ of two policies gives a measure of how far the new policy $\pi_{\theta}$ is from the old policy $\pi_{\theta_{k}}$. As having a large policy update at a time would destabilize the training, PPO-clip puts a limit on the extent of policy update with the following objective.
\begin{equation}\label{eq:6}
    L^{CLIP}(\theta) = \hat{E}_{t} \left[ \min(r_{t}(\theta)\hat{A}_{t}, clip(r_{t}(\theta), 1-\epsilon,1+\epsilon)\hat{A}_{t})\right],
\end{equation}
where $\epsilon$ is the hyperparameter that clips the probability ratio and prevents $r_{t}$ from moving outside the interval $[1-\epsilon, 1+\epsilon]$ depending on whether the advantage is positive or negative. The surrogate objective for PPO-clip can be generalized with the following equation: 
\begin{equation}\label{eq:7}
    L_{(s,a,\theta_{l},\theta)}^{CLIP} = \min \left( \dfrac{\pi_{\theta}(a|s)}{\pi_{\theta_{l}}(a|s)}A^{\pi_{\theta_{l}}}(s,a), g(\epsilon,A^{\pi_{\theta_{l}}}(s,a))\right)
\end{equation}
where,
\begin{equation}\label{eq:8}
    g(\epsilon,A) =
        \begin{cases}
            (1+\epsilon)A & A \ge 0 \\
            (1-\epsilon)A & A < 0
    \end{cases}
\end{equation}

IPPO is a natural extension of standard PPO in multi-agent settings. In IPPO, each agent uses the standard PPO training pipeline, making it a versatile baseline for multi-agent reinforcement learning tasks with reliable performance. In IPPO, the agents do not have access to global state. Thus, the policy learning can be obtained by modifying equation \ref{eq:7} as following,

\begin{equation}\label{eq:9}
    L_{(o,a,\theta_{l},\theta)}^{CLIP} = \min \left( \dfrac{\pi_{\theta}(a|o)}{\pi_{\theta_{l}}(a|o)}A^{\pi_{\theta_{l}}}(o,a), g(\epsilon,A^{\pi_{\theta_{l}}}(o,a))\right)
\end{equation}

The details of implementation for IPPO is given in Algorithm \ref{alg:marl}.

\begin{algorithm}
\setlength{\baselineskip}{10pt}
\caption{MARL Framework for Optimal Interventions in the HIV Epidemic}\label{alg:marl}
Input: Initialize policy parameters $\theta_{l}^{j}$, initialize value function parameters $\phi_{l}^{j}$ for $j =1,2,...,N$ agents, set episode length $T$ (here we used $T = 12$ years corresponding to year 2019 to 2030), set number of episodes $M$, set buffer size $B$ (here we used $B = 10$ episodes), set total timesteps $X = 0$, set $l = 0$. Initialize replay buffer $\mathcal{D}_{l}^{j}$ for each agent $j$. 
\begin{algorithmic}[1]
\For{episode $m = 0, 1, 2, ..., M$} 
    \For{year $t = 1, 2, ..., T$}
        \State \parbox[t]{0.9\linewidth}{At $t=1$, extract the initial observation for each agent $j$ as: $o_{t}^{j} = [p_{k,j,t}, \mu_{u,k,j,t}, \mu_{a,k,j,t}, \mu_{ART,k,j,t}, \mu_{prep,k,j,t}; \forall k \in \{HM, HF, MSM\}].$}
        
        \State \parbox[t]{0.9\linewidth}{Sample actions $a_{t}^{j}$  comprising the change in proportions from each agent $j$ by running policy $\pi_{l}^{j} = \pi_{\theta_{l}^{j}}$; $a_{t}^{j} = [\textbf{a}_{unaware,k,t}^{j}, \textbf{a}_{ART,k,t}^{j}, \textbf{a}_{prep,k,t}^{j}; \forall k \in \{HM,HF,MSM\}].$}
        
        \State \parbox[t]{0.9\linewidth}{Calculate the diagnostic and dropout rates using the change in unaware and ART proportions, respectively.}
        
        \State{Simulate the actions in the compartmental simulation model.}
        
        \State \parbox[t]{0.9\linewidth}{Observe next observation $o_{t+1}^{j}$ and reward ${R}_{t+1}^{j}$, and collect set of trajectories $\{\tau_{t}^{j}\}$, where $\tau_{t}^{j} = (o_{t}^{j}, a_{t}^{j}, R_{t+1}^{j}, done).$}
        
        \State{Add the trajectories to the replay buffers $\mathcal{D}_{l}^{j}$.}
        
        \State{$X \gets X + 1$}
        
        \If{\textbf{($X$ mod $B \times T$) = 0}}
            \State \parbox[t]{0.85\linewidth}{Compute advantage estimates, ${A}_{t}^{j}$ based on the current value function $V_{\phi_{l}^{j}}^{j}$.}
            
            \State \parbox[t]{0.85\linewidth}{Update the policy by maximizing objective (via stochastic gradient ascent): $\theta_{l+1}^{j} = argmax_{\theta^{j}} \\ \dfrac{1}{|\mathcal{D}_{l}^{j}|T} \displaystyle\sum_{\tau_{t}^{j} \in \mathcal{D}_{l}^{j}} \displaystyle\sum_{t=0}^{T} \min \left( \dfrac{\pi_{\theta^{j}}(a_{t}^{j}|o_{t}^{j})}{\pi_{\theta_{l}^{j}}(a_{t}^{j}|o_{t}^{j})} A^{\pi_{\theta_{l}^{j}}} (o_{t}^{j},a_{t}^{j}), g(\epsilon, A^{\pi_{\theta_{l}^{j}}} (o_{t}^{j},a_{t}^{j})) \right).$}
            
            \State \parbox[t]{0.85\linewidth}{Fit value function by regression on mean-squared error:
            \[
            \phi_{l+1}^{j} = argmin_{\phi^{j}} \dfrac{1}{|\mathcal{D}_{l}^{j}| T} \displaystyle\sum_{\tau_{t}^{j} \in \mathcal{D}_{l}^{j}} \displaystyle\sum_{t=0}^{T} \left( V_{\phi_{l}^{j}}^{j}(o_{t}^{j}) - {R}_{t}^{j} \right) ^ {2}.
            \]}
            \State{Clear replay buffer for each agent $j$.}
            \State{$l \gets l + 1$}
        \EndIf
    \EndFor 
\EndFor
\end{algorithmic}
\end{algorithm}

IPPO is a basic form of MARL that does not require information sharing between the agents. In CTDE, there is information sharing across agents, where the actor produces the action from its local observation while the critic utilizes the observation of its local agent and the observation of all other agents in its training. Further, in another form of above, CTDE (action), the critic additionally utilizes the actions taken by all agents. Here we evaluate all three algorithms. 

The objective in CTDE can be stated as:

\begin{equation}\label{eq:10}
    L_{(o,s,a,a^{-},\theta_{l},\theta)}^{CLIP} = \min \left( \dfrac{\pi_{\theta}(a|o)}{\pi_{\theta_{l}}(a|o)}A^{\pi_{\theta_{l}}}(o,s,a), g(\epsilon,A^{\pi_{\theta_{l}}}(o,s,a))\right)
\end{equation}

Here, $a$ is the action of the current agent and $s$ is the global agent.

Similarly, critic for CTDE can be updated with the following equation:

\begin{equation}\label{eq:11}
    \phi_{l+1}^{j} = argmin_{\phi^{j}} \dfrac{1}{|\mathcal{D}_{l}^{j}| T} \displaystyle\sum_{\tau_{t}^{j} \in \mathcal{D}_{l}^{j}} \displaystyle\sum_{t=0}^{T} \left( V_{\phi_{l}^{j}}^{j}(o_{t}^{j}, s_{t}, a_{t}^{j}) - {R}_{t}^{j} \right) ^ {2}
\end{equation} 

The objective in CTDE (action) can be stated as:

\begin{equation}\label{eq:12}
    L_{(o,s,a,a^{-},\theta_{l},\theta)}^{CLIP} = \min \left( \dfrac{\pi_{\theta}(a|o)}{\pi_{\theta_{l}}(a|o)}A^{\pi_{\theta_{l}}}(o,s,a^{-}), g(\epsilon,A^{\pi_{\theta_{l}}}(o,s,a^{-}))\right)
\end{equation}

Here, $a$ is the action of the current agent, $a^{-}$ is the action of all the agents except the current agent, and $s$ is the global agent.

Similarly, critic for CTDE (action) can be updated with the following equation:

\begin{equation}\label{eq:13}
    \phi_{l+1}^{j} = argmin_{\phi^{j}} \dfrac{1}{|\mathcal{D}_{l}^{j}| T} \displaystyle\sum_{\tau_{t}^{j} \in \mathcal{D}_{l}^{j}} \displaystyle\sum_{t=0}^{T} \left( V_{\phi_{l}^{j}}^{j}(o_{t}^{j}, s_{t}, a_{t}^{j^{-}}) - {R}_{t}^{j} \right) ^ {2}
\end{equation}

Lines 13 and 14 in Algorithm \ref{alg:marl} can be replaced by Equation \ref{eq:10} and \ref{eq:11} to implement CTDE. Similarly, lines 13 and 14 in Algorithm \ref{alg:marl} can be replaced by Equation \ref{eq:12} and \ref{eq:13} to implement CTDE (action).

\section{Budget Allocation} \label{budget_appendix}

\setcounter{figure}{0}  
\setcounter{table}{0}   

The budget allocated for Florida (baseline budget) is presented in Table \ref{florida_budget}.

\begin{table}[H]
\centering
\begin{tabular}{|M{4cm}|M{4cm}|}
 \hline
 \textbf{Jurisdiction } & \textbf{Budget (\$ in  millions)} \\
 \hline
 Broward County & 89.8 \\
 \hline
 Duval County & 8.1 \\
 \hline
 Hillsborough County & 27.0 \\
 \hline
 Miami-Dade County & 75.9 \\
 \hline
 Orange County & 34.7 \\
 \hline
 Palm Beach County & 19.5 \\
 \hline
 Pinellas County & 17.2 \\
 \hline
 Rest of Florida & 117.8  \\ \hline
 \end{tabular}
\caption{Budget Allocation for Florida}
\label{florida_budget}
\end{table}

\section{Hyperparameter Tuning} \label{hyperparamters_tune}

\setcounter{figure}{0}  
\setcounter{table}{0}   

\begin{figure}[h]
\centering  
\includegraphics[width=0.95\linewidth]{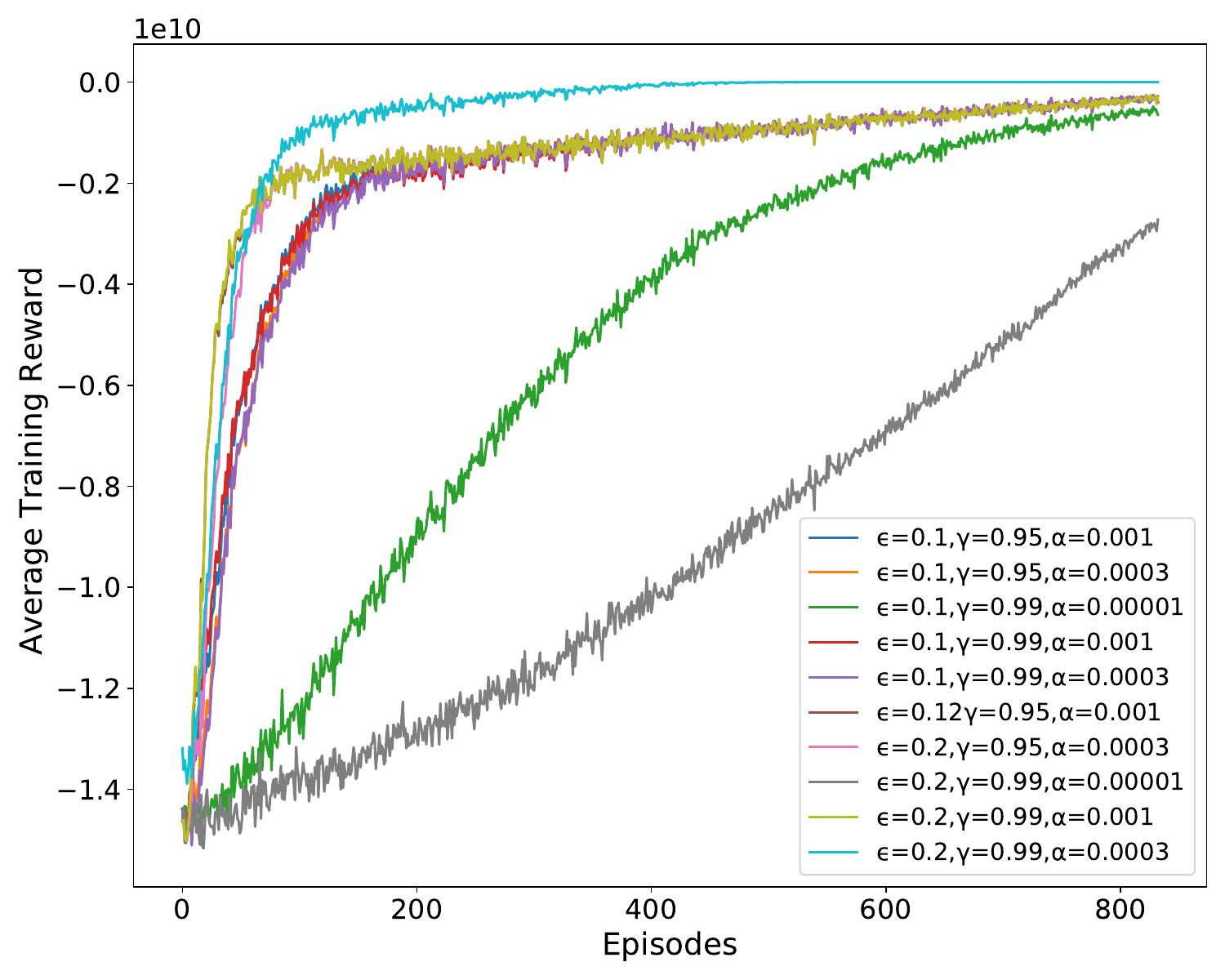} 
\caption{Training reward trajectories across IPPO hyperparameter settings (10 runs).} \label{fig:hyperparameter_graph}
\end{figure}

Tuned hyperparameters, along with experimented values and corresponding performance metrics (average reward), are presented in Table \ref{tab:hyper_tuning} and Figure \ref{fig:hyperparameter_graph}. Each experiment was run for 100,000 time steps with training horizon of 12 steps accounting for the years from 2019 to 2030. Learning rate had the most influence on average reward: a higher value (0.001) had more variance, while a lower value (0.00001) slowed convergence (see Table \ref{tab:hyper_tuning}). A moderate learning rate of 0.0003 consistently achieved the best trade-off between stability and performance. A higher discount factor of $\lambda=0.99$ outperformed smaller value of $\lambda=0.95$, showing the importance of valuing long-term epidemic outcomes. The clipping parameter of $\epsilon=0.2$ provided better stability compared to $\epsilon=0.1$. Thus, for the analyses, hyperparameter values were set as $\lambda=0.99$, $\alpha=0.0003$, $\epsilon=0.2$. Although the lowest total infections corresponded to a different combination, the selected one was the only combination that did not violate the budget constraint, thus resulting in the highest average reward. Further, new infections were not substantially different across these combinations.

\begin{table}[htbp]
\centering
\caption{Performance of PPO under different hyperparameter settings.}
\renewcommand{\arraystretch}{1.2}
\begin{tabular}{|M{2cm}|M{1.5cm}|M{1.5cm}|M{2.5cm}|M{2cm}|}
\hline
\textbf{Discount Factor ($\gamma$)} & \textbf{Learning Rate} & \textbf{Clipping ($\epsilon$)} & \textbf{Average Reward} & \textbf{Total Infections} \\
\hline
0.95 & 0.001   & 0.1 &  -393884313.42 &  44861.79 \\
0.95 & 0.001   & 0.2 &  -276331572.34 &  43434.05 \\
0.95 & 0.0003  & 0.1 &  -397030731.32 &  44309.17 \\
0.95 & 0.0003  & 0.2 &  -279835943.71 &  \textbf{42434.54} \\
0.99 & 0.001   & 0.1 &  -394757799.22 &  44881.87 \\
0.99 & 0.001   & 0.2 &  -279474822.68 &  42862.11 \\
0.99 & 0.0003  & 0.1 &  -398200831.17 &  44202.68 \\
0.99 & 0.0003  & 0.2 &  \textbf{-43250.73} &  43250.73 \\
0.99 & 0.00001 & 0.1 &  -644864912.26 &  44741.91 \\
0.99 & 0.00001 & 0.2 &  -2721804941.81 &  43777.18 \\
\hline
\end{tabular}
\label{tab:hyper_tuning}
\end{table}

\section{Training Rewards}

\setcounter{figure}{0}  
\setcounter{table}{0}   

\begin{figure}[H]
\begin{center}
\includegraphics[width=0.99\linewidth]{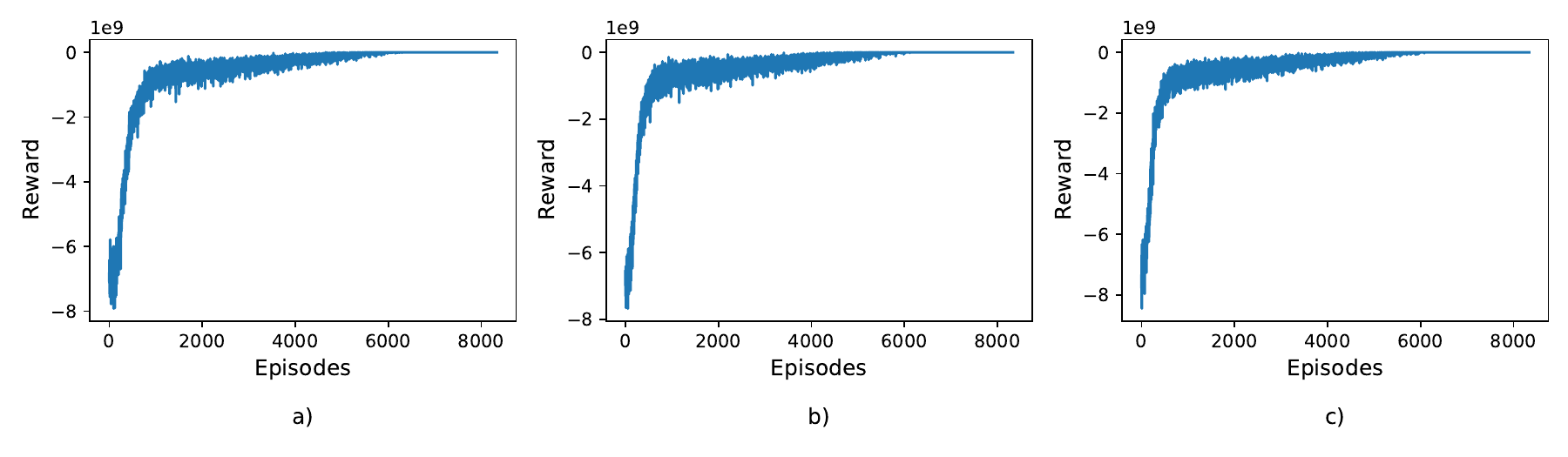}
\end{center}
\caption{Episode rewards for MARL agents using (a) IPPO, (b) CTDE, and (c) CTDE (action) in Florida. Training includes eight actor–critic pairs for the eight jurisdictions.}
\label{fig:reward_marl_fl}
\end{figure}

\begin{figure}[H]
\begin{center}
\includegraphics[width=0.99\linewidth]{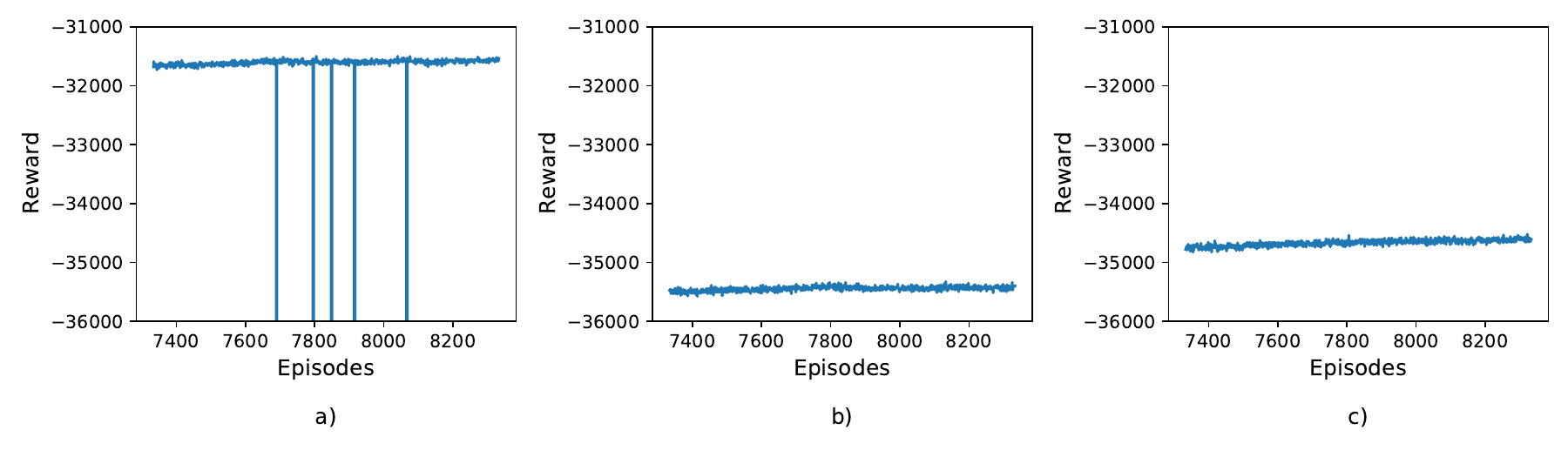}
\end{center}
\caption{Episode rewards over the last 1,000 episodes for MARL agents using (a) IPPO, (b) CTDE, and (c) CTDE (action) in Florida.}
\label{fig:last_1000_reward_marl_fl}
\end{figure}

Figure \ref{fig:reward_marl_fl} shows the learning curve for MARL in Florida. As the reward for each agent is same for one MARL framework, single plot is show as the representative of all the agents. Figure \ref{fig:last_1000_reward_marl_fl} show the episode rewards for final 1000 episodes for all three MARL frameworks in  Florida. These plots show the convergence of the reward values in the final phase of the training of various MARL frameworks.

Figure \ref{fig:last_1000_reward_marl_fl} shows the reward for last 1000 episodes for three MARL algorithms; IPPO, CTDE and CTDE (action) for Florida. It shows the convergence of rewards for all three methods at different points. As HIV is a slow spreading disease, the state space of HIV  performance of IPPO is better than CTDE and CTDE (action). As the objective is to minimize the total infection over the state and each agent in IPPO is trying to minimize its own reward of total infection, it tries to minimize the infections in the jurisdictions with largest number of infections without trying to co-operate with other agents, thus outperforms the other two algorithms.

\section{Comparison of new HIV incidence with historical data}

\setcounter{figure}{0}  
\setcounter{table}{0}   

\begin{figure}[!httb]
\begin{center}
\includegraphics[width=0.99\linewidth]{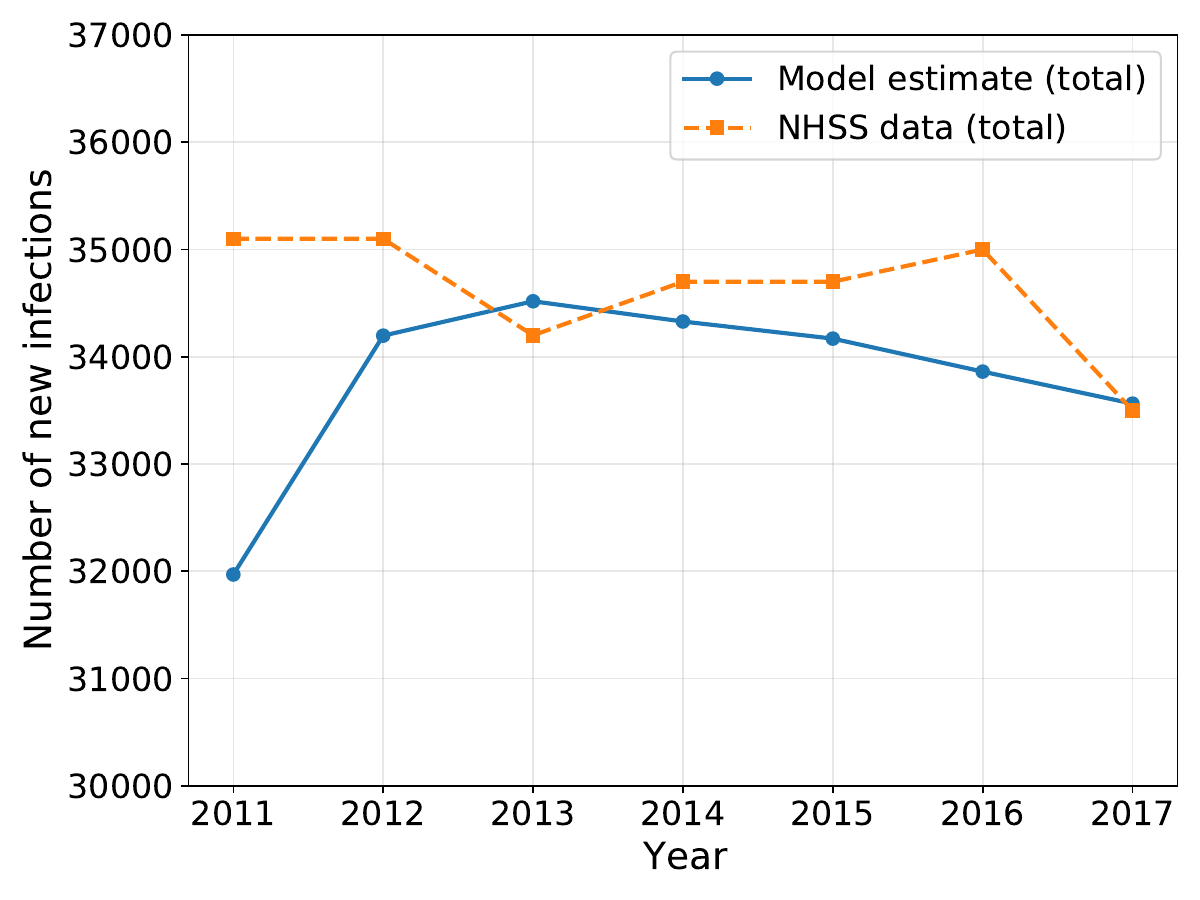}
\end{center}
\caption{Comparison of simulated and NHSS-estimated new infections in the U.S. for the years 2011–2017.}
\label{fig:historical_nhss}
\end{figure}

Figure \ref{fig:historical_nhss} shows comparison of the national-level estimates of new infections calculated from the simulator used in our model with the historical data reported by National Health Surveillance System (NHSS) for the years 2011-2017, taken from \cite{tatapudi2022evaluating}. The new infection estimates are very close to the actual historical HIV data, and thus provide the credibility for the simulator used in our framework.

\section{New incidence under baseline}

\setcounter{figure}{0}  
\setcounter{table}{0}   

Figure \ref{fig:incidence_FL} compares aggregated incidence (new infections) estimates in Florida, from 2019 to 2030 under the baseline budget and baseline action for independent single-agent reinforcement learning (I-SARL), aggregated single-agent reinforcement learning (A-SARL) and multi-agent reinforcement 
learning (MARL). Here, we present incidence estimates specific to each jurisdiction, solved using MARL (IPPO).

\begin{figure}[!httb]
\begin{center}
\includegraphics[width=0.99\linewidth]{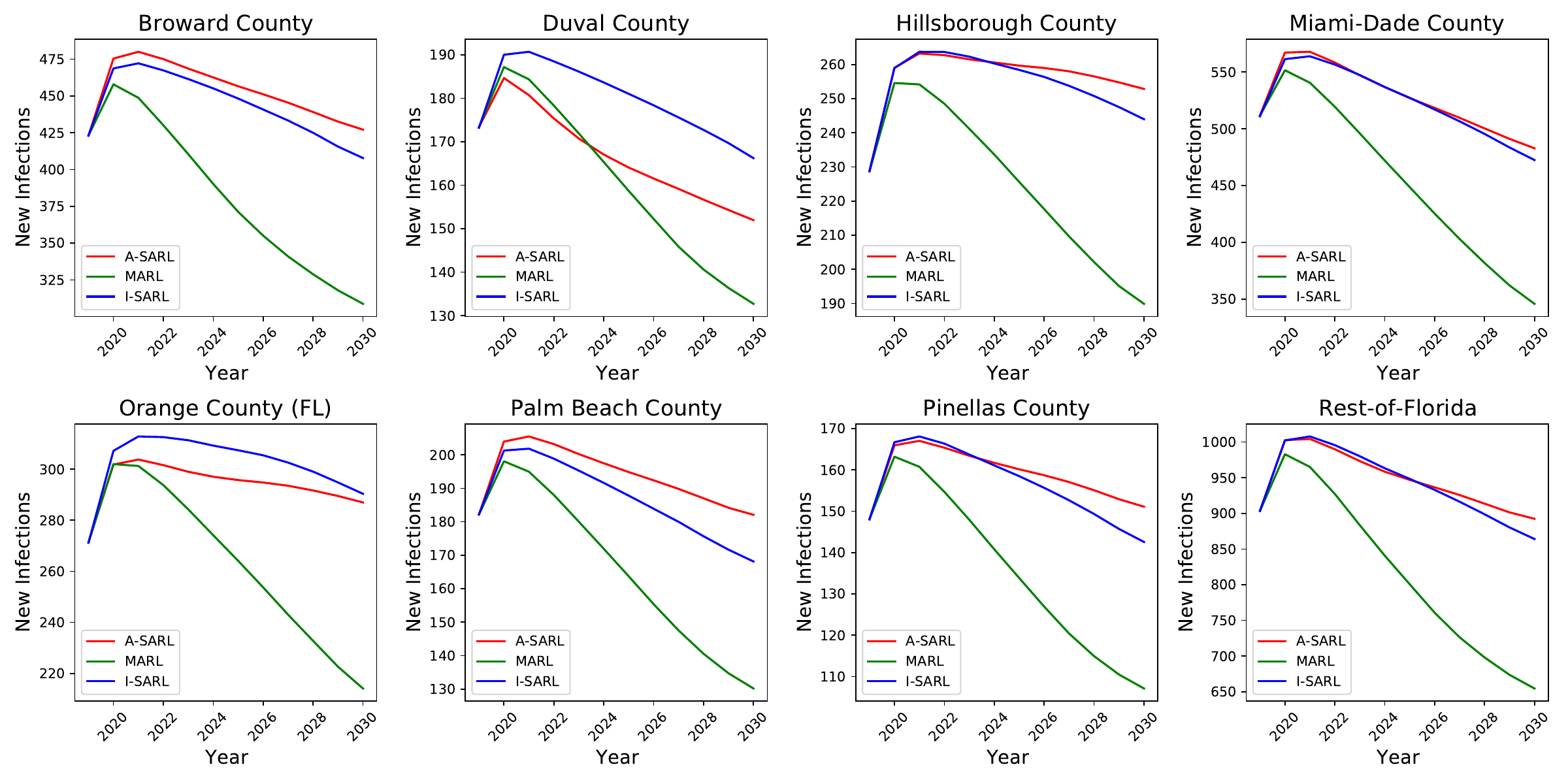}
\end{center}
\caption{New HIV incidence under I-SARL, A-SARL, and MARL with baseline budget and actions in Florida.}
\label{fig:incidence_florida}
\end{figure}

\section{Antiretroviral Therapy (ART) Scale-up Under Optimal Policy}

\setcounter{figure}{0}  
\setcounter{table}{0}   

\begin{figure}
\begin{center}
\includegraphics[width=0.8\linewidth]{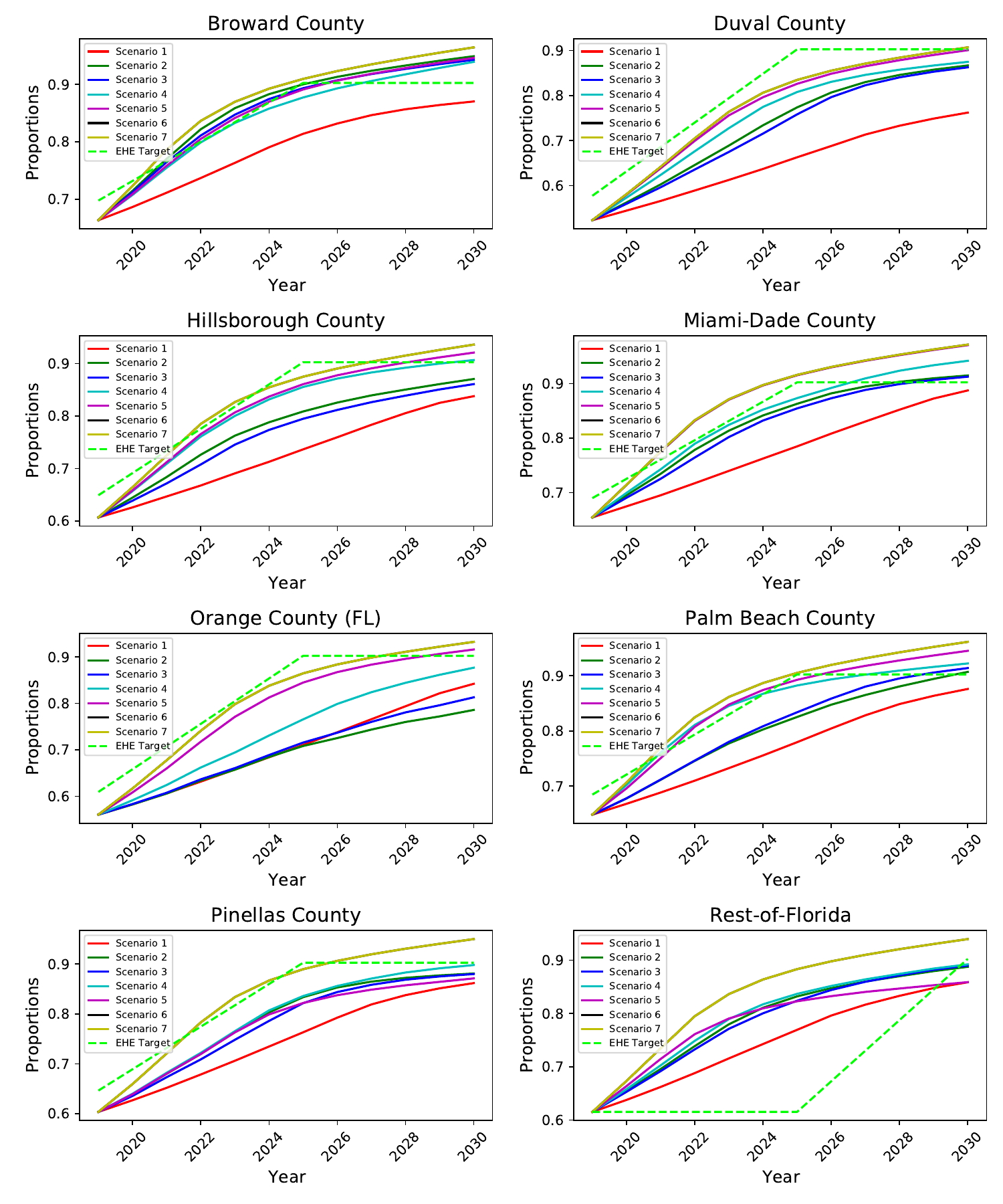}
\end{center}
\caption{ART scale-up under the optimal policy across jurisdictions in Florida. EHE target denotes the national Ending the HIV Epidemic goal.}
\label{fig:art_proportions_florida}
\end{figure}

ART scale-up under optimal policy were similar across risk groups within each scenario. Figure \ref{fig:art_proportions_florida} presents optimal ART scale-up aggregated over all risk groups, in each jurisdiction in Florida.

\section{Summary of Optimal Policies} \label{optimal_policiies_appendix}

\setcounter{figure}{0}  
\setcounter{table}{0}   

\begin{figure}[H]
  \centering
  \begin{subfigure}{0.7\textwidth}
    \centering
    \includegraphics[width=1\textwidth]{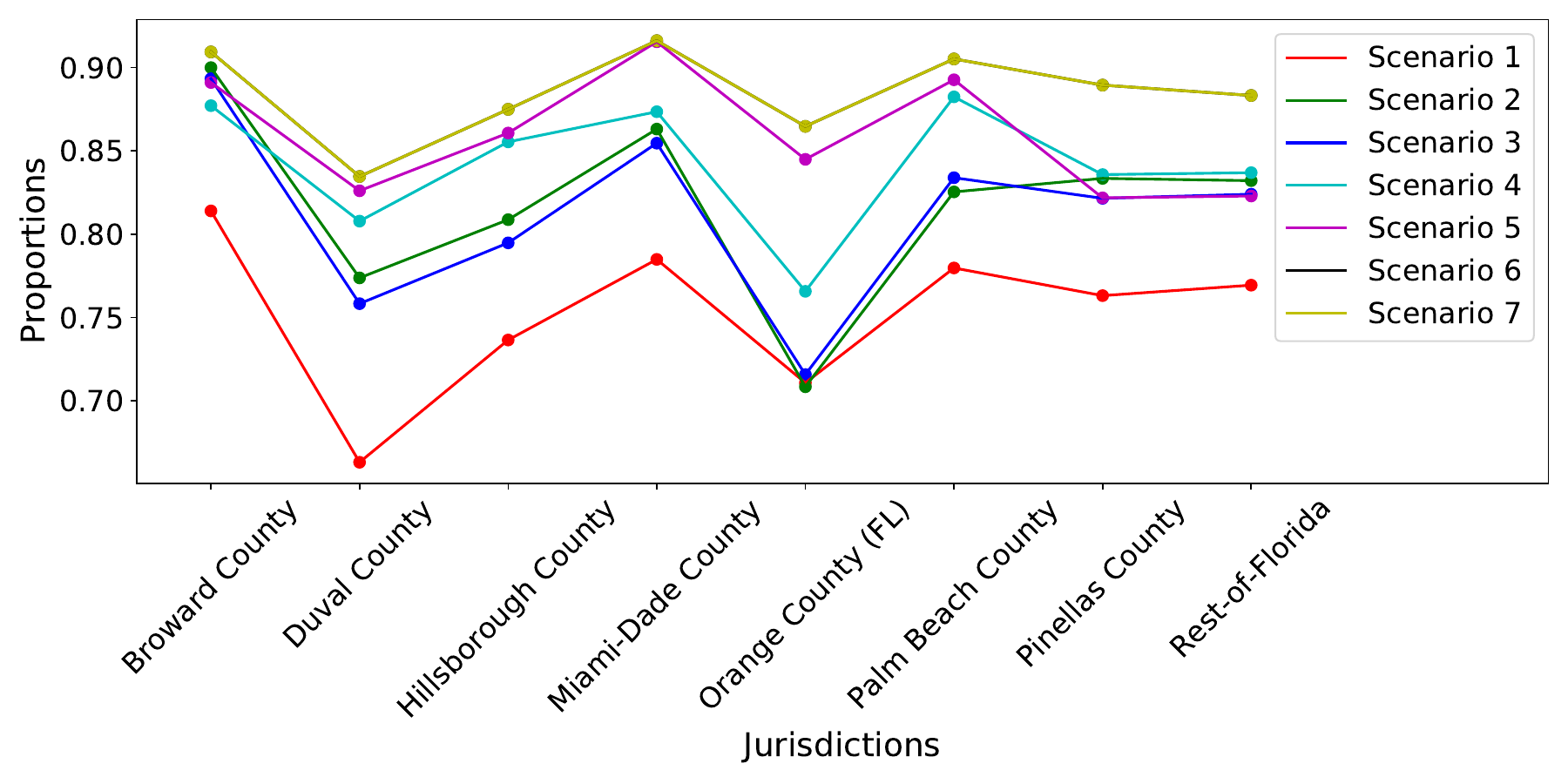}
      \caption[Image 1]{}
      \label{fig:art_2025_fl}
  \end{subfigure}
  \begin{subfigure}{0.7\textwidth}
    \centering
    \includegraphics[width=1\textwidth]{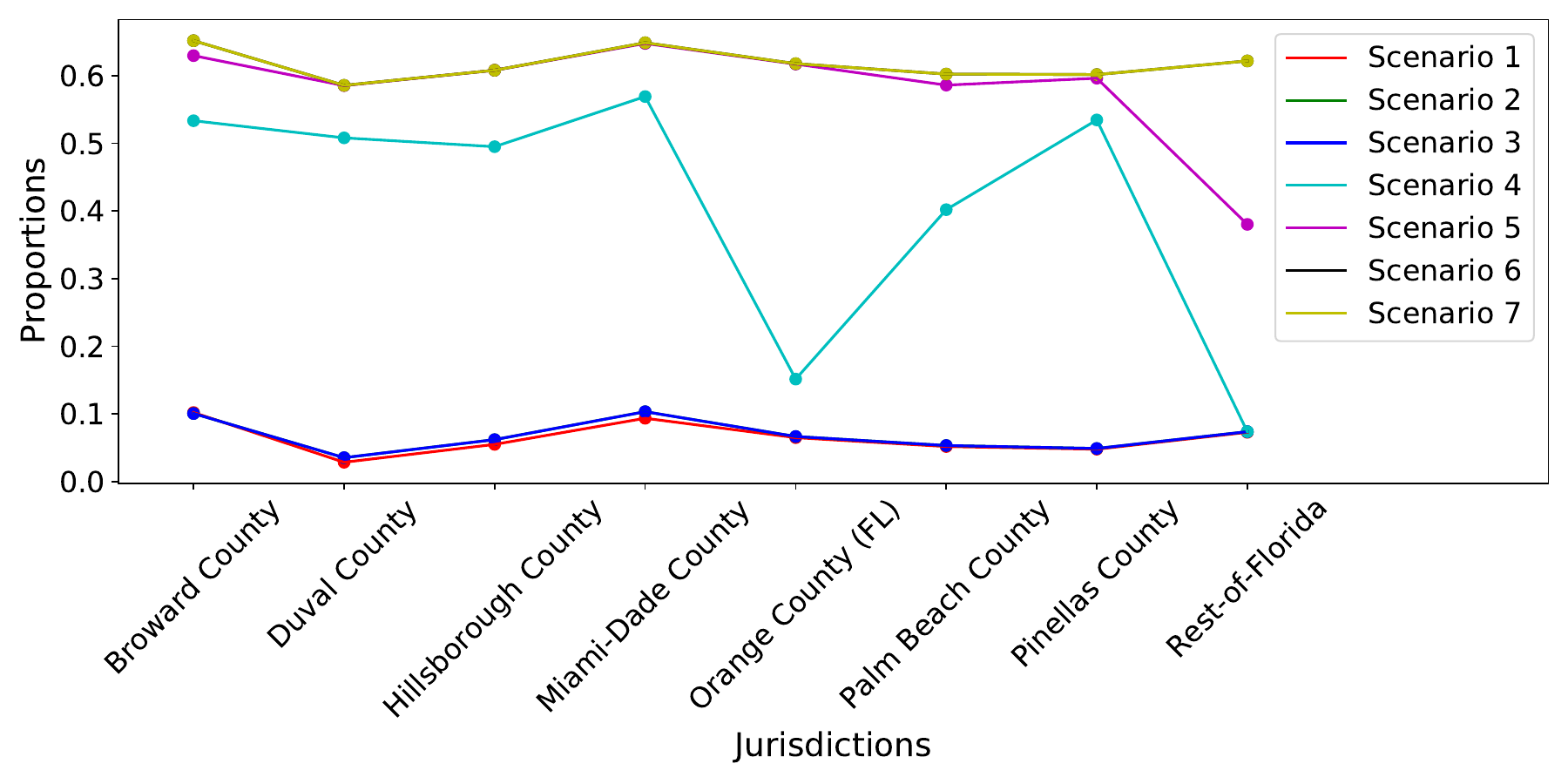}
      \caption[Image 2]{}
      \label{fig:prep_2025_fl}
  \end{subfigure}

  \begin{subfigure}{0.7\textwidth}
    \centering
    \includegraphics[width=1\textwidth]{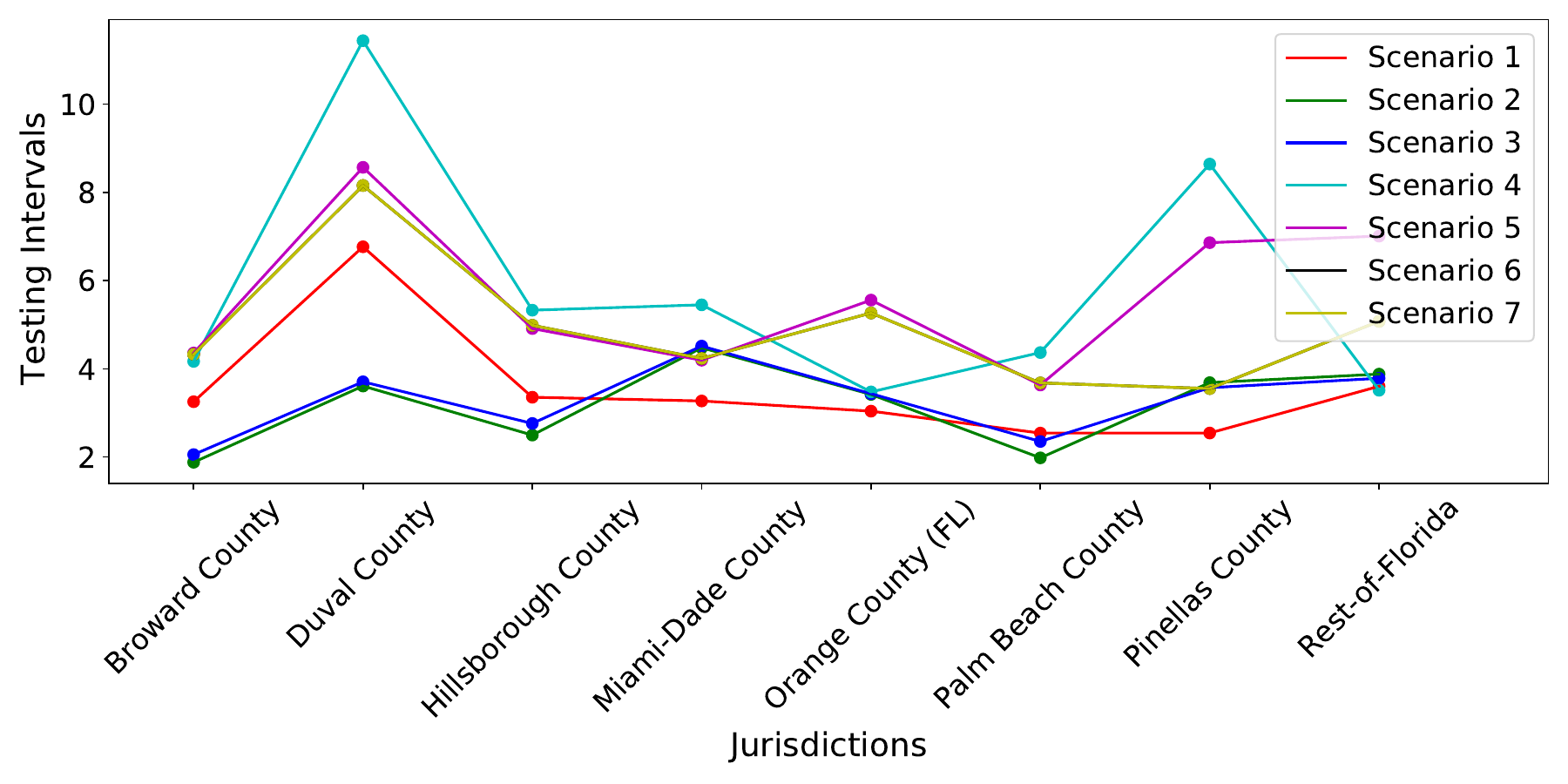}
    \caption[Image 2]{}
    \label{fig:testing_msm_2025_fl}
  \end{subfigure}
  \caption[Images]{(a) ART, (b) PrEP, and (c) testing intervals for MSM in 2025 under different scenarios  across jurisdictions in Florida.}
  \label{fig:intervention_fl}
\end{figure}

To synthesize the outcomes of the optimal policies under varying constraints (Experiment 3 Scenarios in Table \ref{scenario_table}) we observe the results specific to 2025. 
Figures \ref{fig:art_2025_fl}, \ref{fig:prep_2025_fl}, and \ref{fig:testing_msm_2025_fl} present the optimal scale-up in ART, PrEP, and testing, respectively, in 2025 for Florida. In general, we  see that, when only the constraints on the action space are relaxed (Scenarios 2 and 3) without changing the budget, the optimal policy is to scale-up ART. Relaxing budget constraints further leads to incremental scale-up in ART (Scenarios 4 and above). In Scenarios 1, 2 and 3, where budget constraints are the same but the action space is relaxed or there is a penalty, there is not much change in PrEP proportions. Going from scenario 3 to 4, where there is an increase in budget, the optimal policy is to scale-up PrEP while decreasing testing. 
Further, when budget is further increased, the optimal policy is to continue to increase PrEP while also maintaining high testing (going from Scenario 4 to  5 and above in Florida 
). These findings suggest that the optimal policy is to first prioritize ART if budget is limited to current levels. Second, to scale-up ART and PrEP when budget is increased from current levels and, as they collectively decrease new infections, reduce the testing frequency. However, if budget was further increased, there could be a scale-up in all three interventions. Comparing across jurisdictions, while priority in ART scale-up was equivalent in all jurisdictions, the scale-up in PrEP and testing varied across jurisdictions, generally first prioritizing jurisdictions with higher incidence or MSM populations.

\bibliography{sample}

@article{avancena2020optimization,
  title={Optimization models for HIV/AIDS resource allocation: A systematic review},
  author={Avance{\~n}a, Anton LV and Hutton, David W},
  journal={Value in Health},
  volume={23},
  number={11},
  pages={1509--1521},
  year={2020},
  publisher={Elsevier}
}

@article{wang2024optimal,
  title={Optimal resource allocation model for COVID-19: a systematic review and meta-analysis},
  author={Wang, Yu-Yuan and Zhang, Wei-Wen and Lu, Ze-xi and Sun, Jia-lin and Jing, Ming-xia},
  journal={BMC Infectious Diseases},
  volume={24},
  number={1},
  pages={200},
  year={2024},
  publisher={Springer}
}

@article{silal2021operational,
  title={Operational research: A multidisciplinary approach for the management of infectious disease in a global context},
  author={Silal, Sheetal Prakash and others},
  journal={European journal of operational research},
  volume={291},
  number={3},
  pages={929--934},
  year={2021},
  publisher={Elsevier}
}

@book{powell2007approximate,
  title={Approximate Dynamic Programming: Solving the curses of dimensionality},
  author={Powell, Warren B},
  volume={703},
  year={2007},
  publisher={John Wiley \& Sons}
}

@book{rubinstein2004cross,
  title={The cross-entropy method: a unified approach to combinatorial optimization, Monte-Carlo simulation and machine learning},
  author={Rubinstein, Reuven Y and Kroese, Dirk P},
  year={2004},
  publisher={Springer Science \& Business Media}
}

@article{schulman2017proximal,
  title={Proximal policy optimization algorithms},
  author={Schulman, John and Wolski, Filip and Dhariwal, Prafulla and Radford, Alec and Klimov, Oleg},
  journal={arXiv preprint arXiv:1707.06347},
  year={2017}
}

@inproceedings{schulman2015trust,
  title={Trust region policy optimization},
  author={Schulman, John and Levine, Sergey and Abbeel, Pieter and Jordan, Michael and Moritz, Philipp},
  booktitle={International conference on machine learning},
  pages={1889--1897},
  year={2015},
  organization={PMLR}
}

@article{lowe2017multi,
  title={Multi-agent actor-critic for mixed cooperative-competitive environments},
  author={Lowe, Ryan and Wu, Yi I and Tamar, Aviv and Harb, Jean and Pieter Abbeel, OpenAI and Mordatch, Igor},
  journal={Advances in neural information processing systems},
  volume={30},
  year={2017}
}

@article{yu2022surprising,
  title={The surprising effectiveness of ppo in cooperative multi-agent games},
  author={Yu, Chao and Velu, Akash and Vinitsky, Eugene and Gao, Jiaxuan and Wang, Yu and Bayen, Alexandre and Wu, Yi},
  journal={Advances in Neural Information Processing Systems},
  volume={35},
  pages={24611--24624},
  year={2022}
}

@article{krebs2019developing,
  title={Developing a dynamic HIV transmission model for 6 US cities: an evidence synthesis},
  author={Krebs, Emanuel and Enns, Benjamin and Wang, Linwei and Zang, Xiao and Panagiotoglou, Dimitra and Del Rio, Carlos and Dombrowski, Julia and Feaster, Daniel J and Golden, Matthew and Granich, Reuben and others},
  journal={PloS one},
  volume={14},
  number={5},
  pages={e0217559},
  year={2019},
  publisher={Public Library of Science San Francisco, CA USA}
}

@article{zang2020development,
  title={Development and calibration of a dynamic HIV transmission model for 6 US cities},
  author={Zang, Xiao and Krebs, Emanuel and Min, Jeong E and Pandya, Ankur and Marshall, Brandon DL and Schackman, Bruce R and Behrends, Czarina N and Feaster, Daniel J and Nosyk, Bohdan},
  journal={Medical Decision Making},
  volume={40},
  number={1},
  pages={3--16},
  year={2020},
  publisher={SAGE Publications Sage CA: Los Angeles, CA}
}

@article{nosyk2019ending,
  title={Ending the epidemic in America will not happen if the status quo continues: modeled projections for human immunodeficiency virus incidence in 6 US cities},
  author={Nosyk, Bohdan and Zang, Xiao and Krebs, Emanuel and Min, Jeong Eun and Behrends, Czarina N and Del Rio, Carlos and Dombrowski, Julia C and Feaster, Daniel J and Golden, Matthew and Marshall, Brandon DL and others},
  journal={Clinical Infectious Diseases},
  volume={69},
  number={12},
  pages={2195--2198},
  year={2019},
  publisher={Oxford University Press US}
}

@article{khatami2021reinforcement,
  title={A reinforcement learning model to inform optimal decision paths for HIV elimination},
  author={Khatami, Seyedeh N and Gopalappa, Chaitra},
  journal={Mathematical biosciences and engineering: MBE},
  volume={18},
  number={6},
  pages={7666},
  year={2021},
  publisher={NIH Public Access}
}

@article{khurana2018impact,
  title={Impact of improved HIV care and treatment on PrEP effectiveness in the United States, 2016--2020},
  author={Khurana, Nidhi and Yaylali, Emine and Farnham, Paul G and Hicks, Katherine A and Allaire, Benjamin T and Jacobson, Evin and Sansom, Stephanie L},
  journal={JAIDS Journal of Acquired Immune Deficiency Syndromes},
  volume={78},
  number={4},
  pages={399--405},
  year={2018},
  publisher={LWW}
  }

@article{wheatley2022cost,
  title={Cost-Effectiveness of Interventions to Improve HIV Pre-exposure Prophylaxis Initiation, Adherence, and Persistence Among Men Who Have Sex With Men},
  author={Wheatley, Margo M and Knowlton, Gregory and Kao, Szu-Yu and Jenness, Samuel M and Enns, Eva A},
  journal={JAIDS Journal of Acquired Immune Deficiency Syndromes},
  volume={90},
  number={1},
  pages={41--49},
  year={2022},
  publisher={Wolters Kluwer}
}

@article{tatapudi2022evaluating,
  title={Evaluating the sensitivity of jurisdictional heterogeneity and jurisdictional mixing in national level HIV prevention analyses: context of the US ending the HIV epidemic plan},
  author={Tatapudi, Hanisha and Gopalappa, Chaitra},
  journal={BMC Medical Research Methodology},
  volume={22},
  number={1},
  pages={1--20},
  year={2022},
  publisher={BioMed Central}
}

@misc{usehe,
    author = {HIV.gov},
    title = {{EHE Priority Jurisdictions}},
    url  = {https://www.hiv.gov/federal-response/ending-the-hiv-epidemic/overview},
    note = "(Accessed March 2025)",
    year = 2025
}

@misc{usstatistics,
    author = {HIV.gov},
    title = {{U.S. Statistics}},
    howpublished  = "\url{https://www.hiv.gov/hiv-basics/overview/data-and-trends/statistics}",
    note = "(Accessed February 2025)",
    year = 2025
}

@misc{hivaheadfaq,
    author = {AHEAD},
    title = {{Ending the HIV Epidemic}},
    howpublished  = "\url{https://ahead.hiv.gov/ehe/}",
    note = "(Accessed August 2022)",
    year = 2019
}

@misc{hivprep,
    author = {CDC},
    title = {{Effectiveness of Prevention Strategies to Reduce the Risk of Acquiring or Transmitting HIV}},
    howpublished  = "\url{https://www.cdc.gov/hivpartners/php/riskandprevention/index.html}",
    note = "(Accessed: August 2024)",
    year = 2024
}

@misc{hivvlsnum,
    author = {CDC},
    title = {{Monitoring selected national HIV prevention and
    care objectives by using HIV surveillance data—United States and 6 dependent areas, 2020.HIV Surveillance Supplemental Report 2022;27(No. 3). Revised edition}},
    howpublished  = "\url{http://www.cdc.gov/hiv/
    library/reports/hiv-surveillance.html}",
    note = "(Published August 2022. Accessed September 2023)",
    year = 2022
}

@misc{hivunawarenum,
    author = {CDC},
    title = {{Estimated HIV incidence and prevalence in the United
    States, 2015–2019. HIV Surveillance Supplemental Report 2021;26(No. 1)}},
    howpublished  = "\url{http://www.cdc.gov/
    hiv/library/reports/hiv-surveillance.html}",
    note = "(Published May 2021. Accessed September 2023)",
    year = 2021
}

@misc{hivprepnum,
    author = {AHEAD},
    title = {{PrEP Coverage}},
    howpublished  = "\url{https://ahead.hiv.gov/data/prep-coverage}",
    note = "(Accessed December 2023)",
    year = 2023
}

@article{kwak2021deep,
  title={Deep reinforcement learning approaches for global public health strategies for COVID-19 pandemic},
  author={Kwak, Gloria Hyunjung and Ling, Lowell and Hui, Pan},
  journal={PloS one},
  volume={16},
  number={5},
  pages={e0251550},
  year={2021},
  publisher={Public Library of Science San Francisco, CA USA}
}

@inproceedings{libin2021deep,
  title={Deep reinforcement learning for large-scale epidemic control},
  author={Libin, Pieter JK and Moonens, Arno and Verstraeten, Timothy and Perez-Sanjines, Fabian and Hens, Niel and Lemey, Philippe and Now{\'e}, Ann},
  booktitle={Joint European Conference on Machine Learning and Knowledge Discovery in Databases},
  pages={155--170},
  year={2021},
  organization={Springer}
}

@article{awasthi2022vacsim,
  title={Vacsim: Learning effective strategies for covid-19 vaccine distribution using reinforcement learning},
  author={Awasthi, Raghav and Guliani, Keerat Kaur and Khan, Saif Ahmad and Vashishtha, Aniket and Gill, Mehrab Singh and Bhatt, Arshita and Nagori, Aditya and Gupta, Aniket and Kumaraguru, Ponnurangam and Sethi, Tavpritesh},
  journal={Intelligence-Based Medicine},
  pages={100060},
  year={2022},
  publisher={Elsevier}
}

@article{kompella2020reinforcement,
  title={Reinforcement learning for optimization of COVID-19 mitigation policies},
  author={Kompella, Varun and Capobianco, Roberto and Jong, Stacy and Browne, Jonathan and Fox, Spencer and Meyers, Lauren and Wurman, Peter and Stone, Peter},
  journal={arXiv preprint arXiv:2010.10560},
  year={2020}
}

@article{kumar2021recurrent,
  title={Recurrent neural network and reinforcement learning model for COVID-19 prediction},
  author={Kumar, R Lakshmana and Khan, Firoz and Din, Sadia and Band, Shahab S and Mosavi, Amir and Ibeke, Ebuka},
  journal={Frontiers in public health},
  volume={9},
  year={2021},
  publisher={Frontiers Media SA}
}

@article{bednarski2021collaborative,
  title={On collaborative reinforcement learning to optimize the redistribution of critical medical supplies throughout the COVID-19 pandemic},
  author={Bednarski, Bryan P and Singh, Akash Deep and Jones, William M},
  journal={Journal of the American Medical Informatics Association},
  volume={28},
  number={4},
  pages={874--878},
  year={2021},
  publisher={Oxford University Press}
}

@article{weltz2022reinforcement,
  title={Reinforcement learning methods in public health},
  author={Weltz, Justin and Volfovsky, Alex and Laber, Eric B},
  journal={Clinical therapeutics},
  volume={44},
  number={1},
  pages={139--154},
  year={2022},
  publisher={Elsevier}
}

@article{mnih2013playing,
  title={Playing atari with deep reinforcement learning},
  author={Mnih, Volodymyr and Kavukcuoglu, Koray and Silver, David and Graves, Alex and Antonoglou, Ioannis and Wierstra, Daan and Riedmiller, Martin},
  journal={arXiv preprint arXiv:1312.5602},
  year={2013}
}

@inproceedings{haarnoja2018soft,
  title={Soft actor-critic: Off-policy maximum entropy deep reinforcement learning with a stochastic actor},
  author={Haarnoja, Tuomas and Zhou, Aurick and Abbeel, Pieter and Levine, Sergey},
  booktitle={International conference on machine learning},
  pages={1861--1870},
  year={2018},
  organization={PMLR}
}

@article{chu2019multi,
  title={Multi-agent deep reinforcement learning for large-scale traffic signal control},
  author={Chu, Tianshu and Wang, Jie and Codec{\`a}, Lara and Li, Zhaojian},
  journal={IEEE Transactions on Intelligent Transportation Systems},
  volume={21},
  number={3},
  pages={1086--1095},
  year={2019},
  publisher={IEEE}
}

@article{nasir2019multi,
  title={Multi-agent deep reinforcement learning for dynamic power allocation in wireless networks},
  author={Nasir, Yasar Sinan and Guo, Dongning},
  journal={IEEE Journal on Selected Areas in Communications},
  volume={37},
  number={10},
  pages={2239--2250},
  year={2019},
  publisher={IEEE}
}

@inproceedings{lin2018efficient,
  title={Efficient large-scale fleet management via multi-agent deep reinforcement learning},
  author={Lin, Kaixiang and Zhao, Renyu and Xu, Zhe and Zhou, Jiayu},
  booktitle={Proceedings of the 24th ACM SIGKDD international conference on knowledge discovery \& data mining},
  pages={1774--1783},
  year={2018}
}

@article{yu2020multi,
  title={Multi-agent deep reinforcement learning for HVAC control in commercial buildings},
  author={Yu, Liang and Sun, Yi and Xu, Zhanbo and Shen, Chao and Yue, Dong and Jiang, Tao and Guan, Xiaohong},
  journal={IEEE Transactions on Smart Grid},
  volume={12},
  number={1},
  pages={407--419},
  year={2020},
  publisher={IEEE}
}

@article{yang2020urban,
  title={Urban traffic control in software defined internet of things via a multi-agent deep reinforcement learning approach},
  author={Yang, Jiachen and Zhang, Jipeng and Wang, Huihui},
  journal={IEEE Transactions on Intelligent Transportation Systems},
  volume={22},
  number={6},
  pages={3742--3754},
  year={2020},
  publisher={IEEE}
}

@article{oroojlooy2022review,
  title={A review of cooperative multi-agent deep reinforcement learning},
  author={Oroojlooy, Afshin and Hajinezhad, Davood},
  journal={Applied Intelligence},
  pages={1--46},
  year={2022},
  publisher={Springer}
}

@article{kok2015optimizing,
  title={Optimizing an HIV testing program using a system dynamics model of the continuum of care},
  author={Kok, Sarah and Rutherford, Alexander R and Gustafson, Reka and Barrios, Rolando and Montaner, Julio SG and Vasarhelyi, Krisztina and Vancouver HIV Testing Program Modelling Group},
  journal={Health care management science},
  volume={18},
  pages={334--362},
  year={2015},
  publisher={Springer}
}

@article{gopalappa2012cost,
  title={Cost effectiveness of the National HIV/AIDS Strategy goal of increasing linkage to care for HIV-infected persons},
  author={Gopalappa, Chaitra and Farnham, Paul G and Hutchinson, Angela B and Sansom, Stephanie L},
  journal={JAIDS Journal of Acquired Immune Deficiency Syndromes},
  volume={61},
  number={1},
  pages={99--105},
  year={2012},
  publisher={LWW}
}

@article{lin2016cost,
  title={Cost effectiveness of HIV prevention interventions in the US},
  author={Lin, Feng and Farnham, Paul G and Shrestha, Ram K and Mermin, Jonathan and Sansom, Stephanie L},
  journal={American journal of preventive medicine},
  volume={50},
  number={6},
  pages={699--708},
  year={2016},
  publisher={Elsevier}
}

@article{lasry2011model,
  title={A model for allocating CDC’s HIV prevention resources in the United States},
  author={Lasry, Arielle and Sansom, Stephanie L and Hicks, Katherine A and Uzunangelov, Vladislav},
  journal={Health care management science},
  volume={14},
  pages={115--124},
  year={2011},
  publisher={Springer}
}

@article{zaric2001optimal,
  title={Optimal investment in a portfolio of HIV prevention programs},
  author={Zaric, Gregory S and Brandeau, Margaret L},
  journal={Medical Decision Making},
  volume={21},
  number={5},
  pages={391--408},
  year={2001},
  publisher={SAGE Publications Sage CA: Thousand Oaks, CA}
}

@article{lasry2012allocating,
  title={Allocating HIV prevention funds in the United States: recommendations from an optimization model},
  author={Lasry, Arielle and Sansom, Stephanie L and Hicks, Katherine A and Uzunangelov, Vladislav},
  journal={PloS one},
  volume={7},
  number={6},
  pages={e37545},
  year={2012},
  publisher={Public Library of Science San Francisco, USA}
}

@article{yaylali2016theory,
  title={From theory to practice: implementation of a resource allocation model in health departments},
  author={Yaylali, Emine and Farnham, Paul G and Schneider, Karen L and Landers, Stewart J and Kouzouian, Oskian and Lasry, Arielle and Purcell, David W and Green, Timothy A and Sansom, Stephanie L},
  journal={Journal of public health management and practice: JPHMP},
  volume={22},
  number={6},
  pages={567},
  year={2016},
  publisher={NIH Public Access}
}

@article{juusola2016hiv,
  title={HIV treatment and prevention: a simple model to determine optimal investment},
  author={Juusola, Jessie L and Brandeau, Margaret L},
  journal={Medical decision making},
  volume={36},
  number={3},
  pages={391--409},
  year={2016},
  publisher={SAGE Publications Sage CA: Los Angeles, CA}
}

@article{gromov2018numerical,
  title={Numerical optimal control for HIV prevention with dynamic budget allocation},
  author={Gromov, Dmitry and Bulla, Ingo and Silvia Serea, Oana and Romero-Severson, Ethan O},
  journal={Mathematical Medicine and Biology: A Journal of the IMA},
  volume={35},
  number={4},
  pages={469--491},
  year={2018},
  publisher={Oxford University Press}
}

@article{yaylali2018optimal,
  title={Optimal allocation of HIV prevention funds for state health departments},
  author={Yaylali, Emine and Farnham, Paul G and Cohen, Stacy and Purcell, David W and Hauck, Heather and Sansom, Stephanie L},
  journal={Plos one},
  volume={13},
  number={5},
  pages={e0197421},
  year={2018},
  publisher={Public Library of Science San Francisco, CA USA}
}

@article{matrajt2021vaccine,
  title={Vaccine optimization for COVID-19: Who to vaccinate first?},
  author={Matrajt, Laura and Eaton, Julia and Leung, Tiffany and Brown, Elizabeth R},
  journal={Science Advances},
  volume={7},
  number={6},
  pages={eabf1374},
  year={2021},
  publisher={American Association for the Advancement of Science}
}

@article{yu2021dynamic,
  title={Dynamic optimization of COVID-19 vaccine prioritization in the context of limited supply},
  author={Yu, Hongjie and Han, Shasha and Cai, Jun and Yang, Juan and Zhang, Juanjuan and Wu, Qianhui and Zheng, Wen and Shi, Huilin and Ajelli, Marco and Zhou, Xiao-Hua},
  year={2021}
}

@article{libotte2020determination,
  title={Determination of an optimal control strategy for vaccine administration in COVID-19 pandemic treatment},
  author={Libotte, Gustavo Barbosa and Lobato, Fran S{\'e}rgio and Platt, Gustavo Mendes and Neto, Ant{\^o}nio J Silva},
  journal={Computer methods and programs in biomedicine},
  volume={196},
  pages={105664},
  year={2020},
  publisher={Elsevier}
}

@article{han2021time,
  title={Time-varying optimization of COVID-19 vaccine prioritization in the context of limited vaccination capacity},
  author={Han, Shasha and Cai, Jun and Yang, Juan and Zhang, Juanjuan and Wu, Qianhui and Zheng, Wen and Shi, Huilin and Ajelli, Marco and Zhou, Xiao-Hua and Yu, Hongjie},
  journal={Nature communications},
  volume={12},
  number={1},
  pages={4673},
  year={2021},
  publisher={Nature Publishing Group UK London}
}

@article{tsay2020modeling,
  title={Modeling, state estimation, and optimal control for the US COVID-19 outbreak},
  author={Tsay, Calvin and Lejarza, Fernando and Stadtherr, Mark A and Baldea, Michael},
  journal={Scientific reports},
  volume={10},
  number={1},
  pages={10711},
  year={2020},
  publisher={Nature Publishing Group UK London}
}

@article{rawson2020and,
  title={How and when to end the COVID-19 lockdown: an optimization approach},
  author={Rawson, Thomas and Brewer, Tom and Veltcheva, Dessislava and Huntingford, Chris and Bonsall, Michael B},
  journal={Frontiers in Public Health},
  volume={8},
  pages={262},
  year={2020},
  publisher={Frontiers Media SA}
}

@article{sansom2021optimal,
  title={Optimal allocation of societal HIV prevention resources to reduce HIV incidence in the United States},
  author={Sansom, Stephanie L and Hicks, Katherine A and Carrico, Justin and Jacobson, Evin U and Shrestha, Ram K and Green, Timothy A and Purcell, David W},
  journal={American Journal of Public Health},
  volume={111},
  number={1},
  pages={150--158},
  year={2021},
  publisher={American Public Health Association}
}

@article{pino2023optimization,
  title={An optimization model with simulation for optimal regional allocation of COVID-19 vaccines},
  author={Pino, Rodney and Mendoza, Victoria May and Enriquez, Erika Antonette and Velasco, Arrianne Crystal and Mendoza, Renier},
  journal={Healthcare Analytics},
  volume={4},
  pages={100244},
  year={2023},
  publisher={Elsevier}
}

@article{olayiwola2023caputo,
  title={A Caputo fractional order epidemic model for evaluating the effectiveness of high-risk quarantine and vaccination strategies on the spread of COVID-19},
  author={Olayiwola, Morufu Oyedunsi and Alaje, Adedapo Ismaila and Olarewaju, Akeem Yunus and Adedokun, Kamilu Adewale},
  journal={Healthcare Analytics},
  volume={3},
  pages={100179},
  year={2023},
  publisher={Elsevier}
}

@article{silveira2024multi,
  title={A multi-stage optimization model for managing epidemic outbreaks and hospital bed planning in Intensive Care Units},
  author={Silveira, Ingrid Machado and de Freitas Almeida, Jo{\~a}o Fl{\'a}vio and Pinto, Luiz Ricardo and Epaminondas, Luiz Ant{\^o}nio Resende and Concei{\c{c}}{\~a}o, Samuel Vieira and Machado, Elaine Leandro},
  journal={Healthcare Analytics},
  volume={5},
  pages={100342},
  year={2024},
  publisher={Elsevier}
}

@article{raffin2021stablebaselines3,
  title={Stable-Baselines3: Reliable Reinforcement Learning Implementations},
  author={Raffin, Antonin and Hill, Ashley and Gleave, Adam and Kanervisto, Anssi and Ernestus, Maximilian and Dormann, Noah},
  journal={Journal of Machine Learning Research},
  volume={22},
  number={268},
  pages={1--8},
  year={2021}
}

\end{document}